\documentclass[runningheads]{llncs}

\usepackage{eccv}

\usepackage{eccvabbrv}

\usepackage{graphicx}
\usepackage{booktabs}

\usepackage[accsupp]{axessibility}  

\usepackage{hyperref}

\usepackage{orcidlink}

\usepackage{amsmath}
\usepackage{amssymb}
\usepackage{bbm}
\usepackage{multirow}
\usepackage{tabularx}
\usepackage{siunitx}
\usepackage{xspace}
\usepackage{acronym}
\usepackage{pifont}
\usepackage{subcaption}
\usepackage{xcolor}
\usepackage{listings}
\definecolor{codebg}{RGB}{245,245,245}
\usepackage{colortbl}
\usepackage{ragged2e}
\usepackage{rotating}
\usepackage[export]{adjustbox}
\usepackage{caption}

\usepackage{tikz}
\usetikzlibrary{positioning}
\usetikzlibrary{shapes.geometric, shapes, arrows, arrows.meta, bending}
\usetikzlibrary{backgrounds}
\usetikzlibrary{spy}
\usepackage{pgfplots}
\usepackage{pgfplotstable}
\pgfplotsset{compat=1.18}

\usepackage{amsmath,mathtools}
\usepackage{algorithm}
\usepackage{algpseudocode}
\usepackage{soul}
\usepackage[normalem]{ulem}
\usepackage{array}

\definecolor{colOurs}{HTML}{1f77b4}
\definecolor{colTensorRT}{HTML}{ff7f0e}
\definecolor{colGeneric}{HTML}{2ca02c}
\definecolor{colGeometric}{HTML}{d62728}
\definecolor{colExtra}{HTML}{9467bd}

\definecolor{codeblue}{RGB}{0,0,180}
\definecolor{codegreen}{RGB}{0,140,0}
\definecolor{codecomment}{RGB}{0,120,120}
\definecolor{codered}{RGB}{180,30,30}

\newcommand{\getcolor}[1]{%
  \ifstrequal{#1}{Ours}{\color{colOurs}}{%
  \ifstrequal{#1}{Ours w/ TensorRT}{\color{colTensorRT}}{%
  \ifstrequal{#1}{Generic methods}{\color{colGeneric}}{%
  \ifstrequal{#1}{Geometric assumptions}{\color{colGeometric}}{%
  \ifstrequal{#1}{Extra data}{\color{colExtra}}{\color{black}}}}}%
}}

\newcommand{\baseline}{YOLOv10-M3D\xspace}
\newcommand{\ourmethod}{LeAD-M3D\xspace}

\newcommand{\ourmethodn}{\ourmethod N\xspace}
\newcommand{\ourmethods}{\ourmethod S\xspace}
\newcommand{\ourmethodm}{\ourmethod M\xspace}
\newcommand{\ourmethodb}{\ourmethod B\xspace}
\newcommand{\ourmethodx}{\ourmethod X\xspace}

\newcommand{\baselinen}{\baseline N\xspace}
\newcommand{\baselines}{\baseline S\xspace}
\newcommand{\baselinem}{\baseline M\xspace}
\newcommand{\baselineb}{\baseline B\xspace}
\newcommand{\baselinex}{\baseline X\xspace}

\newcommand{\noCode}{$^\dagger$}

\newcommand{\APDDDRFS}{$\text{AP}_{\text{3D$|$R40}}^{0.7}$}
\newcommand{\APBEVRFS}{$\text{AP}_{\text{BEV$|$R40}}^{0.7}$}
\newcommand{\APDDDRFF}{$\text{AP}_{\text{3D$|$R40}}^{0.5}$}

\newcommand{\APDDDS}{$\text{AP}_{\text{3D}}^{0.7}$}
\newcommand{\APDDDF}{$\text{AP}_{\text{3D}}^{0.5}$}
\newcommand{\APHDDDF}{$\text{APH}_{\text{3D}}^{0.5}$}
\newcommand{\APDDDRF}{$\text{AP}_{\text{3D$|$R40}}$}
\newcommand{\APBEVRF}{$\text{AP}_{\text{BEV$|$R40}}$}
\newcommand{\APH}{$\text{APH}_{\text{3D}}$}
\newcommand{\AP}{$\text{AP}_{\text{3D}}$}

\newcommand{\cbarm}[4]{%
  {\begingroup\color{#1}\rule{#2}{#3}\endgroup}%
  \hspace{0.5em}#4%
}

\newcolumntype{L}{>{\RaggedRight\arraybackslash}X}

\newcommand{\midrulegray}{\arrayrulecolor[rgb]{0.8, 0.8, 0.8} \midrule \arrayrulecolor{black}}

\newcommand{\cmidrulewaymogray}{\arrayrulecolor[rgb]{0.8, 0.8, 0.8} \cmidrule{2-18} \arrayrulecolor{black}}

\definecolor{tGreenStrong}{HTML}{52B69A} 
\definecolor{tGreenLight}{HTML}{ACDCCF} 
\definecolor{tRedStrong}{HTML}{FF595E} 
\definecolor{tRedLight}{HTML}{FF595E} 
\definecolor{tBlueStrong}{HTML}{064789} 
\definecolor{tBlueLight}{HTML}{56A7F8} 
\definecolor{tBlueLightLight}{HTML}{AED5FC} 

\definecolor{qualGT}{HTML}{63B59A}
\definecolor{qualFOV}{HTML}{A098A6}
\definecolor{qualBL}{HTML}{F45B57}
\definecolor{qualOUR}{HTML}{FAC91B}

\definecolor{qualGTcr}{RGB}{244,179,1}
\definecolor{qualBLcr}{RGB}{219,16,72}
\definecolor{qualOURcr}{RGB}{16,85,154}

\newcommand \colorindicator[1]{%
	{\textcolor{#1}{$\blacksquare\!\!\!\!\!\blacksquare$}}%
}

\newcommand{\best}{\cellcolor{tBlueLight}}
\newcommand{\second}{\cellcolor{tBlueLightLight}}

\definecolor{codeblue}{RGB}{0,0,180}
\definecolor{codegreen}{RGB}{0,140,0}
\definecolor{codecomment}{RGB}{0,120,120}
\definecolor{codered}{RGB}{180,30,30}

\lstdefinelanguage{mypython}{
    language=Python,
    morekeywords={self},
    morekeywords={[2]@invariant,cgi_3d}, 
    keywordstyle={[2]\color{codeblue}\bfseries}, 
}

\lstdefinestyle{pythonstyle}{
    language=mypython,
    backgroundcolor=\color{white},
    basicstyle=\ttfamily\footnotesize,
    keywordstyle=\color{codegreen}\bfseries,
    commentstyle=\color{codecomment}\itshape,
    stringstyle=\color{codered}\itshape,
    identifierstyle=\color{black},
    frame=none,
    keepspaces=true,
    framesep=0mm,
    showstringspaces=false,
    breaklines=true,
    tabsize=1,
    lineskip=-2pt
}

\newcommand{\cmark}{\ding{51}}%
\newcommand{\xmark}{\ding{55}}%

\definecolor{red1}{HTML}{FF595E}
\definecolor{yellow1}{HTML}{FFCA3A}
\definecolor{green1}{HTML}{52B69A}
\definecolor{blue1}{HTML}{064789}
\definecolor{purple1}{HTML}{805D93}

\begin{document}

\title{LeAD-M3D: Leveraging Asymmetric Distillation\\ for Real-Time Monocular 3D Detection} 

\titlerunning{LeAD-M3D}


\author{
Johannes Meier\,\textsuperscript{1,2,3,4,\dag,*\,}\orcidlink{0009-0000-2227-8271}\and
Jonathan Michel\,\textsuperscript{3,4,\dag\,}\orcidlink{0009-0004-2314-8283}\and
Oussema Dhaouadi\,\textsuperscript{1,2,3,4,5\,}\orcidlink{0009-0008-6842-5220}\and
Yung-Hsu Yang\,\textsuperscript{2\,}\orcidlink{0000-0003-0044-515X}\and
Christoph Reich\,\textsuperscript{3,4,6,7\,}\orcidlink{0000-0002-8616-1627}\and
Zuria Bauer\,\textsuperscript{2\,}\orcidlink{0000-0001-8447-2344}\and
Stefan Roth\,\textsuperscript{6,7,8\,}\orcidlink{0000-0001-9002-9832}\and
Marc Pollefeys\,\textsuperscript{2,9\,}\orcidlink{0000-0003-2448-2318}\and
Jacques Kaiser\,\textsuperscript{1\,}\orcidlink{0000-0001-7487-6185}\and
Daniel Cremers\,\textsuperscript{3,4,7\,}\orcidlink{0000-0002-3079-7984}
}

\authorrunning{J. Meier et al.}



\institute{
\textsuperscript{1}\,DeepScenario\;\;\;\;
\textsuperscript{2}\,ETH Zurich\;\;\;\;
\textsuperscript{3}\,TU Munich\;\;\;\;
\textsuperscript{4}\,MCML\;\;\;\;\\
\textsuperscript{5}\,University of Cambridge\;\;\;
\textsuperscript{6}\,TU Darmstadt\;\;\;
\textsuperscript{7}\,ELIZA\;\;\;
\textsuperscript{8}\,hessian.AI\;\;\;
\textsuperscript{9}\,Microsoft\\
\url{https://deepscenario.github.io/LeAD-M3D/}
}


\maketitle
\footnotetext[1]{j.meier@tum.de\space\space\space \textsuperscript{\dag} Equal contribution} 
\begin{abstract}
Real-time monocular 3D object detection remains challenging due to severe depth ambiguity, viewpoint shifts, and the high computational cost of 3D reasoning. Existing approaches either rely on LiDAR or geometric priors to compensate for missing depth or sacrifice efficiency to achieve competitive accuracy. We introduce LeAD-M3D, a monocular 3D detector that achieves state-of-the-art accuracy and real-time inference without extra modalities. Our method is enabled by three key components. Asymmetric Augmentation Denoising Distillation (A2D2) transfers geometric knowledge from a clean-image teacher to a MixUp-noised student via a quality- and importance-weighted depth-feature loss, enabling stronger depth reasoning without LiDAR. 3D-aware Consistent Matching (CM$_{\text{3D}}$) improves prediction-to-ground truth assignment by integrating 3D MGIoU into the matching score, yielding stable and precise supervision. Finally, Confidence-Gated 3D Inference (CGI$_{\text{3D}}$) accelerates inference by restricting expensive 3D regression to confident regions. Together, these contributions set a new Pareto frontier for monocular 3D detection: LeAD-M3D achieves state-of-the-art accuracy on KITTI and Waymo, and the best reported car AP on Rope3D, while running up to \num{3.6}\,$\times$ faster than prior high-accuracy models (\eg, MonoDiff). LeAD-M3D demonstrates that high fidelity and real-time monocular 3D detection is simultaneously attainable, without LiDAR, stereo, or strong geometric assumptions.
\end{abstract}
\begin{figure}[ht]
    \centering
    \centering 

\pgfdeclareplotmark{starourrt}{
    \node[star, star point ratio=2.0, minimum size=5pt,
          inner sep=0pt,draw=black,solid,fill=red1] {};
}
\pgfdeclareplotmark{starour}{
    \node[star, star point ratio=2.0, minimum size=5pt,
          inner sep=0pt,draw=black,solid,fill=yellow1] {};
}

\begin{tikzpicture}[clip, every node/.style={font=\sffamily\footnotesize}, >={Stealth[inset=0pt,length=2pt,angle'=45]}]
    \tikzset{every picture/.style={/utils/exec={\sffamily\footnotesize}}}
    \tikzset{every major tick/.append style={line width=.5pt, major tick length=3.5pt, gray!85}}
	\begin{axis}[
        enlargelimits=false,
        clip=true,
        height=0.325\textwidth,
        xlabel shift=-2.95pt,
        width=0.95\textwidth,
        grid=both,
        xtick pos=bottom,
        ytick pos=left,
        grid style={line width=.08pt, draw=white, dash pattern=on 1pt off 1pt},
        major grid style={line width=.2pt,draw=gray!65},
        minor tick num=1,
        xmin=0,
        xmax=90,
        ylabel shift=2.95pt,
        xtick={
            0, 10, 20, 30, 40, 50, 60, 70, 80, 90, 100
        },
        xticklabels={
            0, 10, 20, 30, 40, 50, 60, 70, 80, 90, 100
        },
        ticklabel style = {font=\fontsize{7}{2}},
        ylabel=$\text{AP}_{\text{3D$|$R40}}^{\text{0.7}}$,
        xlabel=Runtime (in ms),
        ymin=13.5,
        ymax=22.5,
        ytick={14, 16, 18, 20, 22},
        yticklabels={14, 16, 18, 20, 22},
        nodes near coords,
        point meta=explicit symbolic, 
        every node near coord/.append style={
            font=\tiny, color=black, anchor=west, xshift=1.0pt, yshift=-1.0pt
        },
        ]
    	\addplot[color=gray!70, mark=starourrt, mark size=2.0pt, mark options={draw=red1, fill=red1}] table[x=t,y=ap, meta=name] {ourrtdata.dat};\label{pgfplots:ourrt};
        \addplot[color=gray!70, mark=starour, mark size=2.0pt, mark options={draw=yellow1, fill=yellow1}] table[x=t,y=ap, meta=name] {oursdata.dat};\label{pgfplots:our};
        
        \addplot[color=green1, only marks, mark=square*, mark size=1.35pt] coordinates {(270.0, 19.43)} node[anchor=south east, font=\tiny, black] {MonoTAKD};\label{pgfplots:lidar};
        \addplot[color=green1, only marks, mark=square*, mark size=1.35pt] coordinates {(30.1, 16.82)} node[anchor=south west, font=\tiny, black] {DK3D};
        \addplot[color=green1, only marks, mark=square*, mark size=1.35pt] coordinates {(36.0, 16.70)} node[anchor=north west, font=\tiny, black] {MonoSG};
        \addplot[color=green1, only marks, mark=square*, mark size=1.35pt] coordinates {(213.9, 17.02)} node[anchor=south east, font=\tiny, black] {OccupancyM3D};
        \addplot[color=green1, only marks, mark=square*, mark size=1.35pt] coordinates {(25.7, 16.13)} node[anchor=north, font=\tiny, black] {MonoSTL}; 
        \addplot[color=green1, only marks, mark=square*, mark size=1.35pt] coordinates {(56, 16.46)} node[anchor=north, font=\tiny, black] {MonoFG};  
        \addplot[color=green1, only marks, mark=square*, mark size=1.35pt] coordinates {(1380.3, 17.13)} node[anchor=south east, font=\tiny, black] {MonoNERD};
        \addplot[color=green1, only marks, mark=square*, mark size=1.35pt] coordinates {(46.7, 16.03)} node[anchor=north, xshift=-1pt, font=\tiny, black] {MonoDistill}; 
        
        \addplot[color=blue1, only marks, mark=triangle*, mark size=1.5pt] coordinates {(68.1, 18.72)} node[anchor=south, font=\tiny, black] {MonoDGP};\label{pgfplots:geometry}; 
        \addplot[color=blue1, only marks, mark=triangle*, mark size=1.5pt] coordinates {(23.2, 16.73)} node[anchor=south, font=\tiny, black] {MonoUNI};
        \addplot[color=blue1, only marks, mark=triangle*, mark size=1.5pt] coordinates {(56, 17.37)} node[anchor=south, font=\tiny, black] {MonoATT}; 
        \addplot[color=blue1, only marks, mark=triangle*, mark size=1.5pt] coordinates {(29, 16.14)} node[anchor=west, font=\tiny, black] {PDR}; 
        
        \addplot[color=purple1, only marks, mark=pentagon*, mark size=1.5pt] coordinates {(82.0, 18.85)} node[anchor=north, font=\tiny, black] {GATE3D};\label{pgfplots:noextra};  
        \addplot[color=purple1, only marks, mark=pentagon*, mark size=1.5pt] coordinates {(38.0, 18.84)} node[anchor=south, font=\tiny, black] {MonoMAE}; 
        \addplot[color=purple1, only marks, mark=pentagon*, mark size=1.5pt] coordinates {(86.0, 21.02)} node[anchor=east, font=\tiny, black] {MonoDiff}; 
        \addplot[color=purple1, only marks, mark=pentagon*, mark size=1.5pt] coordinates {(20.2, 19.15)} node[anchor=west, font=\tiny, black] {MonoLSS}; 
        \addplot[color=purple1, only marks, mark=pentagon*, mark size=1.5pt] coordinates {(40.0, 17.12)} node[anchor=south, font=\tiny, black] {FD3D}; 
        \addplot[color=purple1, only marks, mark=pentagon*, mark size=1.5pt] coordinates {(15.5, 16.36)} node[anchor=west, font=\tiny, black] {DDML}; 
        \addplot[color=purple1, only marks, mark=pentagon*, mark size=1.5pt] coordinates {(47.8, 16.47)} node[anchor=south, font=\tiny, black] {MonoDETR}; 
        \addplot[color=purple1, only marks, mark=pentagon*, mark size=1.5pt] coordinates {(27.3, 15.01)} node[anchor=north, font=\tiny, black] {Cube R-CNN}; 
        
	\end{axis}
    
    \node[draw,fill=white, text width=6.85cm, inner sep=0.5pt, anchor=center] at (0.41\textwidth, 2.6) {\renewcommand{\arraystretch}{0.875}\setlength{\tabcolsep}{2.0pt}\scriptsize
    \begin{tabular}{
    clcl
    }
    \;\;\;\;\;\;\;\;\ref*{pgfplots:our} & \sffamily \ourmethod\vphantom{$i^i$} & \ref*{pgfplots:ourrt} & \sffamily \ourmethod (TensorRT) \\
    \end{tabular}\\[-2.2pt]
    \begin{tabular}{
    lclclcl
    }
    \sffamily Baselines: & \ref*{pgfplots:noextra} & (w/o extra) & \ref*{pgfplots:geometry} & \sffamily (w/ geom.) & \ref*{pgfplots:lidar} & \sffamily (w/ LiDAR, \etc) \\
    \end{tabular}};
    \draw[<-] (0.2, -0.45) -- node[midway, below, yshift=2pt] {\tiny lower better} (1.2, -0.45);
    \draw[->] (-0.535, 1.4) -- node[midway, above, xshift=2pt, rotate=90] {\tiny higher better} (-0.535, 2.4);
\end{tikzpicture}


    \caption{
        \textbf{Runtime \vs Accuracy} on KITTI \emph{test}, using \APDDDRFS{} Mod. (in \%, $\uparrow$) and runtime (in ms, $\downarrow$).
        We provide a model family (sizes N to X) to balance runtime and 3D detection accuracy.
        \ourmethod{} offers a Pareto frontier over existing approaches. 
        Using TensorRT further improves the runtime, enabling $>$\SI{60}{FPS} real-time inference of even our largest model size (X). Runtime is reported on the same hardware (NVIDIA RTX 8000) wherever possible (\ie, code is publicly available).}
    \label{fig:teaser}
\end{figure}

\section{Introduction}
\label{sec:introduction}
Monocular 3d object detection (M3D) aims to predict the 3D position, orientation, and size of a scene's objects from a \emph{single} RGB image.
This setup is broadly available and M3D finds widespread application in autonomous driving~\cite{progress_driving,progress_driving2}, robotics~\cite{app_robotics}, medicine~\cite{app_medical}, and city infrastructure~\cite{app_traffic_monitoring}.
However, inferring depth information from a single 2D image introduces strong ambiguities, and is the dominant source of error in M3D~\cite{monodle,monocd}.
This problem intensifies for non-zero roll and pitch viewpoints (\eg, roadside camera views) \cite{cdrone,rope3d}.
Moreover, recent methods emphasize accuracy over runtime efficiency \cite{monodiff,monotakd}, hindering deployment in real-world scenarios.
A model that learns exclusively from 3D box supervision, providing accurate detections and efficient real-time inference, remains currently lacking.


To obtain both accurate and efficient models, knowledge distillation (KD) has been shown to be an effective paradigm~\cite{survey_knowledge_distillation, survey_knowledge_distillation2}. In KD, a large and computationally expensive teacher model transfers its knowledge to a smaller, more efficient student model~\cite{survey_knowledge_distillation, survey_knowledge_distillation2}. This process yields a student model that retains high accuracy while significantly reducing computational cost. To perform effective KD for M3D, existing approaches create a teacher-student asymmetry by giving the teacher access to additional information, typically in the form of LiDAR~\cite{hsrdn,monodistill,cmkd,monotakd}. However, LiDAR is not always available in real-world applications, and the use of multiple modalities (\eg, imagery and LiDAR) leads to complex pipelines. Here, we propose Asymmetric Augmentation Denoising Distillation (A2D2), an image-only KD pipeline that leverages an asymmetric denoising distillation~\cite{knowledge_disstill} strategy for efficient and accurate M3D. Unlike prior M3D distillation pipelines, A2D2 is conceptually simpler as we do not rely on additional modalities (\eg, LiDAR).

Using our A2D2, we build an M3D model family composed of five models (\cf \cref{fig:teaser}) with different model sizes, \ie, N\ /\ S\ /\ M\ /\ B\ /\ X, following the YOLO family~\cite{yolov10}. In particular, we train our largest variant (X) as a teacher model to provide expressive target representations. We freeze the teacher and distill different student models using A2D2 by denoising a MixUp-based~\cite{mixup} objective. Denoising is performed on object-specific depth features. By using a dynamic feature loss that weights features according to the teacher's prediction quality and importance, we allow for effective feature distillation. This enables the student to focus learning on reliable and informative cues, improving the student's depth reasoning and detection accuracy.

To complement our A2D2, we introduce two additional components.
\emph{First}, we introduce 3D-aware Consistent Matching (CM$_{\text{3D}}$), which improves the prediction-to-ground truth assignment by incorporating a 3D Intersection over Union (IoU) term into the matching criterion.
This joint 2D-3D alignment allows the model to learn from supervision that better reflects the actual 3D localization quality.
\emph{Second}, we propose Confidence-Gated 3D Inference (CGI$_{\text{3D}}$), a lightweight inference strategy that restricts the costly 3D regression head to high-confidence regions identified by an efficient 2D classifier.
This reduces redundant computation and accelerates inference with virtually no loss in accuracy.

Together, we present \textbf{\ourmethod}, \textbf{Le}veraging \textbf{A}symmetric \textbf{D}istillation for Real-time \textbf{M}onocular \textbf{3}D \textbf{D}etection. Our \ourmethod model family yields a new accuracy-efficiency Pareto frontier over existing M3D models (\cf \cref{fig:teaser}) \emph{without} relying on LiDAR, stereo, or ground-plane inputs. For example, on KITTI~\cite{kitti} we achieve the highest \AP\space across all difficulty levels among monocular methods and even surpass LiDAR- and geometry-assisted approaches, while running 3.6$\times$ faster than the next-best high-accuracy model (MonoDiff~\cite{monodiff}). \ourmethod also achieves strong cross-view and domain generalization results, all while running in real-time on a single GPU.

Our main contributions can be summarized as follows: \emph{(i)}~We propose A2D2, a novel, LiDAR-free knowledge-distillation scheme for M3D. A2D2 transfers knowledge using MixUp-based information asymmetry and a quality- and impor\-tance-weighted feature loss, strengthening the student's depth reasoning without requiring privileged depth information. \emph{(ii)}~We introduce CM$_{\text{3D}}$, a prediction-to-ground truth assignment strategy that improves supervision by employing 3D cues. \emph{(iii)}~We propose CGI$_{\text{3D}}$, a lightweight inference strategy that restricts costly 3D regression to high-confidence regions, effectively reducing inference efficiency without a degradation in detection accuracy. \emph{(iv)}~We demonstrate a new accuracy-efficiency Pareto frontier (\cf \cref{fig:teaser}) of \ourmethod across a wide range of datasets, without relying on LiDAR, stereo, or geometric priors (\eg, inverse height-depth consistency). Additionally, \ourmethod leads to strong cross-view and domain generalization results.

\section{Related Work}
\label{sec:related_work}

\subsubsection{Monocular 3D Object Detection (M3D).} Existing M3D methods can be broadly grouped into \emph{three} categories.
The \emph{first} uses extra data, such as LiDAR~\cite{mononerd,occupancym3d,cmkd,CaDDN,add,3dmood}, object shape cues~\cite{dcd,autoshape}, temporal consistency~\cite{tempm3d}, ground planes~\cite{cobev,mose}, or stereo images~\cite{monosg}, during training to compensate for sparse 3D box supervision.
The dependency on additional data limits the applicability to real-world scenarios where these modalities are unavailable, \eg, LiDAR is unavailable in certain traffic~\cite{roscenes,cityflow} or drone~\cite{cdrone,dsc3d} scenarios. While LiDAR is often used to aid the annotation process, recent work removed LiDAR entirely from the M3D annotation pipeline~\cite{dsc3d}.
The \emph{second} category of M3D models injects geometric constraints to regularize depth~\cite{monocd,MoGDE,monouni,monodgp,gupnet,monoflex,monodde}.
A common assumption is that the apparent 2D height of an object is inversely proportional to its depth, which yields a scale prior by relating the estimated 3D height to the 2D height~\cite{monoflex,gupnet}. Current approaches~\cite{monocd,MoGDE,monouni,monodgp} encode variants of perspective geometric priors.
These assumptions largely hold under car-view settings with near-zero roll and pitch but break for different viewpoints, restricting deployment to forward-facing automotive cameras~\cite{cdrone,roscenes,rope3d}.
The \emph{third} category relies only on 3D bounding boxes for supervision without auxiliary modalities or hard-coded priors~\cite{monolss,monomae,gate3d,QD3DT,cc3dt}.
We focus on this category of models, maximizing applicability across camera viewpoints and datasets.
Our approach fits this family, as it allows full flexibility with respect to object orientation, explicitly supporting arbitrary $\operatorname{SO}(3)$ rotations rather than being constrained by viewpoint-specific priors like the inverse height-depth consistency.

\subsubsection{Knowledge Distillation (KD)}~\cite{distill_bert,survey_knowledge_distillation,survey_knowledge_distillation2,knowledge_distillation_od} is a general technique for transferring knowledge from a large, high-capacity, and accurate teacher model to a smaller, more efficient student model. Introducing asymmetry into the distillation process has been shown to improve the student's accuracy~\cite{distill_bert,survey_knowledge_distillation,survey_knowledge_distillation2,knowledge_distillation_od}. Previous M3D methods achieve this teacher-student asymmetry by giving the teacher an easier problem and/or more informative data, \eg, supplying LiDAR, allowing the monocular student to learn high-quality 3D geometric cues~\cite{hsrdn,monodistill,cmkd,monotakd}.
On the other hand, ADD~\cite{add} and FD3D~\cite{fd3d} remove the LiDAR dependency by providing the teacher with ground-truth object positions or object-wise depth maps.
Our approach takes a different angle on this paradigm.
Instead of giving the teacher more information, we give the student strongly augmented MixUp images~\cite{mixup}, while the teacher receives both clean images. Compared to existing M3D distillation approaches, our approach is conceptually simpler, as we do not use multiple modalities. This also allows both student and teacher to share the same meta-architecture, reducing overall architectural complexity.

\subsubsection{Real-Time Object Detection.} Significant progress has been made in 2D real-time detection~\cite{yolo_series,yolov12,yolov13,lw-detr,rt-detr,rt_detrv2,rt_detrv3,d-fine} on the accuracy-efficiency frontier.
Among real-time 2D detection approaches, the YOLO series~\cite{yolo_series,yolov12,yolov13} has continuously improved the accuracy-efficiency frontier. We base our real-time M3D model on YOLOv10~\cite{yolov10}, which enables accurate, efficient, and non-maximum suppression-free inference.
Existing M3D methods require long~\cite{CaDDN,mononerd,occupancym3d} or moderate inference times~\cite{monolss,monocon}, and most provide only a single model size~\cite{monocd,gupnet,monolss}.
To address this, we offer a model family with five model sizes where accuracy is improved through A2D2 and task-aligned learning without increasing inference cost, while CGI$_{\text{3D}}$ optimizes runtime. This allows our second-largest model (B) to outperform the previous fastest baseline in terms of speed.

\begin{figure*}[t]
  \centering
  \includegraphics[width=1.0\textwidth, trim={0.325cm 0.2cm 0.32cm 0}, clip]{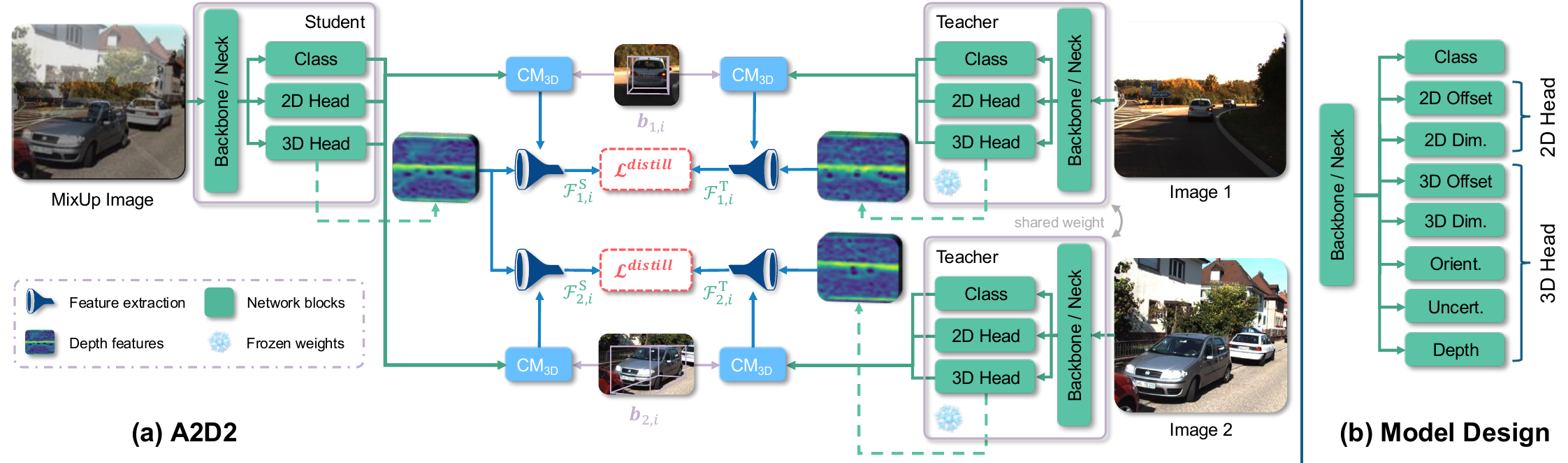}
  \caption{
      \textbf{Overview of \ourmethod.}
      \emph{(a)} We distill high-dimensional instance-depth features from a large teacher to a compact student.
      To create an information gap, the teacher sees clean images.
      The student receives a MixUp image and must reproduce the teacher's intermediate features (\cref{sec:method:distill}).
      This frames distillation as a denoising task, which removes MixUp-induced artifacts.
      CM$_{\text{3D}}$ uses ground truth to pair corresponding teacher and student predictions (\cf \cref{sec:tal}).
      \emph{(b)} Detailed model architecture.
      \emph{Dim.} stands for dimension head, \emph{Orient.} denotes for orientation head, and \emph{Uncert.} stands the depth uncertainty head.
    }
  \label{fig:method_overview}
\end{figure*}
\section{Method: Accurate \& Real-Time M3D}
\label{sec:method}

\subsubsection{Problem Statement.}
\label{sec:method:problem}
The goal of M3D is to predict 3D bounding boxes and object categories from a single RGB image.
Formally, let $\mathbf{I}\in\mathbb{R}^{3\times H \times W}$ be an RGB image and let $\hat{B}(\mathbf{I}) = \{\mathbf{\hat{b}}^{\text{3D}}_1, \ldots, \mathbf{\hat{b}}^{\text{3D}}_{M'}\}$ denote the set of estimated 3D bounding boxes for objects in $\mathbf{I}$.
Each box $\mathbf{\hat{b}}^{\text{3D}}_i$ is defined by its 3D center location $(x_i, y_i, z_i) \in \mathbb{R}^3$, 3D size $(w_i, h_i, l_i) \in \mathbb{R}^3$, orientation represented by a rotation matrix $R_{i} \in \operatorname{SO}(3)$, and category $c_i \in \mathbb{C}$ (\eg, ``Vehicle'' or ``Pedestrian'').

\subsubsection{Overview.}
\label{sec:method:overview}
Prior M3D methods largely optimize accuracy, while runtime efficiency is often secondary.
We address this by introducing LeAD-M3D, \textbf{Le}veraging \textbf{A}symmetric \textbf{D}istillation for Real-Time \textbf{M}onocular \textbf{3}D \textbf{D}etection, a novel M3D method built on YOLOv10~\cite{yolov10}.
As shown in \cref{fig:method_overview}b, we follow~\cite{monolss} and extend YOLOv10 with standard 3D detection heads (\ie, 3D offset, 3D dimension, orientation, depth, and uncertainty heads), a postprocessing pipeline, and 3D losses.
This baseline establishes an efficient architectural foundation for M3D, which we denote as \baseline.
We show more details in the supplementary material. Our overall pipeline (\cf \cref{fig:method_overview}a) further extends the baseline M3D model with three proposed components.
A2D2 enhances standard KD techniques by integrating a novel augmentation-based denoising task.
CM$_{\text{3D}}$ lifts the standard 2D prediction-to-ground truth assignment to 3D space.
Finally, CGI$_{\text{3D}}$ reduces head FLOPs during inference without impacting the predictions.

\subsection{Asymmetric Augmentation Denoising Distillation (A2D2)}
\label{sec:method:distill}
KD is an effective approach to balance accuracy with real-time requirements. KD methods for M3D often create an information asymmetry, easing the teacher's task and/or complicating the student's task.
As depth estimation is the core bottleneck in M3D, these methods~\cite{monodistill,hsrdn,cmkd} simplify the teacher's task enormously by adding LiDAR~\cite{hsrdn,monodistill,cmkd,add,fd3d}.
This further strengthens the transfer for M3D yet introduces a modality dependency.
Moreover, standard KD losses transfer features uniformly~\cite{add,fd3d,cmkd,hsrdn,monodistill},
ignoring variation in the teacher's quality and the varying influence of feature channels on the final depth.
To address these issues, we propose Asymmetric Augmentation Denoising Distillation (A2D2), a LiDAR-free KD scheme that couples an asymmetric MixUp-based denoising task with quality- and importance-weighted feature alignment.

\subsubsection{MixUp-Based Information Asymmetry.}
Unlike previous methods, we use MixUp \cite{monolss, cdrone} to augment the student input, enabling the student to learn to match the teacher's instance-depth features despite augmentation, creating a desired asymmetry without requiring extra modalities.
In particular, we blend two images at the pixel level and require the student to detect all objects for both images.
Unlike most other augmentation strategies, MixUp preserves the full spatial extent of all objects in both images while still creating a denoising task, so no GT annotations are lost.
Crucially, MixUp preserves object geometry in image coordinates, \ie, projected centers, depths, dimensions, and orientation remain consistent.
We exploit this invariance for distillation as a denoising task in feature space.

As shown in \cref{fig:method_overview}a, first, we feed two clean images to the teacher and the corresponding MixUp image to the student.
Second, we use 3D-aware Consistent Matching (CM$_{\text{3D}}$) to assign each ground-truth object to its best-matching prediction from the teacher and separately to its best match from the student (\cref{sec:tal}).
It is worth noting that we distill the depth features from the depth head rather than generic backbone activations, where the scalar depth is obtained by regression using the teachers weight matrix $W^{\rm T}\in\mathbb{R}^{Q}$ and the teachers depth features $\mathcal{F}_{i}^{\rm T}\in\mathbb{R}^{Q}$, where $i$ denotes a specific object in the image.
This establishes explicit student-teacher pairs for the same underlying object, enabling feature-level distillation between corresponding predictions.
We found that depth features yield more effective transfer for the 3D task than generic backbone features.

\subsubsection{Quality- and Importance-Weighted Distillation Loss.}
The standard KD losses ignore variation in teacher quality and the unequal importance of feature channels. To address this, we propose reweighting the feature loss based on the teacher's prediction quality and the importance of each feature.
For the teacher's quality, we weight the distillation loss by a relative depth-error-based score:
\begin{equation}
    \label{eq:quality_indicator}
    \eta_{i} = \frac{z_{i}}{\max\bigl(\lvert z_{i} - \hat{z}_{i}^{\rm T}\rvert,\ \epsilon\bigr)},
\end{equation}
where $z_{i}$ and $\hat{z}_{i}^{\rm T}$ denote ground truth and teacher-predicted depth, respectively, for object $i$.
We set $\epsilon=0.1$ for numerical stability.
Using a relative error mitigates a higher weighting of nearby over distant objects, as nearby objects naturally exhibit smaller absolute errors.
For feature importance, channels with larger absolute weights contribute more to the final prediction.
Therefore, we define the normalized importance per channel $q\in \{1,\ldots,Q\}$ as
\begin{equation}
    \label{eq:feature_relevance_indicator}
    \omega_{q} = \frac{\lvert W_{q}^{\rm T}\rvert}{\sum_{q'=1}^{Q} \lvert W_{q'}^{\rm T}\rvert}.
\end{equation}

We combine these weights with an L1 alignment on instance-depth features.
Let $B(\mathbf{I}) = \{\mathbf{b}^{\text{3D}}_1, \ldots, \mathbf{b}^{\text{3D}}_M\}$ be the ground-truth bounding box set of $\mathbf{I}$ matched to the predictions by CM$_{\text{3D}}$.
We define our quality- and importance-weighted feature-loss between the teacher $\mathcal{F}^{\rm T}_{i}$ and student features $\mathcal{F}^{\rm S}_{i}\in\mathbb{R}^{Q}$ as:
\begin{equation}
    \label{eq:distill_loss}
    \mathcal{L}^{\text{distill}} = \frac{1}{\lvert B(\mathbf{I})\rvert}\;
    \sum_{i=1}^{|B(\mathbf{I})|}\;\sum_{q=1}^{Q}\;
    \omega_q \, \eta_{i} \, \bigl\lvert \mathcal{F}^{\rm T}_{i,q} - \mathcal{F}^{\rm S}_{i,q}\bigr\rvert.
\end{equation}
\subsubsection{Training.}
We adopt offline KD~\cite{knowledge_disstill} with a frozen teacher.
Regardless of the student model size, the teacher is always uses our largest size (\ie, \ourmethodx trained \emph{w/o} A2D2).
We first train the teacher using standard supervision \cite{yolov10}, \ie, classification $\mathcal{L}^{\text{cls}}$, 2D bounding-box $\mathcal{L}^{\text{2D}}$, and 3D bounding-box $\mathcal{L}^{\text{3D}}$ loss, then freeze for distillation.
During distillation, we train a student model, which can be any model size from our family (N,\ S,\ M,\ B,\ or X), using the total loss
\begin{equation}
    \begin{aligned}
    \mathcal{L} &= \mathcal{L}^{\text{cls}} + \mathcal{L}^{\text{2D}} + \mathcal{L}^{\text{3D}} + \mathcal{L}^{\text{distill}},
    \end{aligned}
    \label{eq:total_loss}
\end{equation}
composed of our distillation loss $\mathcal{L}^{\text{distill}}$ and the standard supervised losses. We provide more details on the training and loss functions in the supplement. 

\begin{figure}[t]
\begin{minipage}{.585\linewidth}
    \centering%
    \includegraphics[trim={0 0 0 -0.08cm},clip, width=0.96\linewidth]{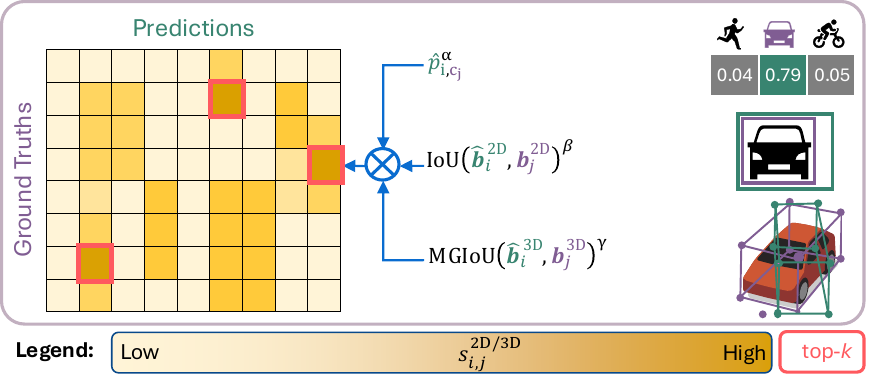}
    \caption{
        \textbf{3D-aware Consistent Matching (CM\textsubscript{3D})} utilizes both 2D and 3D overlaps alongside classification scores to disambiguate prediction-to-ground truth assignments, enabling stable and precise supervision during training and distillation.}%
    \label{fig:tal}
\end{minipage}%
\hfill
\begin{minipage}{.387\linewidth}
    \centering%
    \includegraphics[trim={0 0 0 0.02cm},clip, width=0.96\linewidth]{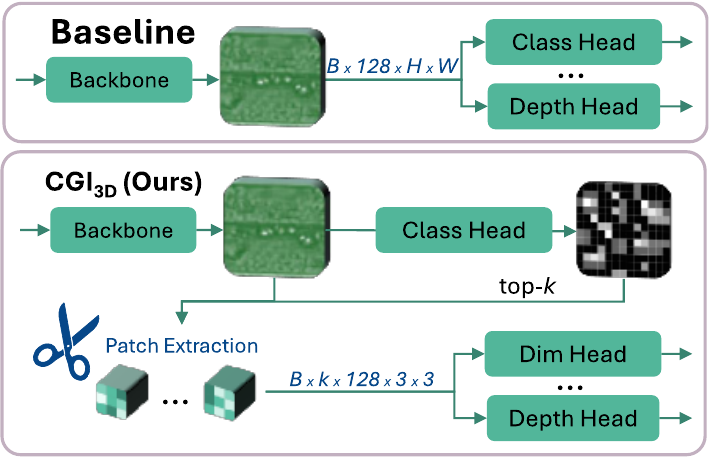}
    \caption{
    \textbf{Confidence-Gated 3D Inference (CGI\textsubscript{3D})} speeds up inference by restricting 2D/3D regression to confident locations (\emph{bottom}), while the baseline processes every location (\emph{top}).}
    
    \label{fig:instance_targeting}
\end{minipage}
\end{figure}

\subsection{3D-aware Consistent Matching (CM$_{\text{3D}}$)}
\label{sec:tal}
Both supervised training and our A2D2 method require reliable assignments between model predictions and ground-truth objects.
Without robust matching, noisy or misaligned pairs can degrade both training stability and final accuracy, especially in challenging scenarios like MixUp, where multiple objects with similar 2D projections coexist.
To address this, we propose 3D-aware Consistent Matching (CM$_{\text{3D}}$), which ranks prediction-ground truth pairs using class confidence and both 2D as well as 3D box agreement (\cf \cref{fig:tal}).

2D matches in our baseline (\baseline \cite{yolov10}) are obtained by ranking all prediction-ground truth pairs using the score $s_{i,j}^{\text{2D}}$, and top-$k$ selection. $s_{i,j}^{\text{2D}}$ is computed by:
\begin{equation}
    s_{i,j}^{\text{2D}} = \hat{p}_{i,c_j}^{\alpha} \,
    \operatorname{IoU}\bigl(\mathbf{\hat{b}}_i^{\text{2D}},\, \mathbf{b}_j^{\text{2D}}\bigr)^{\beta},
\end{equation}
where $i$ indexes predictions, $j$ indexes ground-truth bounding boxes, and $\hat{p}_{i,c_j}$ is the predicted probability for class $c_j$. $\mathbf{\hat{b}}_i^{\text{2D}}$ and $\mathbf{b}_j^{\text{2D}}$ are the predicted and ground-truth 2D bounding boxes. $\alpha\in\mathbb{R}_{+}$ and $\beta\in\mathbb{R}_{+}$ are weighting bounding box overlap and classification.
We extend this to 3D by incorporating the Marginalized Generalized IoU (MGIoU)~\cite{mgiou} between predicted and ground-truth 3D bounding boxes (\cf \cref{fig:tal}):
\begin{equation}
    s_{i,j}^{\text{2D/3D}} = s_{i,j}^{\text{2D}} \,
    \operatorname{MGIoU}\bigl(\mathbf{\hat{b}}_i^{\text{3D}},\, \mathbf{b}_j^{\text{3D}}\bigr)^{\gamma},
\end{equation}
with the predicted $\mathbf{\hat{b}}_i^{\text{3D}}$ and the ground-truth $\mathbf{b}_j^{\text{3D}}$ 3D bounding boxes. $\gamma\in\mathbb{R}_{+}$ is used for weighting.
MGIoU offers a 3D-overlap surrogate that remains informative even without 3D box intersection, which is common early in training or for small objects.
It is computed by projecting the 3D shapes onto a set of unique directional normals and marginalizing the one-dimensional generalized IoU values across these projections.
MGIoU
aggregates location, size, and orientation, and is invariant to object dimensions, unlike corner-based losses~\cite{monodde}. Note, using MGIoU leads to a negligible increase in training time by less than \SI{2}{\%}.

Compared to static anchor-based assignments used in many monocular 3D object detection (M3D) methods~\cite{monocon,cdrone}, our dynamic 2D/3D scoring better disambiguates crowded scenes and mixed content (\eg, under MixUp).
The 2D term stabilizes learning when 3D estimates are crude, while the 3D term helps to separate overlapping predictions as accuracy improves.

\subsection{Confidence-Gated 3D Inference (CGI$_{\text{3D}}$)}
\label{sec:speedup}
M3D detectors typically run their regression heads densely over full feature maps.
As most locations correspond to background, a substantial portion of the computation is wasted.
Therefore, we propose Confidence-Gated 3D Inference (CGI$_{\text{3D}}$) to restrict expensive regression to likely object locations as shown in \cref{fig:instance_targeting}. Specifically, \emph{(1)} after obtaining the neck features, we run the classification head densely across the entire feature map to \emph{(2)} select the top-$k$ locations based on class confidence. We then \emph{(3)} extract $3{\times}3$ local patches centered at these locations and \emph{(4)} apply the 2D and 3D regression heads only to these sparse patches rather than the full map.

During training, we keep the dense head computation for simplicity.
At inference, we move the baseline top-$k$ selection earlier and run the 2D/3D heads only on $3{\times}3$ patches around top-$k$ locations, yielding outputs identical to dense evaluation, because the heads' effective receptive field is exactly $3{\times}3$ (one $3{\times}3$ convolution followed by two $1{\times}1$ convolutions).
This is simpler than Region of Interest (RoI)-Align (typically $7{\times}7$ grids~\cite{mask_rcnn,fastrcnn}) and avoids bilinear interpolation, substantially reducing head-level FLOPs with no accuracy loss.

\section{Experiments}
\label{sec:experiments}

We first compare our approach against lightweight and state-of-the-art (SOTA) M3D methods, followed by domain generalization evaluations. Finally, we analyze individual model components and provide qualitative results.

\subsubsection{Datasets and Metrics.} We perform experiments on the primary M3D benchmarks, KITTI \cite{kitti} and Waymo \cite{waymo}. To evaluate diverse camera perspectives, we conduct experiments on Rope3D \cite{rope3d}. Rope3D includes different cameras and viewpoints, enabling cross-view validation. To showcase domain generalization, we use nuScenes (Front) \cite{nuScenes}. We strictly follow standard protocols and metrics across all datasets. In particular, for KITTI \cite{chen20153d}, we report \APDDDRF\; and \APBEVRF\; at IoU thresholds of \num{0.7} (``Cars'') and \num{0.5} (``Pedestrians''/``Cyclists''). Both metrics are computed for easy, moderate (Mod.), and hard objects. Note that while our main experiments are performed on the KITTI \emph{test} set (requiring a submission to the official test server), analysis and ablations are performed on the \emph{validation} set. Waymo evaluation follows the DEVIANT \cite{deviant} protocol for front-camera images ($\text{AP}_{\text{3D}}$/$\text{APH}_{\text{3D}}$ at \num{0.5}/\num{0.7} IoU). For Rope3D, we use the heterologous split and report \APDDDRF\ and the Rope score \cite{rope3d} for ``Car'' and ``Big Vehicle'' classes. To measure inference speed and efficiency, we report model parameters, runtime, and floating point operations (FLOPs). Both runtime and FLOPs are reported on the same hardware (NVIDIA RTX 8000), if code is available, and for a single image forward pass.

\subsubsection{Implementation Details.} \label{sec:impl_details_main}
We train our student and teacher models using the Adam optimizer \cite{adam} with an initial learning rate of \num{0.001}, a weight decay of \num{0.0005}, and a \num{3}-epoch warmup cosine learning rate schedule \cite{sgdr}. CM$_{\text{3D}}$ hyperparameters are set to $\alpha=0.5$, $\beta=1.0$, and $\gamma=1.0$. All experiments are conducted on a single NVIDIA RTX 8000. The teacher training (\ourmethod w/o A2D2) requires 34 hours; subsequent student training scales with model size, peaking at \SI{60}{h} for the largest model (size X) on KITTI. Despite the focus on optimized inference speed for real-world applications, our training remains resource-efficient, requiring only a \emph{single} GPU. As our direct baseline, we use YOLOv10 by adding 3D detection heads (\baseline). For all implementation details, please refer to the supplement.

\begin{table}[t]
    \fontsize{6pt}{7pt}\selectfont
    \setlength{\tabcolsep}{1.73pt}
    \caption{
        \textbf{Comparison with lightweight M3D methods ($\mathbf{<30\,M}$ parameters)} on the KITTI \cite{kitti} \emph{test} set for the category ``Car'' using \APDDDRFS\ and \APBEVRFS\ (both in \%, $\uparrow$). \emph{Extra} indicates the use of auxiliary training data. \emph{Params} reports no.\ of model parameters in millions. \emph{GFLOPs} measured for single image inference. \emph{Time} is reported in ms for single image inference without TensorRT on an NVIDIA RTX 8000 GPU. Best and second-best results are highlighted in blue \colorindicator{tBlueLight} and light blue \colorindicator{tBlueLightLight}, respectively.
    }
    \begin{tabularx}{\linewidth}{>{\hspace{-\tabcolsep}\raggedright\columncolor{white}[\tabcolsep][\tabcolsep]}XcS[table-format=2.1]S[table-format=4.0]S[table-format=4.1]S[table-format=2.1]S[table-format=2.2]S[table-format=2.2]S[table-format=2.2]S[table-format=2.2]S[table-format=2.2]}
    \toprule
    \multirow{3}{*}{\textbf{Method}} & \multirow{3}{*}{\textbf{Extra}} & {\multirow{3}{*}{\textbf{Params\ $\downarrow$}}} & {\multirow{3}{*}{\textbf{GFLOPs\ $\downarrow$}}} & {\multirow{3}{*}{\textbf{Time\ $\downarrow$}}} & \multicolumn{3}{c}{\textbf{\APDDDRFS\ $\uparrow$}} & \multicolumn{3}{c}{\textbf{\APBEVRFS\ $\uparrow$}}\\
    \cmidrule(lr){6-8} \cmidrule(lr){9-11}
    & & & & & {\textbf{Easy}} & {\textbf{Mod.}} & {\textbf{Hard}} & {\textbf{Easy}} & {\textbf{Mod.}} & {\textbf{Hard}}\\ \midrule

    MonoNERD \cite{mononerd} & LiDAR & \second {6.6} & 4220 & 1380.3 & 22.75 & 17.13 & 15.63 & 31.13 & 23.46 & 20.97\\
    MonoSGC \cite{monosgc} & LiDAR & 23.1 & 173 & 35.0 & 27.01 & 16.77 & 14.61 & 35.78 & 23.27 & 19.92\\
    OccupancyM3D \cite{occupancym3d} & LiDAR & 28.3 & 389 & 213.9 & 25.55 & 17.02 & 14.79 & 35.38 & 24.18 & 21.37\\  
    DPL$_\mathit{FLEX}$ \cite{dpl_flex} & Unlabeled & 21.5 & 152 & 28.9 & 24.19 & 16.67 & 13.83 & 33.16 & 22.12 & 18.74 \\ 

    \midrule

    MonoUNI \cite{monouni} & Geometry & 22.7 & 122 & 23.2 & 24.75 & 16.73 & 13.49 & 33.28 & 23.05 & 16.39\\ 
    MonoCD \cite{monocd} & Geometry & 21.8 & 171 & 28.1 & 25.53 & 16.59 & 14.53 & 33.41 & 22.81 & 19.57 \\
    \midrule

    MonoCon \cite{monocon} & {---} & 19.6 & 115 & 15.5 & 22.50 & 16.46 & 13.95 & 31.12 & 22.10 & 19.00\\
    DDML \cite{depth-discriminative} & {---} & 19.6 & 115 & 15.5 & 23.31 & 16.36 & 13.73 & {---} & {---} & {---}\\
    MonoLSS \cite{monolss} & {---} & 21.5 & 127 & 20.2 & 26.11 & 19.15 & 16.94 & 34.89 & \second 25.95 & 22.59\\
    \midrule

    \ourmethodn (Ours) & --- & \best 3.8 & \best 14 & \best 9.7 & 24.31 & 16.49 & 14.14 & 32.22 & 21.72 & 19.41 \\
    \ourmethods (Ours) & --- & 10.1 & \second 38 & \second 10.2 & 27.28 & 18.87 & 16.37 & 34.86 & 24.17 & 21.32 \\
    \ourmethodm  (Ours) & ---  & 19.7 & 88 & 13.3 & \second 28.08 & \second 19.47 & \second 17.66 & \second 36.21 & 25.46 & \second 22.89 \\
    \ourmethodb (Ours) & --- & 24.9 & 133 & 13.9 & \best 29.10 & \best 20.17 & \best 18.34 & \best 37.65 & \best  26.63 & \best 23.75 \\
    \bottomrule
    \end{tabularx}
    \label{tab:exp_kittI_test_fast}
\end{table}

\subsection{Comparison with Lightweight M3D Models}
In \cref{tab:exp_kittI_test_fast}, we compare our model with existing lightweight M3D models ($<$\SI{30}{M} parameters) in terms of accuracy, FLOPs, model size, and runtime.
Mono\-NeRD~\cite{mononerd} uses fewer parameters than any prior method (\SI{6.6}{M}).
Despite this, our second-smallest model (S) surpasses MonoNeRD in accuracy while being trained LiDAR-free. Notably, \ourmethods has an over \num{100}$\times$ faster inference than MonoNeRD. We attribute MonoNeRD's low speed to its compute‑heavy 3D volume processing, which dominates its inference cost. Among models trained without extra data, MonoLSS is the most accurate existing approach. Still, \ourmethodb outperforms MonoLSS on almost all accuracy metrics, while being \SI{34}{\%} faster in runtime. In terms of FLOPs, our third-largest model (M) requires \SI{25}{\%} fewer FLOPs than the fastest alternatives, DDML~\cite{depth-discriminative} and MonoCon~\cite{monocon}, yet outperforms all existing models on \num{5} out of \num{6} accuracy metrics. Our second-largest model (B) maintains a faster runtime than the fastest existing lightweight baseline while surpassing all prior methods in accuracy.

\subsection{Comparison with SOTA M3D Models}

\begin{table}[t]
    \centering
    \begin{minipage}[t]{0.485\linewidth}
    \fontsize{6pt}{7pt}\selectfont
    \centering
    \setlength{\tabcolsep}{1.32pt}
    \caption{
        \textbf{KITTI~\cite{kitti} test results.} Comparison with state-of-the-art M3D methods on the KITTI \emph{test} set for the category ``Car'' using \APDDDRFS\ (in \%). \emph{Extra} indicates the use of auxiliary train. data.} 
    \begin{tabularx}{\linewidth}{>{\hspace{-\tabcolsep}\raggedright\columncolor{white}[\tabcolsep][\tabcolsep]}XcS[table-format=2.2]S[table-format=2.2]S[table-format=2.2]}
    \toprule
    \raisebox{1.1pt}[0pt][0pt]{\multirow{3}{*}{\textbf{Method}}} & \raisebox{1.1pt}[0pt][0pt]{\multirow{3}{*}{\textbf{Extra}}} & \multicolumn{3}{c}{\textbf{\APDDDRFS\ $\uparrow$}}\\
    \cmidrule(lr){3-5}
    & & \textbf{Easy} & \textbf{Mod.} & \textbf{Hard} \\
    \midrule

    MonoDistill \cite{monodistill} & LiDAR  & 24.31 & 18.47 & 15.76 \\
    ADD \cite{add} & LiDAR & 25.61 & 16.81 & 13.79 \\    HSRDN \cite{hsrdn} & LiDAR & 22.01 & 13.61 & 13.10\\
    MonoFG \cite{monofg} & LiDAR & 24.35 & 16.46 & 13.84\\
    MonoSTL \cite{monostl} & LiDAR & 25.33 & 16.13 & 13.35\\    MonoSG \cite{monosg} & Stereo & 25.77 & 16.70 & 14.22\\
    DK3D \cite{dk3d} & LiDAR & 25.63 & 16.82 & 13.81 \\
    MonoTAKD \cite{monotakd} & LiDAR & 27.91 & 19.43 & 16.51\\

    \midrule

    MoGDE \cite{MoGDE} & Geometry & 27.25 & 17.93 & 15.80\\
    MonoDGP \cite{monodgp} & Geometry & 26.35 & 18.72 & 15.97\\

    \midrule

    Cube R-CNN \cite{brazil2023omni3d} & --- & 23.59 & 15.01 & 12.56\\
    MonoDETR \cite{monodetr} & --- & 25.00 & 16.47 & 16.38\\
    MonoPSTR \cite{monopstr} & --- & 26.15 & 17.01 & 13.70\\
    FD3D \cite{fd3d} & {---} & 25.38 & 17.12 & 14.50\\
    MonoDiff \cite{monodiff} & --- & \second 30.18 & \second 21.02 & \second 18.16\\
    MonoMAE \cite{monomae} & {---} & 25.60 & 18.84 & 16.78\\GATE3D \cite{gate3d} & --- & 26.07 & 18.85 & 16.76\\
    MonoA$^2$ \cite{monoa2} &{---} & 23.24 & 17.55 & 15.26 \\
    \midrule
    \ourmethodx (Ours) & --- & \best 30.76 & \best 21.20 & \best 18.76\\
    \bottomrule
    \end{tabularx}
    \label{tab:exp_kitti_test_sota}

    \end{minipage}%
    \hfill%
    \begin{minipage}[t]{0.485\linewidth}
    \centering
    \fontsize{6pt}{7pt}\selectfont
    \setlength{\tabcolsep}{0.8pt}
    \caption{
        \textbf{Rope3D~\cite{rope3d} results.} Comparison with with state-of-the-art M3D methods on the Rope3D heterlog.\ benchmark. GPF indicates ground-plane-free models. We report \APDDDRFS\ and Rope score in \%. BV denotes ``Big Vehicle'' class.}
    
    \begin{tabularx}{\linewidth}{
        >{\hspace{-\tabcolsep}\raggedright\columncolor{white}[\tabcolsep][\tabcolsep]}XcS[table-format=2.2]S[table-format=2.2]S[table-format=2.2]S[table-format=2.2]
    }
    \toprule
    
    \raisebox{-2.5pt}[0pt][0pt]{\multirow{2}{*}{\textbf{Method}}} & \raisebox{-2.5pt}[0pt][0pt]{\multirow{2}{*}{\textbf{GPF}}} & \multicolumn{2}{c}{\textbf{Car\ $\uparrow$}} & \multicolumn{2}{c}{\textbf{BV\ $\uparrow$}}\\ 
    \cmidrule(lr){3-4} \cmidrule(lr){5-6} 
    & & {\textbf{AP}} & {\textbf{Rope}} & {\textbf{AP}} & {\textbf{Rope}} \\
    
    \midrule
    M3D-RPN~\cite{m3drpn} & \xmark & 11.09 & 28.17 & 3.39 & 21.01 \\
    MonoDLE~\cite{monodle} & \xmark & 12.16 & 28.39 & 3.02 & 19.96 \\
    MonoFlex~\cite{monoflex} & \xmark & 11.24 & 27.79 & \best 13.10 & \best 28.22 \\
    BEVHeight~\cite{bevheight} & \xmark & 5.41 & 23.09 & 1.16 & 18.53 \\
    CoBEV~\cite{cobev} & \xmark & 6.59 & 24.01 & 2.26 & 19.71 \\ \midrule
    
    M3D-RPN~\cite{m3drpn} & \cmark & 6.05 & 23.84 & 2.78 & 20.82 \\
    Kinematic3D \cite{kinematic3d} & \cmark & 5.82 & 23.06 & 1.27 & 18.92 \\
    MonoDLE~\cite{monodle} & \cmark & 3.77 & 21.42 & 2.31 & 19.55 \\
    MonoFlex \cite{monoflex} & \cmark & 10.86  & 27.39 & 0.97 & 18.18 \\
    BEVFormer~\cite{bevformer} & \cmark & 3.87 & 21.84  & 0.84 & 18.42 \\
    BEVDepth~\cite{bevdepth} & \cmark & 0.85 & 19.38 & 0.30 & 17.84 \\
    MonoCon \cite{monocon} & \cmark & 10.71 & 27.55 & 1.61 & 19.25 \\ 
    GroundMix \cite{cdrone} & \cmark & 12.86 & 29.37 & 3.90 & 21.06 \\ \midrule
    
    \ourmethodn (Ours) & \cmark & 9.67 & 25.46 & 1.80 & 19.17\\
    
    \ourmethods (Ours)  & \cmark & 13.33 & 28.59 & 4.16 & 21.25 \\
    
    \ourmethodm (Ours)  & \cmark & 14.31 &  29.62 & 4.95 & 22.14\\
    
    \ourmethodb (Ours)  & \cmark & \second 15.05 & \second 30.13 & 5.40 & 22.41 \\
    
    \ourmethodx (Ours) & \cmark & \best 16.45 & \best 31.34 & \second 8.71 & \second 25.15\\

    \bottomrule
    \end{tabularx}
    \label{tab:exp_rope3d_light}
    \end{minipage}%
\end{table}

\subsubsection{KITTI~\cite{kitti} Test Set.}
In \cref{tab:exp_kitti_test_sota}, we compare our largest model \ourmethodx with SOTA models on the KITTI test set (requiring submission to the test server).
\ourmethod outperforms all existing methods in \APDDDRF, including those that leverage perspective geometric priors (\eg, inverse height-depth consistency) or extra training data.
Furthermore, our model runs \num{3.6}\,$\times$ faster than the previous SOTA MonoDiff~\cite{monodiff} (\cf supplement).

\begin{table}[t]
    \centering
    \begin{minipage}[t]{0.485\linewidth}
    \centering
    \fontsize{6pt}{7pt}\selectfont
    \setlength{\tabcolsep}{2.4pt}
    \caption{\textbf{Waymo~\cite{waymo} results.} Comparison with state-of-the-art M3D methods on the Waymo \emph{validation} set. We report \APDDDF\ \& \APDDDS\ (both in \%, $\uparrow$) for the ``Vehicle'' category. We compare with methods following the DEVIANT~\cite{deviant} protocol.}
    \begin{tabularx}{\linewidth}{
    >{\hspace{-\tabcolsep}\raggedright\columncolor{white}[\tabcolsep][\tabcolsep]}XS[table-format=2.2]S[table-format=2.2]S[table-format=1.2]S[table-format=1.2]
    }
    \toprule
    \multirow{2}{*}{\raisebox{-2pt}[0pt][0pt]{\textbf{Method}}} & \multicolumn{2}{c}{\textbf{\APDDDF\ $\uparrow$}} & \multicolumn{2}{c}{\textbf{\APDDDS\ $\uparrow$}} \\ 
    \cmidrule(lr){2-3} \cmidrule(lr){4-5}
    & {\textbf{L1}} & {\textbf{L2}} & {\textbf{L1}} & {\textbf{L2}} \\
    \midrule
    
    PatchNet \cite{PatchNet} & 2.92 & 2.42 & 0.39 & 0.38\\
    PCT \cite{PCT} & 4.20 & 4.03 & 0.89 & 0.66\\
    GUPNet~\cite{gupnet} & 10.02 & 9.39 & 2.28 & 2.14\\  
    DEVIANT \cite{deviant} & 10.98 & 10.29 & 2.69 & 2.52\\
    MonoJSG \cite{monojsg} & 5.65 & 5.34 & 0.97 & 0.91\\
    MonoCon \cite{monocon} & 10.07 & 9.44 & 2.30 & 2.16\\ 
    Stereoscopic \cite{stereoscopic} & 7.18 & 7.17 & 1.72 & 1.61 \\
    MonoRCNN++ \cite{monorcnn++} & 11.37 & 10.79 & 4.28 & 4.05\\
    MonoUNI \cite{monouni} & 10.98 & 10.38 & 3.20 & 3.04\\
    MonoXiver-GUPNet \cite{monoxiver} & 11.47 & 10.67 & {---} &  {---} \\      
    MonoXiver-DEVIANT \cite{monoxiver} & 11.88 & 11.06 & {---} & {---} \\
    DDML \cite{depth-discriminative} & 10.14 & 9.50 & 2.50 & 2.34 \\
    MonoAux \cite{monoaux} & 9.82 & 9.25 & 3.92 & 3.57 \\
    NF-DVT \cite{nf-dvt} & 11.32 & 11.24 & 2.76 & 2.64\\
    MonoLSS \cite{monolss} & 13.49 & 13.12 & 3.71 & 3.27\\
    SSD-MonoDETR \cite{ssd-monodetr} & 11.83 & 11.34 & \second 4.54 & \second 4.12 \\
    GroundMix \cite{cdrone} & 11.89 & 10.50 & 3.10 & 2.73\\
    MonoCD \cite{monocd} & 11.62 & 11.14 & 3.85 & 3.50 \\
    MonoDGP \cite{monodgp} & 12.36 & 11.71 & 4.28 & 4.00\\
    \midrule
      
    \ourmethodn (Ours) & 12.14 & 10.73 & 2.96 & 2.61\\ 

    \ourmethods (Ours) & 13.24 & 12.08 & 3.53 & 3.11\\ 

    \ourmethodm (Ours) & 14.55 & 12.86 & 3.98 & 3.51\\ 

    \ourmethodb (Ours) & \second 15.04 & \second 13.29 & 4.29 & 3.78\\ 

    \ourmethodx (Ours) & \best 16.46 & \best 14.54 & \best 4.82 &  \best 4.24\\ 
    \bottomrule
    \end{tabularx}
    \label{tab:exp_waymo_light}

    \end{minipage}%
    \hfill%
    \begin{minipage}[t]{0.485\linewidth}
    \centering
    \fontsize{6pt}{7pt}\selectfont
    \setlength{\tabcolsep}{2.15pt}
    \caption{\textbf{Domain generalization results} for nuScenes \cite{nuScenes} to KITTI \cite{kitti} \emph{val.} using \APDDDRFF{} (in \%, $\uparrow$). We compare to domain generalization (DG) and unsupervised adaptation approaches (UDA, in gray) that use target-domain data.}
    \begin{tabularx}{\columnwidth}{>{\hspace{-\tabcolsep}\raggedright\columncolor{white}[\tabcolsep][\tabcolsep]}XcS[table-format=2.2] S[table-format=2.2] S[table-format=2.2]}
        \toprule
        \textbf{Method} & \textbf{Type} & \textbf{Easy} & \textbf{Mod.} & \textbf{Hard} \\ 
        \midrule  
        \color{gray} STMono3D~\cite{stmono3d} & \color{gray}UDA & \color{gray}29.01 & \color{gray}19.88 & \color{gray}17.17 \\
        \color{gray} MonoCT~\cite{monoct} & \color{gray}UDA & \color{gray} 42.80 & \color{gray}32.24 & \color{gray}27.36 \\ 
        \midrule
        DGMono3D~\cite{dgmono3d} & DG & 28.77 & 24.82 & 23.67 \\ 
        MonoGDG~\cite{monogdg} & DG & \second 33.48 & \second 27.14 & \second 26.37 \\ 
        LeAD-M3D B (Ours) & DG & \best 45.50 & \best 32.14 & \best 28.56 \\ 
        \bottomrule
    \end{tabularx}
    \label{tab:rebuttal_domgen}
        \vphantom{Bayer 04}\\[1.0pt]
    \fontsize{6pt}{7pt}\selectfont
    \centering
    \caption{\textbf{CM$_{\text{3D}}$\;\&\;A2D2 analysis} on KITTI \emph{val.} using \APDDDRFS\,(in \%,\,$\uparrow$, ``Car'') \& median depth error (MDE, in \%, $\downarrow$). Configuration \num{1} is \baselineb (our baseline); config. \num{5} is \ourmethodb.}
    \setlength{\tabcolsep}{1.3pt} 
    \begin{tabularx}{\linewidth}{
        >{\hspace{-\tabcolsep}\raggedright\columncolor{white}[\tabcolsep][\tabcolsep]}Xccc|S[table-format=2.2]S[table-format=2.2]S[table-format=2.2]|S[table-format=2.0]
        }
    \toprule
     &  & \multicolumn{2}{c|}{\textbf{CM$_{\text{3D}}$}} & \multicolumn{3}{c|}{\textbf{\APDDDRFS}\ $\uparrow$} & {\raisebox{-3.0pt}[0pt][0pt]{\multirow[c]{2}{23pt}{\textbf{MDE\,$\downarrow$ in cm}}}} \\
    \cmidrule{3-4} \cmidrule{5-7} 
    \multirow{-2}{*}{\raisebox{3.5pt}[0pt][0pt]{\textbf{Config.}}} & \multirow{-2}{*}{\raisebox{3.5pt}[0pt][0pt]{\textbf{A2D2}}} & \textbf{2D} & \textbf{3D} & {\textbf{Easy}} & {\textbf{Mod.}} & {\textbf{Hard}} &  \\
    \midrule
    1 &        & \cmark &        & 25.68 & 19.60 & 17.47 & 61 \\
    2 &        & \cmark & \cmark & 26.26 & 20.43 & 17.71 & 60 \\
    \midrule
    3 & \cmark & \cmark &        & \second 27.12 & \second 21.81 & \second 19.34 & \second 57 \\
    4 & \cmark &        & \cmark & 26.34 & 21.49 & 19.12 & \second 57 \\
    5 (full) & \cmark & \cmark & \cmark & \best 28.44 & \best 22.65 & \best 19.87 & \best 56 \\
    \bottomrule
    \end{tabularx}
    \label{tab:main_ablation}

    \end{minipage}%
\end{table}

\subsubsection{Rope3D~\cite{rope3d} Validation Set.}
In \cref{tab:exp_rope3d_light}, we provide results on Rope3D \cite{rope3d} (traffic view), evaluating accuracy for views different from car mounted cameras.
\ourmethodx scores the best result for the ``Car'' category.
For the rare category ``Big Vehicle'', it performs second-best to MonoFlex~\cite{monoflex}, which requires ground-plane inputs, while \ourmethod does not.
In a fair comparison (\ie, MonoFlex w/o ground-plane input), \ourmethod significantly outperforms MonoFlex.

\subsubsection{Waymo~\cite{waymo} Validation Set.}
Similarly, our largest model (X) achieves the best results on the Waymo validation set~\cite{waymo} (\cf \cref{tab:exp_waymo_light}) without perspective geometric priors or extra training data.
In \APDDDF\ Level~1, we surpass the previous best model~\cite{monolss} by \SI{2.97} AP.
Even our second-largest model (B) attains a higher \APDDDF\ than all existing methods. 

\subsubsection{Domain Generalization.} In \cref{tab:rebuttal_domgen}, we evaluate the robustness of \ourmethodb to domain shifts. In particular, we analyze domain generalization by training on nuScenes~\cite{nuScenes} and evaluating on the KITTI~\cite{kitti} validation set. \ourmethodb outperforms all domain generalization (DG) approaches, even MonoGDG~\cite{monogdg}, a specialized approach for DG, requiring alignments for focal length, distortion, and camera orientation. In contrast, \ourmethodb generalizes using only virtual depth~\cite{brazil2023omni3d} without specific components designed for DG. Notably, \ourmethod remains competitive with unsupervised domain adaptation (UDA) methods that leverage target data for adaptation. This demonstrates that \ourmethod learns highly generalizable representations effective in diverse sensor setups.

\subsection{Analyzing \ourmethod}
\label{sec:analyzing_method}
We conduct multiple analyses and ablations to demonstrate the effectiveness of our key contributions. We report results using the KITTI~\cite{kitti} validation and use \ourmethodb\space if not noted differently. Further analyses are in the supplement.

\subsubsection{Main Analyses.}
\Cref{tab:main_ablation} analyzes both A2D2 and CM$_{\text{3D}}$ \wrt the detection accuracy. Starting from our baseline (\baselineb), adding 3D cues in CM$_{\text{3D}}$ (config. 2) improves the AP (Moderate) by \SI{0.83}{\%} and the depth error by \SI{1}{cm}. Adding distillation using A2D2 (config. 5) further improves the accuracy by \SI{2.22}{\%} in AP (Mod.) and significantly reduces the depth error by \SI{4}{cm} \wrt configuration 2. These results demonstrate that the A2D2, our core contribution, is the key driver to strong detection and depth accuracy of \ourmethod. Still, both A2D2 and CM$_{\text{3D}}$ improve the accuracy independently of each other (\cf config. 2, 3 \& 4 in \cref{tab:main_ablation}).

\subsubsection{A2D2 Ablation.} In \cref{tab:ablation_distillation}, we ablate individual components of our A2D2 approach in detail by removing or adding specific components from our full configuration (\ourmethodb). Distilling backbone features instead of depth features yields a reduced AP (Mod.) by \SI{0.35}{\%}, confirming that depth features are a more effective KD target than generic backbone features. Removing quality and importance indicators reduces AP (Mod.) by \SI{0.37}{\%} and \SI{0.61}{\%}, respectively. Performing online, \ie, distilling during initial supervised training, instead of offline distillation, reduces AP (Mod.) by \SI{0.85}{\%}. This is likely because the strongest distillation targets emerge only late in training, coinciding with the learning rate decay necessary for depth convergence. Finally, feeding the teacher the same MixUp image as the student, \ie, removing asymmetry from augmentations, yields a drop in AP (Mod.) of \SI{1.08}{\%}, demonstrating that augmentation asymmetry/denoising is beneficial for effective distillation.

\begin{table}[t]
    \centering
    \begin{minipage}[t]{0.485\linewidth}
    \centering
    \fontsize{6pt}{7pt}\selectfont
    \caption{\textbf{Detailed A2D2 ablation} on KITTI \cite{kitti} \emph{validation} using car category \APDDDRFS\ (in \%) and model size B. We ablate the key components of A2D2 and report results without A2D2 for reference.}
    \setlength{\tabcolsep}{1.4pt}
    \begin{tabularx}{\linewidth}{
    >{\hspace{-\tabcolsep}\raggedright\columncolor{white}[\tabcolsep][\tabcolsep]}X
    S[table-format=2.2]
    S[table-format=2.2]
    S[table-format=2.2]
    }
    \toprule
    
    \textbf{Method} & {\textbf{Easy\,$\uparrow$}} & {\textbf{Mod.\,$\uparrow$}} & {\textbf{Hard\,$\uparrow$}} \\ \midrule
    \ourmethodb (Ours) & \best \bfseries 28.44 & \best \bfseries 22.65 & \best \bfseries 19.87 \\
    \hspace{0.25em} w/ backbone features & 27.56 & \second 22.30 & \second 19.82 \\
    \hspace{0.25em} w/o quality indicator & \second 27.79 & 22.28 &  19.63 \\ 
    \hspace{0.25em} w/o importance indicator & 27.33 & 22.04 & 19.37 \\
    \hspace{0.25em} w/ online self-distillation & 27.45 & 21.80 & 19.57 \\
    \hspace{0.25em} w/o clean images & 27.19 & 21.57 & 19.27 \\ \midrule
    Ours w/o A2D2 & 26.26 & 20.43 & 17.71 \\ 
    
    \bottomrule 
    \label{tab:ablation_distillation}
    \end{tabularx}

        \vphantom{.}\\
    \centering
    \fontsize{6pt}{7pt}\selectfont
    \setlength{\tabcolsep}{5.9pt}
    \caption{
        \textbf{A2D2 augmentation analysis.} We analyze different augmentation strategies for distillation on the KITTI~\cite{kitti} \emph{validation} set using our B model and report Car Moderate \APDDDRFS\ in \%. For reference, we also report results without distillation (w/o A2D2).
    }
    \begin{tabularx}{\linewidth}{
    >{\hspace{-\tabcolsep}\raggedright\columncolor{white}[\tabcolsep][\tabcolsep]}XS[table-format=2.2]}
    \toprule
    \textbf{Augmentation strategy} & {\textbf{Car \APDDDRFS\ $\uparrow$}} \\
    \midrule
    \color{gray} \ourmethod w/o A2D2 & \color{gray} 20.43 \\
    \midrule
    RandAugment\cite{randaugment} & 21.10 \\
    CutMix~\cite{cutmix} & \second 22.28 \\
    MixUp~\cite{mixup} & \best 22.65 \\
    \bottomrule 
    \end{tabularx}
    \label{tab:ablation_augmentation}

    \end{minipage}%
    \hfill%
    \begin{minipage}[t]{0.485\linewidth}
    \centering
    \fontsize{6pt}{7pt}\selectfont
    \setlength{\tabcolsep}{2.15pt}
    \caption{\textbf{Runtime analysis of CGI$_{\text{3D}}$} on KITTI \cite{kitti} \emph{validation} using car category \APDDDRFS\ (in \%), model size N, and single image inference. Ch.\ denotes the number of regression head channels. Time is reported in ms.}
    \begin{tabularx}{\linewidth}{
    >{\hspace{-\tabcolsep}\raggedright\columncolor{white}[\tabcolsep][\tabcolsep]}cS[table-format=3.0]|S[table-format=2.2]|S[table-format=2.1]S[table-format=2.1]}
    \toprule
    {\textbf{CGI$_{\text{3D}}$}} & {\textbf{Ch.}} & {\textbf{\APDDDRFS\ $\uparrow$}} & {\textbf{GFLOPs\ $\downarrow$}} & 
    {\textbf{Time\ $\downarrow$}} \\ 
    \midrule
    & 128 & 18.33 & 81.0 & {16.2} \\
    \cmark & 128 & 18.33 & \second 19.5 & \second 10.9 \\
    \midrule
    & 64 & 18.38 & 45.8 & 15.5 \\
    \cmark & 64 & 18.38 & \best 14.2 & \best 8.2 \\
    \bottomrule 
    \end{tabularx}
    \label{tab:ablationd_runtime}
        \vphantom{.}\\
        \vphantom{.}\\[0.6pt]
        \vphantom{.}\\
    \centering
    \fontsize{6pt}{7pt}\selectfont
    \setlength{\tabcolsep}{3.0pt}
    \caption{
        \textbf{Analysis of augmentation difficulty.} We measure teacher (\ourmethodx w/o A2D2) depth error (in cm, $\downarrow$) on augmented KITTI~\cite{kitti} \emph{val.} sets. MixUp induces the highest error, providing an effective asymmetry for distillation.
    }
    \begin{tabularx}{\linewidth}{>{\hspace{-\tabcolsep}\raggedright\columncolor{white}[\tabcolsep][\tabcolsep]}XS[table-format=3.0]S[table-format=2.0]}
    \toprule
    \multirow{2}{*}{\textbf{Augmentation strategy}} & \multicolumn{2}{c}{\textbf{Depth Error\,(cm)\,$\downarrow$}} \\
    \cmidrule(lr){2-3}
    & {\;\;\;\;Mean\;\;\;\;} & {Median} \\
    \midrule
    {\color{gray} No aug. (KITTI val.)} & \color{gray}118 & \color{gray}59 \\
    \midrule
    RandAugment~\cite{randaugment} & 129 & 62 \\
    CutMix~\cite{cutmix} & \second 137 & \second 66 \\
    MixUp~\cite{mixup} & \best 167 & \best 75 \\
    \bottomrule 
    \end{tabularx}
    \label{tab:ablation_augmentation_validation_set}
    \end{minipage}%
\end{table}

\subsubsection{Runtime Analysis.} In \cref{tab:ablationd_runtime}, we analyze the effect of CGI$_{\text{3D}}$ on the inference efficiency and accuracy of \ourmethodn. In particular, we analyze our two core modifications of the baseline (YOLOv10-M3D): First, applying CGI$_{\text{3D}}$ during inference, and second, reducing the head channels from \num{128} to \num{64}. Applying CGI$_{\text{3D}}$ significantly reduces the runtime by two-thirds and FLOPs by about \SI{75}{\%}, while retaining the accuracy. Reducing head channels further improves the runtime by \SI{0.7}{ms} and FLOPs by ca.\ \SI{40}{\%}. The  AP remains about the same. Together, our modifications significantly improve inference efficiency without sacrificing accuracy. Please see the supplement for additional results and insights.

\subsubsection{Augmentation Strategies for A2D2.} A2D2 enables the student to improve the accuracy by artificially complicating the student's task. A key component during this distillation is the data augmentation strategy, which introduces asymmetry and requires denoising. To analyze the effectiveness of MixUp in A2D2, we report results with different augmentation strategies in \cref{tab:ablation_augmentation}. In particular, we perform distillation with MixUp~\cite{mixup}, CutMix~\cite{cutmix}, and RandAugment~\cite{randaugment} (omitting geometric transforms). MixUp outperforms both CutMix and RandAugment. Still, all augmentation strategies improve over performing no distillation, demonstrating the general effectiveness of A2D2.

To further analyze why A2D2 is effective, we generate augmented versions of the KITTI validation set. In particular, we utilize different augmentation strategies (\ie, MixUp, CutMix, and RandAugment) to generate augmented validation sets. On these augmented validation sets, we report the depth error using \ourmethodx \textit{w/o A2D2} in \cref{tab:ablation_augmentation_validation_set}. All augmentation strategies lead to an increased depth error over the error on the clean validation set (no aug.), while MixUp leads to the largest depth error. This shows that augmentations effectively introduce an information asymmetry, \ie, complicating the student's objective. As demonstrated by the results in \cref{tab:ablation_augmentation,tab:ablation_distillation}, the student benefits from the large asymmetry introduced by MixUp, leading to increased detection accuracy and depth reasoning.

\newlength{\imgH}
\setlength{\imgH}{0.77cm}

\newlength{\bevH}
\setlength{\bevH}{2\imgH}

\begin{figure*}[t]
    \centering

    \begin{tabular}{ccc@{\hskip 0.24cm}ccc}
         &  & \sffamily BEV\;\;\;\;\;\;\;\;\, &  &  & \sffamily BEV\;\;\;\;\;\;\;\;\, \\
        \raisebox{0.9\imgH}[0pt][0pt]{\rotatebox[origin=c]{90}{\sffamily Ours}} & \begin{tikzpicture}[spy using outlines={red, magnification=2.75, minimum width=1.656\imgH, minimum height=0.925\imgH, fill=white, connect spies}] 
            \node[inner sep=0pt, outer sep=0pt] at (0, 0) {\includegraphics[height=\imgH]{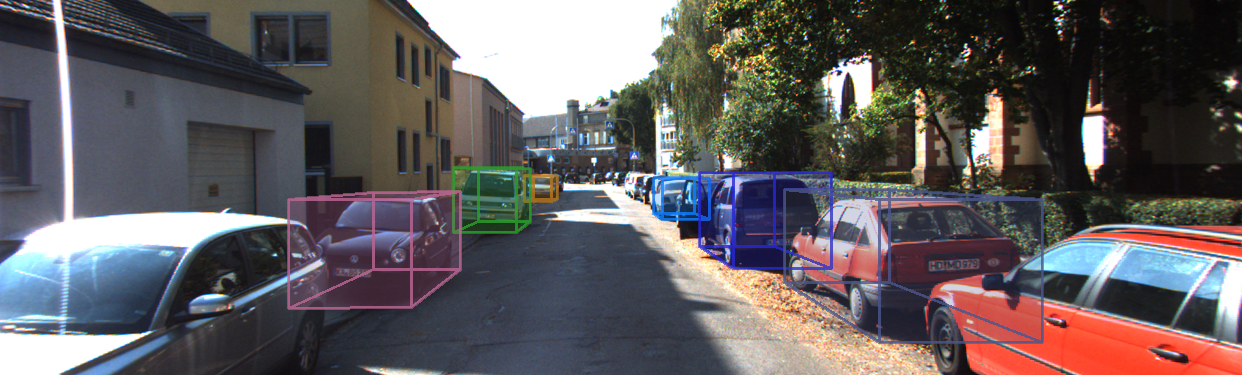}};
            \spy on (0.25, -0.075) in node [draw opacity=1.0, black, fill=white, anchor=center, line width=0.2pt, inner sep=0pt, outer sep=0pt] at (0.828\imgH, 1.05\imgH);
            \spy on (-0.25, 0.0) in node [draw opacity=1.0, black, fill=white, anchor=center, line width=0.2pt, inner sep=0pt, outer sep=0pt] at (-0.828\imgH, 1.05\imgH);
        \end{tikzpicture} & \begin{tikzpicture}[spy using outlines={black, magnification=2.75, minimum width=0.5\bevH, minimum height=0.5\bevH, fill=white, connect spies}]
            \node[inner sep=0pt, outer sep=0pt] at (0, 0) {\includegraphics[height=\bevH, trim={5cm 0.3cm 5cm 0.45cm}, clip, frame=0.2pt]{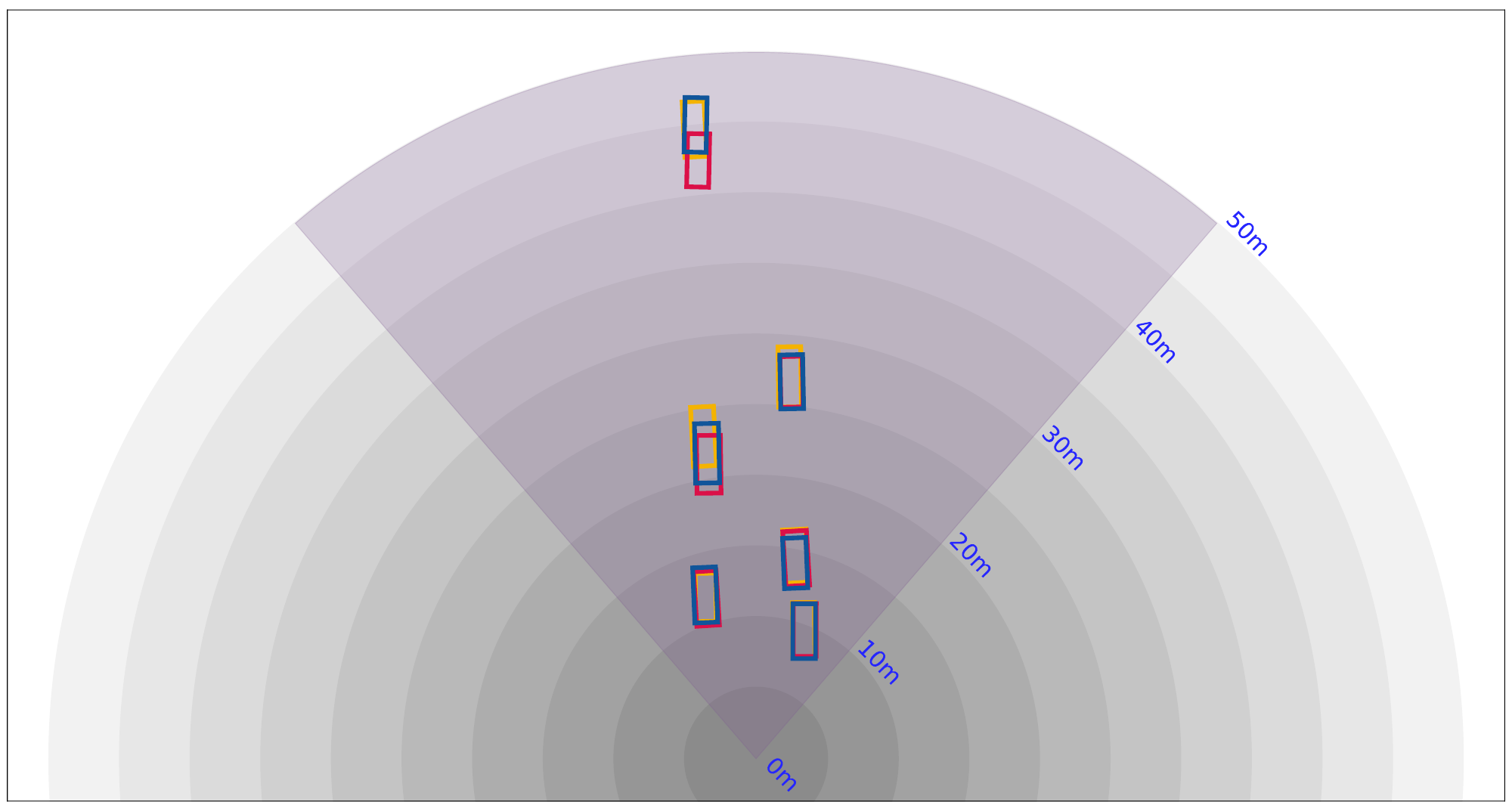}};\spy on (-0.1, 0.525) in node [draw opacity=1.0, black, fill=white, anchor=center, line width=0.2pt, inner sep=0pt, outer sep=0pt] at (0.975\bevH, 0.25\bevH);
            \spy on (-0.025, -0.035) in node [draw opacity=1.0, black, fill=white, anchor=center, line width=0.2pt, inner sep=0pt, outer sep=0pt] at (0.975\bevH, -0.25\bevH);
        \end{tikzpicture} & \raisebox{0.9\imgH}[0pt][0pt]{\rotatebox[origin=c]{90}{\sffamily Ours}} & \begin{tikzpicture}[spy using outlines={red, magnification=2.75, minimum width=1.656\imgH, minimum height=0.925\imgH, fill=white, connect spies}] 
            \node[inner sep=0pt, outer sep=0pt] at (0, 0) {\includegraphics[height=\imgH]{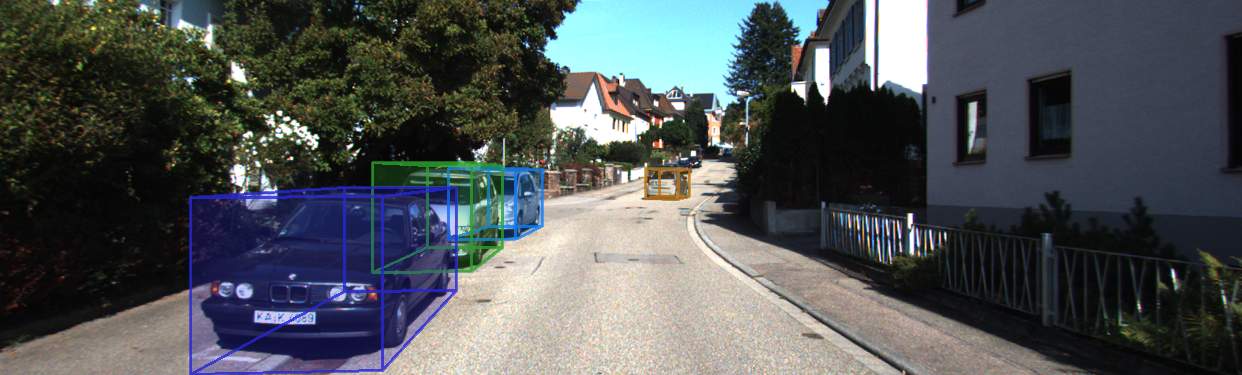}};
            \spy[magnification=5.0] on (0.1, 0.0) in node [draw opacity=1.0, black, fill=white, anchor=center, line width=0.2pt, inner sep=0pt, outer sep=0pt] at (0.828\imgH, 1.05\imgH);
            \spy on (-0.3, -0.05) in node [draw opacity=1.0, black, fill=white, anchor=center, line width=0.2pt, inner sep=0pt, outer sep=0pt] at (-0.828\imgH, 1.05\imgH);
        \end{tikzpicture} & \begin{tikzpicture}[spy using outlines={black, magnification=2.75, minimum width=0.5\bevH, minimum height=0.5\bevH, fill=white, connect spies}]
            \node[inner sep=0pt, outer sep=0pt] at (0, 0) {\includegraphics[height=\bevH, trim={5cm 0.3cm 5cm 0.45cm}, clip, frame=0.2pt]{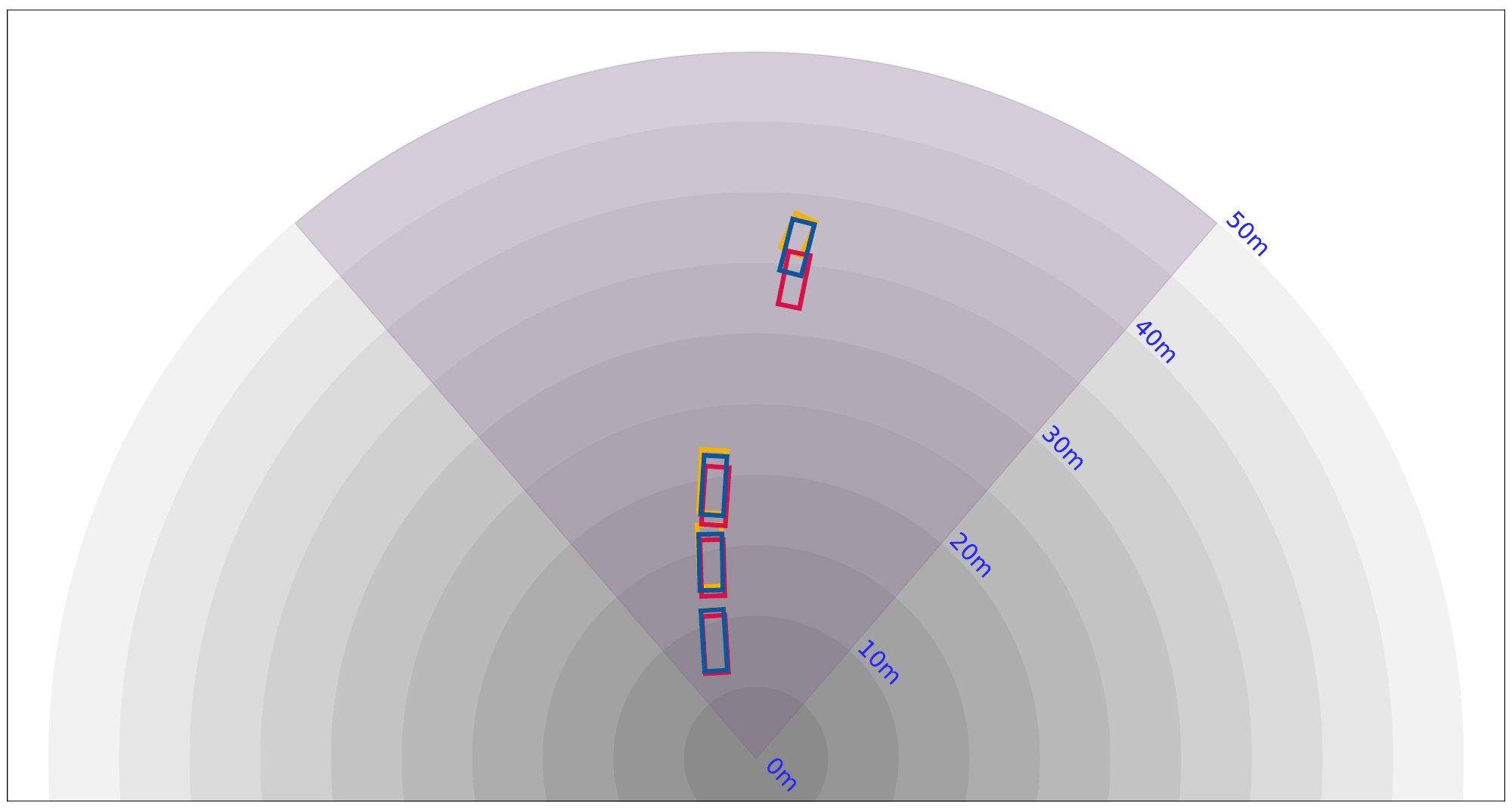}};\spy on (0.1, 0.3) in node [draw opacity=1.0, black, fill=white, anchor=center, line width=0.2pt, inner sep=0pt, outer sep=0pt] at (0.975\bevH, 0.25\bevH);
            \spy on (-0.1, -0.2) in node [draw opacity=1.0, black, fill=white, anchor=center, line width=0.2pt, inner sep=0pt, outer sep=0pt] at (0.975\bevH, -0.25\bevH);
        \end{tikzpicture} \\[-1pt]
        \raisebox{0.9\imgH}[0pt][0pt]{\rotatebox[origin=c]{90}{\sffamily Ours}} & \begin{tikzpicture}[spy using outlines={red, magnification=7.5, minimum width=1.656\imgH, minimum height=0.925\imgH, fill=white, connect spies}] 
            \node[inner sep=0pt, outer sep=0pt] at (0, 0) {\includegraphics[height=\imgH]{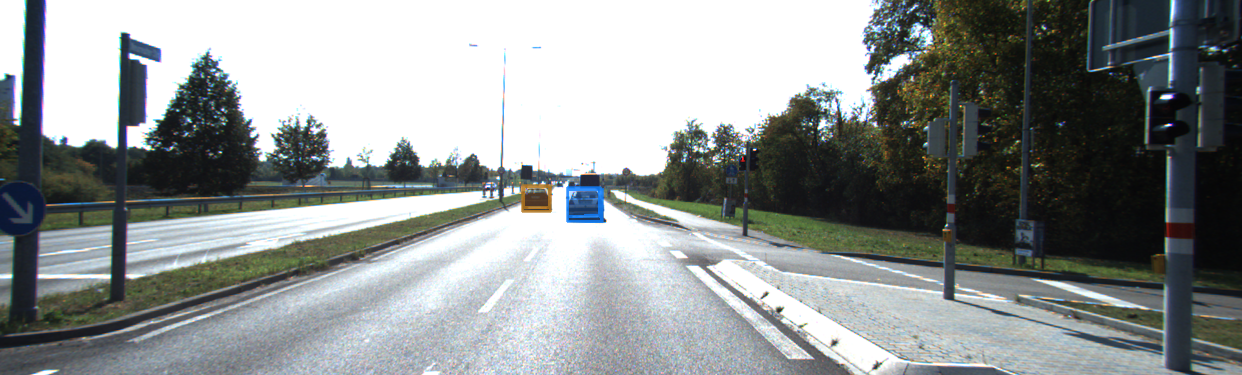}};
            \spy on (-0.035, -0.03) in node [draw opacity=1.0, black, fill=white, anchor=center, line width=0.2pt, inner sep=0pt, outer sep=0pt] at (0.828\imgH, 1.05\imgH);
            \spy on (-0.215, -0.025) in node [draw opacity=1.0, black, fill=white, anchor=center, line width=0.2pt, inner sep=0pt, outer sep=0pt] at (-0.828\imgH, 1.05\imgH);
        \end{tikzpicture} & \begin{tikzpicture}[spy using outlines={black, magnification=2.75, minimum width=0.5\bevH, minimum height=0.5\bevH, fill=white, connect spies}]
            \node[inner sep=0pt, outer sep=0pt] at (0, 0) {\includegraphics[height=\bevH, trim={5cm 0.3cm 5cm 0.45cm}, clip, frame=0.2pt]{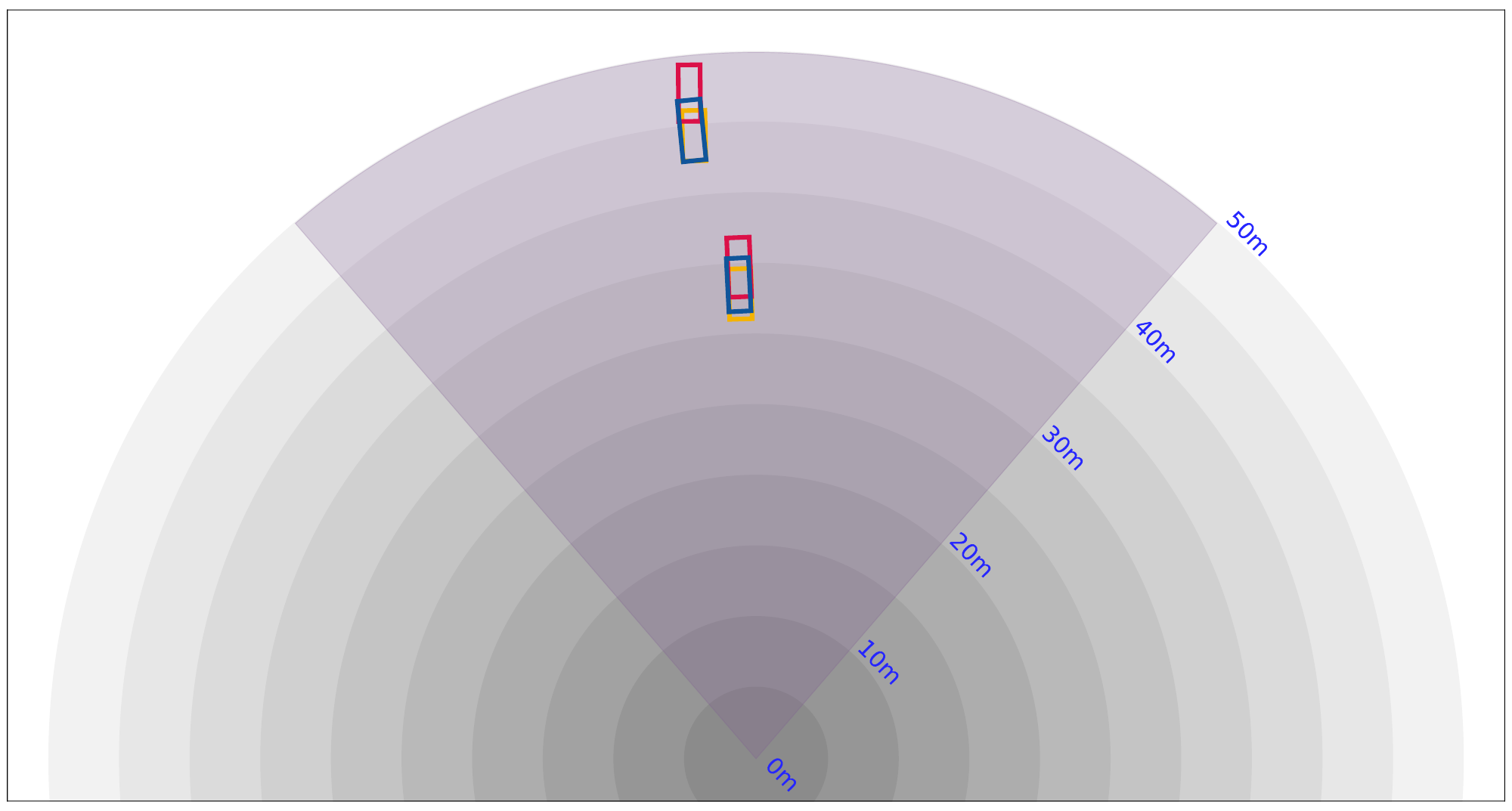}};\spy on (-0.1, 0.6) in node [draw opacity=1.0, black, fill=white, anchor=center, line width=0.2pt, inner sep=0pt, outer sep=0pt] at (0.975\bevH, 0.25\bevH);
            \spy on (-0.025, 0.275) in node [draw opacity=1.0, black, fill=white, anchor=center, line width=0.2pt, inner sep=0pt, outer sep=0pt] at (0.975\bevH, -0.25\bevH);
        \end{tikzpicture} & \raisebox{0.9\imgH}[0pt][0pt]{\rotatebox[origin=c]{90}{\sffamily Ours}} & \begin{tikzpicture}[spy using outlines={red, magnification=2.75, minimum width=1.656\imgH, minimum height=0.925\imgH, fill=white, connect spies}] 
            \node[inner sep=0pt, outer sep=0pt] at (0, 0) {\includegraphics[height=\imgH]{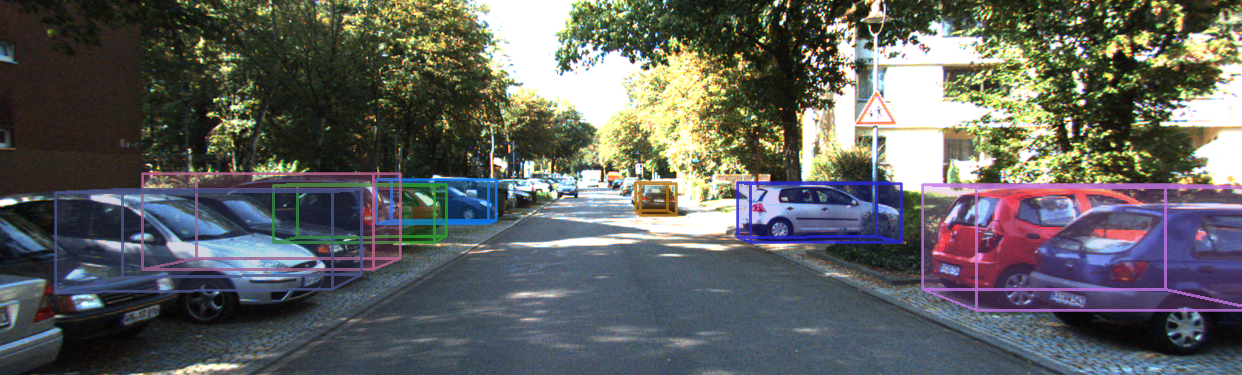}};
            \spy[magnification=5.0] on (0.1, 0.0) in node [draw opacity=1.0, black, fill=white, anchor=center, line width=0.2pt, inner sep=0pt, outer sep=0pt] at (0.828\imgH, 1.05\imgH);
            \spy[magnification=4.0] on (-0.4, 0.0) in node [draw opacity=1.0, black, fill=white, anchor=center, line width=0.2pt, inner sep=0pt, outer sep=0pt] at (-0.828\imgH, 1.05\imgH);
        \end{tikzpicture} & \begin{tikzpicture}[spy using outlines={black, magnification=2.75, minimum width=0.5\bevH, minimum height=0.5\bevH, fill=white, connect spies}]
            \node[inner sep=0pt, outer sep=0pt] at (0, 0) {\includegraphics[height=\bevH, trim={5cm 0.3cm 5cm 0.45cm}, clip, frame=0.2pt]{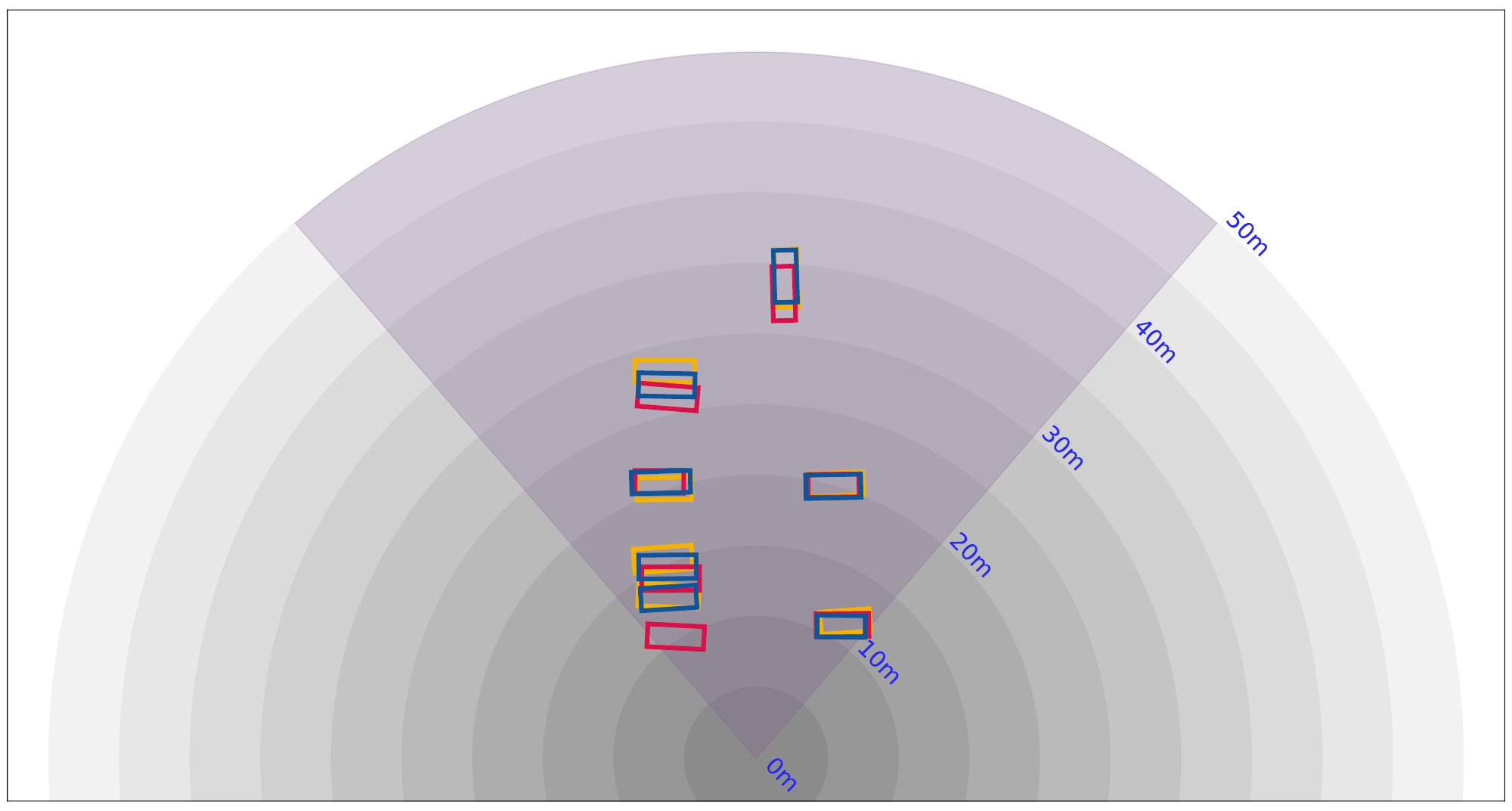}};\spy on (0.1, 0.25) in node [draw opacity=1.0, black, fill=white, anchor=center, line width=0.2pt, inner sep=0pt, outer sep=0pt] at (0.975\bevH, 0.25\bevH);
            \spy on (-0.2, 0.035) in node [draw opacity=1.0, black, fill=white, anchor=center, line width=0.2pt, inner sep=0pt, outer sep=0pt] at (0.975\bevH, -0.25\bevH);
        \end{tikzpicture} \\
    \end{tabular}

    \caption{
        \textbf{Qualitative results} on the KITTI~\cite{kitti} \emph{val.} set. We visualize \ourmethodx's predictions in 2D \emph{(left)} and compare \ourmethodx to our baseline (\baselinex) in the bird's eye view (BEV) \emph{(right)}.
        \ourmethodx achieves more accurate depth estimates than our baseline (\baselinex). We highlight improved detections.
        Best viewed in color; zoom in for details.
        {BEV color coding}: Ground truth \colorindicator{qualGTcr}, \baselinex\ \colorindicator{qualBLcr}, \ourmethodx\ \colorindicator{qualOURcr}, and field of view \colorindicator{qualFOV}.}
        
    \label{fig:kitti_qualitative}
\end{figure*}

\subsection{Qualitative Results}
In \cref{fig:kitti_qualitative}, we show qualitative results on the KITTI~\cite{kitti} validation set. Compared to the baseline, \ourmethod's detections entail noticeably more accurate depth estimates. The improvements are particularly pronounced for moderate- and far-range objects, where estimating depth is especially challenging. For more qualitative results, please refer to the supplement.

\begin{figure}[t]
    \centering
    \tikzset{every picture/.style={/utils/exec={\sffamily\scriptsize}}}
    \begin{tikzpicture}[spy using outlines={white, magnification=2.0, minimum width=1.19cm, minimum height=1.19cm, fill=white, connect spies}]
        \node[inner sep=0pt, outer sep=0pt] at (-0.3333333\linewidth, 0) {\includegraphics[width=0.3333333333\linewidth]{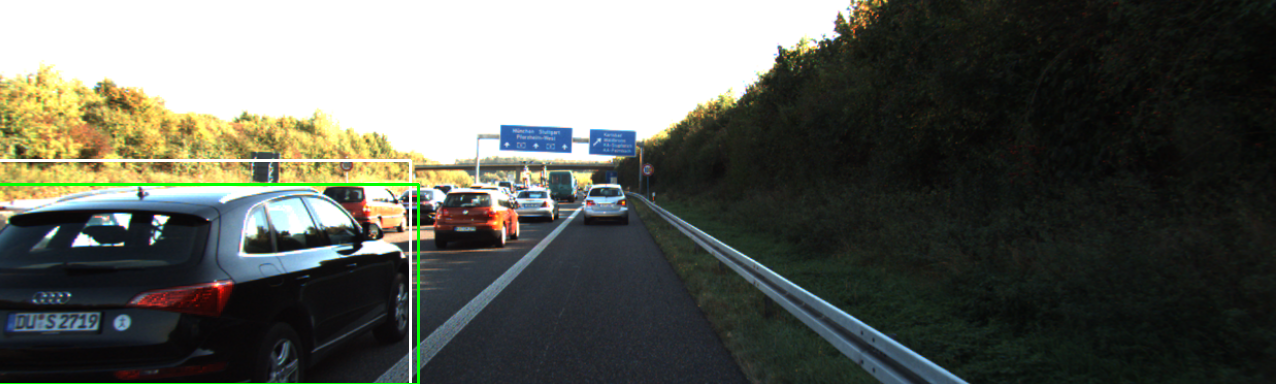}};
        \node[inner sep=0pt, outer sep=0pt] at (0.16666666\linewidth, 0) {\includegraphics[width=0.6666666666\linewidth]{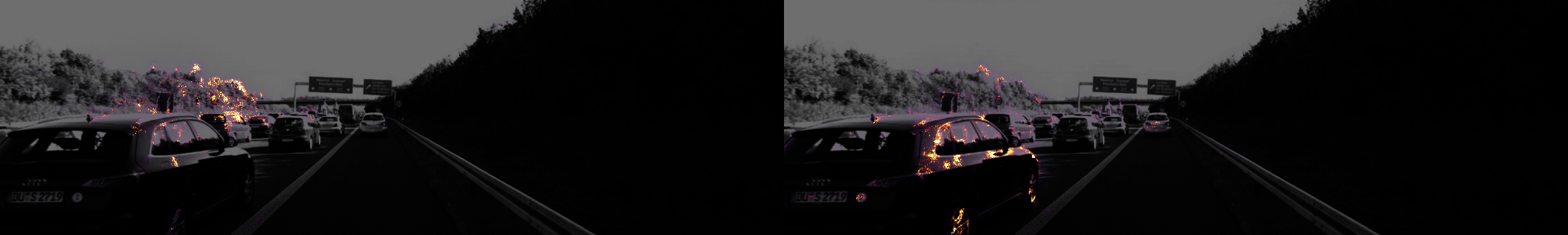}};
        \node[inner sep=0pt, outer sep=0pt, white] at (-0.065\linewidth, 0.46) {w/o denoise \& distill.};
        \node[inner sep=0pt, outer sep=0pt, white] at (0.2683333333\linewidth, 0.46) {w/ denoise \& distill.\hphantom{o}};
        \spy on (-0.9, 0.0) in node [draw opacity=1.0, white, fill=white, anchor=center, line width=0.5pt, inner sep=0pt, outer sep=0pt] at (1.425cm, 0.0cm);
        \spy on (3.0, -0.15) in node [draw opacity=1.0, white, fill=white, anchor=center, line width=0.5pt, inner sep=0pt, outer sep=0pt] at (5.485cm, 0.0cm);
    \end{tikzpicture}
    \caption{\textbf{Saliency analysis.} Integrated gradients~\cite{intgrad} \wrt object depth (left bottom car) on KITTI \emph{val}. Without denoising \& distillation \emph{(middle)}, saliency attends to noisy background regions. With both \emph{(right)}, saliency concentrates on the object.}
    \label{fig:ig_main}
\end{figure}

\subsection{Understanding Denoising and Distillation.} Our approach combines denoising and distillation (\cf \cref{fig:method_overview}) to obtain accurate 3D detections. However, both components entail different mechanistic properties. MixUp-based denoising \emph{increases detection difficulty} (\cf \cref{tab:ablation_augmentation_validation_set}), thereby using annotations more effectively. Denoising also acts as a \emph{regularizer}, forcing the student to rely on robust and MixUp-invariant cues, \eg, object structure, rather than noisy background structure, as demonstrated qualitatively in \cref{fig:ig_main}. Depth feature distillation by A2D2 acts as an effective \emph{guidance} signal for denoising (\cf \cref{tab:ablation_distillation}). Together, denoising and distillation enable the student to improve 3D reasoning and detection accuracy.

\section{Conclusion}
\label{sec:conclusion}
We presented a real-time monocular 3D detection framework that employs asymmetric augmentation denoising distillation, 3D-aware matching, and confidence-gated inference. Our \ourmethod model family establishes a new accuracy-effi\-ciency Pareto frontier for M3D, without using LiDAR supervision, stereo, or perspective geometric assumptions/inputs (\eg, height-depth consistency assumption or ground-plane input). \ourmethod achieves robust accuracy across viewpoints and datasets while also demonstrating strong domain generalization. While we focus on the supervised setting, our distillation could extend to semi-supervised settings, providing an avenue for scaling M3D.

{
\footnotesize
\paragraph{Acknowledgments.}
This work is a result of the joint research project STADT:up and was funded by the German Federal Ministry for Economic Affairs and Climate Action (BMWK), based on a decision of the German Bundestag. The author is solely responsible for the content of this publication. This work was also supported by the ERC Advanced Grant SIMULACRON, the Georg Nemetschek Institute project AI4TWINNING \& the DFG project 4D-YouTube CR 250/26-1. Funding was also received from the Deutsche Forschungsgemeinschaft (German Research Foundation, DFG) under Germany's Excellence Strategy (EXC-3057/1 ``Reasonable Artificial Intelligence'', Project No.\,533677015) and the LOEWE initiative (Hesse, Germany) within the emergenCITY center [LOEWE/1/12/519/03/05.001(0016)/72]. C.\,Reich is supported by the Konrad Zuse School of Excellence in Learning and Intelligent Systems (\href{https://eliza.school}{ELIZA}) through the DAAD programme Konrad Zuse Schools of Excellence in Artificial Intelligence, sponsored by the Federal Ministry of Education \& Research. This work was also funded by the ETH Foundation Project 2025-FS-352, the SNSF Advanced Grant 216260.}

\bibliographystyle{splncs04}
\bibliography{main}

{
    \small
    \begin{thebibliography}{124}

    \bibitem[98]{monorun} Chen, H., Huang, Y., Tian, W., Gao, Z., Xiong, L.: MonoRUn: Monocular 3D object detection by reconstruction and uncertainty propagation. In: CVPR. pp. 10379–10388 (2021)
    
    \bibitem[99]{pseudo_stereo} Chen, Y., Dai, H., Ding, Y.: Pseudo-stereo for monocular 3D object detection in autonomous driving. In: CVPR. pp. 877–887 (2022)
    
    \bibitem[100]{monopair} Chen, Y., Tai, L., Sun, K., Li, M.: MonoPair: Monocular 3D object detection using pairwise spatial relationships. In: CVPR. pp. 12090–12099 (2020)
    
    \bibitem[101]{monomm} Fu, Y., Xu, Z., Fu, J., Xue, H., Tan, S., Li, L., Qing, S.: MonoMM: A multi-scale mamba-enhanced network for real-time monocular 3D object detection. J. Supercomput. \textbf{81}(3), 449 (2025)
    
    \bibitem[102]{gast2018lpn} Gast, J., Roth, S.: Lightweight probabilistic deep networks. In: CVPR. pp. 3369–3378 (2018)
    
    \bibitem[103]{homoflex} Gu, J., Wu, B., Fan, L., Huang, J., Cao, S., Xiang, Z., Hua, X.: Homography loss for monocular 3D object detection. In: CVPR. pp. 1070–1079 (2022)
    
    \bibitem[104]{monodtr} Huang, K., Wu, T., Su, H., Hsu, W.H.: MonoDTR: Monocular 3D object detection with depth-aware transformer. In: CVPR. pp. 4002–4011 (2022)

    \bibitem[105]{uncloss} Kendall, A., Gal, Y.: What uncertainties do we need in Bayesian deep learning for computer vision?. In: NIPS \textbf{30} (2017)
    
    
    
    \bibitem[106]{coco} Lin, T., Maire, M., Belongie, S.J., Hays, J., Perona, P., Ramanan, D., Doll{\'{a}}r, P., Zitnick, C.L.: Microsoft COCO: Common objects in context. In: ECCV. vol. 8693, pp. 740–755 (2014)
    
    \bibitem[107]{ideal-m3d} Meier, J., Günther, F., Marin, R., Dhaouadi, O., Kaiser, J., Cremers, D.: IDEAL-M3D: Instance diversity- enriched active learning for monocular 3D detection. In: WACV. pp. 181-191 (2026)
    
    \bibitem[108]{multi-bin} Mousavian, A., Anguelov, D., Flynn, J., Kosecka, J.: 3D bounding box estimation using deep learning and geometry. In: CVPR. pp. 5632–5640 (2017)
    
    \bibitem[109]{dd3d} Park, D., Ambrus, R., Guizilini, V., Li, J., Gaidon, A.: Is pseudo-LiDAR needed for monocular 3D object detection? In: ICCV. pp. 3122–3132 (2021)
    
    \bibitem[110]{DID-M3D} Peng, L., Wu, X., Yang, Z., Liu, H., Cai, D.: DID-M3D: Decoupling instance depth for monocular 3D object detection. In: ECCV. vol. 13661, pp. 71–88 (2022)
    
    \bibitem[111]{rf_detr} Robinson, I., Robicheaux, P., Popov, M., Ramanan, D., Peri, N.: RF-DETR: Neural architecture search for real-time detection transformers. In: ICLR (2026)
    
    \bibitem[112]{imagenet} Russakovsky, O., Deng, J., Su, H., Krause, J., Satheesh, S., et al.: ImageNet large scale visual recognition challenge. Int. J. Comput. Vis. \textbf{115}(3), 211–252 (2015)
    
    \bibitem[113]{pdr} Sheng, H., Cai, S., Zhao, N., Deng, B., Zhao, M., Lee, G.H.: PDR: Progressive depth regularization for monocular 3D object detection. IEEE Trans. Circuits Syst. Video Technol. \textbf{33}(12), 7591–7603 (2023)
    
    \bibitem[114]{opa-3d} Su, Y., Di, Y., Zhai, G., Manhardt, F., Rambach, J.R., Busam, B., Stricker, D., Tombari, F.: OPA-3D: Occlusion-aware pixel-wise aggregation for monocular 3D object detection. IEEE Robotics Autom. Lett. \textbf{8}(3), 1327–1334 (2023)
    
    \bibitem[115]{sparse_rcnn} Sun, P., Zhang, R., Jiang, Y., Kong, T., Xu, C., Zhan, W., Tomizuka, M., Li, L., Yuan, Z., Wang, C., Luo, P.: Sparse R-CNN: end-to-end object detection with learnable proposals. In: CVPR. pp. 14454–14463 (2021)
    
    \bibitem[116]{efficienetnetv2} Tan, M., Le, Q.V.: EfficientNetV2: Smaller models and faster training. In: ICML. vol. 139, pp. 10096–10106 (2021)
    
    \bibitem[117]{monopgc} Wu, Z., Gan, Y., Wang, L., Chen, G., Pu, J.: MonoPGC: Monocular 3D object detection with pixel geometry contexts. In: ICRA. pp. 4842–4849 (2023)
    
    \bibitem[118]{yolobu} Xiong, K., Zhang, D., Liang, D., Liu, Z., Yang, H., Dikubab, W., Cheng, J., Bai, X.: You only look bottom-up for monocular 3D object detection. IEEE Robotics Autom. Lett. \textbf{8}(11), 7464–7471 (2023)

    \bibitem[119]{MixSKD} Yang, C., An, Z., Zhou, H., Cai, L., Zhi, X., Wu, J., Xu, Y., Zhang, Q.: MixSKD: Self-knowledge distillation from mixup for image recognition. In: ECCV. vol. 13684, pp. 534–551 (2022)
    
    \bibitem[120]{monori} Yao, H., Han, P., Chen, J., Wang, Z., Qiu, Y., Wang, X., Wang, Y., Chai, X., Cao, C., Jin, W.: MonOri: Orientation-guided PnP for monocular 3D object detection. IEEE Trans. Neural Netw. Learn. Syst. \textbf{36}(10), 19068–19080 (2025)
    
    \bibitem[121]{rt_detr} Zhao, Y., Lv, W., Xu, S., Wei, J., Wang, G., Dang, Q., Liu, Y., Chen, J.: DETRs beat YOLOs on real-time object detection. In: CVPR. pp. 16965–16974. (2024)

    \bibitem[122]{RotContinuity} Zhou, Y., Barnes, C., Lu, J., Yang, J., Li, H.: On the continuity of rotation representations in neural networks. In: CVPR. pp. 5745–5753 (2019)
    
    \bibitem[123]{monoef} Zhou, Y., He, Y., Zhu, H., Wang, C., Li, H., Jiang, Q.: Monocular 3D object detection: An extrinsic parameter free approach. In: CVPR. pp. 7556–7566 (2021)
    
    \bibitem[124]{monoatt} Zhou, Y., Zhu, H., Liu, Q., et al.: MonoATT: Online monocular 3D object detection with adaptive token transformer. In: CVPR. pp. 17493–17503 (2023)
    
    \bibitem[125]{deformable_detr} Zhu, X., Su, W., Lu, L., Li, B., Wang, X., Dai, J.: Deformable DETR: Deformable transformers for end-to-end object detection. In: ICLR (2021)
    
    \end{thebibliography}
}
\clearpage
\newpage

\author{
Johannes Meier\,\textsuperscript{1,2,3,4,\dag,*\,}\orcidlink{0009-0000-2227-8271}\and
Jonathan Michel\,\textsuperscript{3,4,\dag\,}\orcidlink{0009-0004-2314-8283}\and
Oussema Dhaouadi\,\textsuperscript{1,2,3,4\,}\orcidlink{0009-0008-6842-5220}\and
Yung-Hsu Yang\,\textsuperscript{2\,}\orcidlink{0000-0003-0044-515X}\and
Christoph Reich\,\textsuperscript{3,4,5,6\,}\orcidlink{0000-0002-8616-1627}\and 
Zuria Bauer\,\textsuperscript{2\,}\orcidlink{0000-0001-8447-2344}\and
Stefan Roth\,\textsuperscript{5,6,7\,}\orcidlink{0000-0001-9002-9832}\and
Marc Pollefeys\,\textsuperscript{2,8\,}\orcidlink{0000-0003-2448-2318}\and
Jacques Kaiser\,\textsuperscript{1\,}\orcidlink{0000-0001-7487-6185}\and
Daniel Cremers\,\textsuperscript{3,4,6\,}\orcidlink{0000-0002-3079-7984}
}

\authorrunning{J. Meier et al.}
\titlerunning{Supplement: LeAD-M3D}



\institute{
\textsuperscript{1}\,DeepScenario\;\;\;\;
\textsuperscript{2}\,ETH Zurich\;\;\;\;
\textsuperscript{3}\,TU Munich\;\;\;\;
\textsuperscript{4}\,MCML\;\;\;\;
\textsuperscript{5}\,TU Darmstadt\\
\textsuperscript{6}\,ELIZA\;\;\;\;
\textsuperscript{7}\,hessian.AI\;\;\;\;
\textsuperscript{8}\,Microsoft \\
\url{https://deepscenario.github.io/LeAD-M3D/}
}

\title{Supplementary Material\\ LeAD-M3D: Leveraging Asymmetric Distillation\\ for Real-Time Monocular 3D Detection} 
\maketitle
\setcounter{section}{0}
\renewcommand\thesection{\Alph{section}}
\setcounter{page}{1}
\pagenumbering{roman}
\setcounter{figure}{5}
\setcounter{table}{10}
\setcounter{equation}{6}

\noindent 
In this appendix, we provide further architecture and training details of LeAD-M3D (\cref{sec:supp_ours_architecture_train_details}), followed by a comprehensive set of additional experiments and benchmark evaluations (\cref{sec:supp_experiments}). We conclude with a discussion on potential research avenues (\cref{sec:future_work}).

\section{\ourmethod Architecture and Training Details}
\label{sec:supp_ours_architecture_train_details}

Here, we provide further model and training details of \ourmethod, complementing \cref{sec:method} of the main paper. These include details of our loss functions, extending our baseline from 2D to 3D detection. Additionally, we provide details of our CGI\textsubscript{3D} approach. Finally, we provide further dataset and implementation details.

\subsection{Extending the 2D Baseline to 3D}
\label{sec:supp_extend_baseline}

While the main paper focuses on our core methodological contributions over the baseline, we provide additional details here on lifting the 2D YOLOv10~\cite{yolov10} detector to the monocular 3D object detection (M3D) setting. To achieve this, we follow established best practices by extending YOLOv10 with the standard 3D detection heads and 3D losses proposed in MonoLSS~\cite{monolss}.

\subsubsection{2D/3D Detection Heads.} 
In \cref{tab:heads}, we detail the architectural modifications to YOLOv10. Specifically, we replace the original 2D-specific outputs with new 2D heads and add standard 3D detection heads from MonoLSS\cite{monolss} (\cf \cref{tab:heads}).

\begin{table}[t]
  \centering
  \caption{\textbf{\ourmethod prediction heads overview.} In case of KITTI~\cite{kitti} / Waymo~\cite{waymo}, 12 bins and 12 residuals are learned (multi-bin). In case of Rope3D~\cite{rope3d}, the $\operatorname{SO}(3)$ orientation matrix is learned.}
  \label{tab:heads}
  \small
  \setlength{\tabcolsep}{3.0pt}
  \begin{tabularx}{\columnwidth}{@{}Xcccc@{}}
    \toprule
    \textbf{Head} & \textbf{Channels} & \textbf{KITTI} & \textbf{Waymo} & \textbf{Rope3D} \\
    \midrule
    2D box-offset head ($\mathbf{\hat{O}}_i^{\rm 2D}$) & \num{2} & \checkmark & \checkmark & \checkmark \\
    2D size head ($\mathbf{\hat{S}}_i^{\rm 2D}$) & \num{2} & \checkmark & \checkmark & \checkmark \\
    Projected 3D center-offset head ($\mathbf{\hat{O}}_i^{\rm 3D}$) & \num{2} & \checkmark & \checkmark & \checkmark \\
    3D size head ($\mathbf{\hat{S}}_i^{\rm 3D}$)       & \num{3} & \checkmark & \checkmark & \checkmark \\
    3D depth head ($z_i$)        & \num{1} & \checkmark & \checkmark & \checkmark \\
    Depth-uncertainty head $(\hat{\sigma}_i)$ & \num{1} & \checkmark & \checkmark & \checkmark \\
    Multi-bin orientation head ($\hat{\Theta}_{i}^{\text{bin}}, \hat{\Theta}_{i}^{\text{res}}$) & \num{24} & \checkmark & \checkmark & \xmark \\
    $\operatorname{SO}(3)$ orientation head ($\hat{R}_i^{\text{a}}$) & \num{6} & \xmark & \xmark & \checkmark \\
    \bottomrule
  \end{tabularx}
\end{table}

\subsubsection{Loss Functions.} While our distillation loss is outlined in the main paper (\cf \cref{eq:distill_loss}), here we provide more details on the supervised part of our overall loss function (\cf \cref{eq:total_loss}). In particular, we provide details of each individual supervised 2D and 3D detection loss function.

\paragraph{Preliminaries.} Extending our notation introduced in \cref{sec:method}, we seek to detect objects using 3D bounding boxes. Additionally, we also detect objects in 2D. In particular, given the input RGB image $\mathbf{I} \in \mathbb{R}^{3\times H \times W}$, we seek to predict both 3D bounding boxes $\hat{B}(\mathbf{I}) = \{\mathbf{\hat{b}}^{\text{3D}}_1, \ldots, \mathbf{\hat{b}}^{\text{3D}}_{M'}\}$ and corresponding 2D bounding boxes $\hat{B}^{\text{2D}}(\mathbf{I}) = \{\mathbf{\hat{b}}^{\text{2D}}_1, \ldots, \mathbf{\hat{b}}^{\text{2D}}_{M'}\}$. Each 3D bounding box $\mathbf{\hat{b}}^{\text{3D}}_i$ is defined by its 3D center location $\mathbf{\hat{C}}^{\text{3D}}_{i}=(x_i, y_i, z_i) \in \mathbb{R}^3$, 3D size $\mathbf{\hat{S}}_{i}^{\text{3D}}=(w_i, h_i, l_i) \in \mathbb{R}^3$, orientation represented by a rotation matrix $R_{i} \in \operatorname{SO}(3)$, and category $c_i \in \mathbb{C}$ (\eg, ``Vehicle'' or ``Pedestrian''). Each 2D bounding box $\mathbf{\hat{b}}^{\text{2D}}_i$ is analogously defined by its 2D center location in pixel space $\mathbf{\hat{C}}^{\text{2D}}_{i}=(\tilde{x}_i, \tilde{y}_{i}) \in \mathbb{R}^2$, 2D size in pixel space $\mathbf{\hat{S}}^{\text{2D}}_{i}=(\tilde{w}_i, \tilde{h}_i) \in \mathbb{R}^2$, and category $c_i\in \mathbb{C}$.

Only during training, we have given ground-truth 3D bounding boxes $B(\mathbf{I}) = \{\mathbf{b}^{\text{3D}}_1, \ldots, \mathbf{b}^{\text{3D}}_M\}$ and corresponding 2D bounding boxes $B^{\text{2D}}(\mathbf{I}) = \{\mathbf{b}^{\text{2D}}_1, \ldots, \mathbf{b}^{\text{2D}}_M\}$. Here, $\mathbf{C}_{i}^{\text{3D}}=(\overline{x}_i, \overline{y}_i, \overline{z}_i)\in \mathbb{R}^3$ is the ground-truth center in 3D, $\mathbf{S}_{i}^{\text{3D}} \in \mathbb{R}^3$ the ground-truth 3D size, $\overline{R}_{i} \in \operatorname{SO}(3)$ the ground-truth rotation, and $\overline{c}_i \in \mathbb{C}$ the ground-truth category of bounding box $\mathbf{b}^{\text{3D}}_i$. Analogously, $\mathbf{C}_{i}^{\text{2D}}\in \mathbb{R}^2$ is the ground-truth center in 2D, $\mathbf{S}_{i}^{\text{2D}} \in \mathbb{R}^2$ the ground-truth 2D size of $\mathbf{b}^{\text{2D}}_i$

Following YOLOv10~\cite{yolov10}, each predicted 3D bounding box $\mathbf{\hat{b}}^{\text{3D}}_i$ is associated with a specific anchor center $\mathbf{\hat{A}}_{i}\in\{s(\Omega - \frac{1}{2})| \Omega\in\mathbb{N}, \Omega s\leq H\}\times \{s(\Omega - \frac{1}{2})\;|\; \Omega\in\mathbb{N}, \omega s\leq W\}$ with stride $s\in\{8, 16\}$. For each anchor center $\mathbf{\hat{A}}_{i}$, our network predicts 2D center offsets $\mathbf{\hat{O}}^{\text{2D}}_{i}\in\mathbb{R}^{2}$, 2D size $\mathbf{\hat{S}}^{\text{2D}}_{i}$, projected 3D center offsets $\mathbf{\hat{O}}^{\text{3D}}_{i}\in\mathbb{R}^{2}$, 3D size $\mathbf{\hat{S}}^{\text{3D}}_{i}$, depth $z_{i}$, and allocentric orientation ${R}^{\text{a}}_{i}\in\mathbb{R}_{6}$. Additionally, we predict depth uncertainty $\hat{\sigma}_{i}$ and a class probability $p_{i}\in[0, 1]^{|\mathbb{C}|}$. Note, we follow existing work~\cite{brazil2023omni3d,cdrone,monoct} and predict allocentric orientation ${R}^{\text{a}}_{i}\in\mathbb{R}^{6}$ instead of $R_{i}$. During inference, the network predicts ${R}^{\text{a}}_{i}$, which is converted to $R_{i}$ using Gram–Schmidt, 2D location $\mathbf{C}_{i}$, and camera intrinsics $k\in \mathbb{R}^{3\times 4}$. Using the anchor center $\mathbf{\hat{A}}_{i}$, 3D offset, depth $\mathbf{\hat{O}}^{\text{3D}}_{i}$, camera intrinsics $k$, and classification $\arg\max p_{i}$, we can obtain the full 3D bounding-box representation $\mathbf{\hat{b}}^{\text{3D}}_{i}$. For an overview, see \cref{tab:heads}.

We use CM$_{\text{3D}}$ to match predictions with ground-truth detections. \ourmethod follows YOLOv10 \cite{yolov10} and uses two identical and parallel heads. For the first head, a one-to-one matching between predictions and ground truth is obtained using CM$_{\text{3D}}$. For the second head, a one-to-many matching is applied (also using CM$_{\text{3D}}$), in which a ground-truth detection is matched to multiple predictions. This head stabilizes training and is removed during inference \cite{yolov10}. For the sake of simplicity, we assume matched bounding boxes and describe our detection losses for one-to-one matching. The detection losses can be extended to the one-to-many by summing over multiple predictions.

\paragraph{2D Offset Loss.} The 2D offset loss supervises the 2D center offsets $\mathbf{\hat{O}}^{\text{2D}}_{i}$ given the ground-truth center $\mathbf{C}^{\text{2D}}_{i}$ and anchor location $\mathbf{\hat{A}}_{i}$ by
\begin{equation}\label{eq:loss_2d_offset}
    \mathcal{L}^{\text{o2D}} = \frac{1}{\lvert B(\mathbf{I})\rvert} \sum_{i=1}^{|B(\mathbf{I})|}  \bigl\lVert \mathbf{\hat{O}}^{\text{2D}}_{i} - \mathbf{\hat{A}}_{i} - \mathbf{C}^{\text{2D}}_{i}\bigr\rVert_{1}.
\end{equation}
Note that the anchor locations are not learned but fixed.

\paragraph{2D Size Loss.} We supervise the predicted 2D size $\mathbf{\hat{S}}^{\text{2D}}_i$ using the ground-truth 2D size $\mathbf{S}^{\text{2D}}_i$ by:
\begin{equation}\label{eq:loss_2d_size}
\mathcal{L}^{\text{s2D}} = \frac{1}{\lvert B(\mathbf{I})\rvert} \sum_{i=1}^{|B(\mathbf{I})|} \bigl\lVert \mathbf{\hat{S}}^{\text{2D}}_i - \mathbf{S}^{\text{2D}}_i\bigr\rVert_{1}.
\end{equation}

\paragraph{3D Offset Loss.} Similar to \cref{eq:loss_2d_offset}, we supervise the projected 3D center offsets $\mathbf{\hat{O}}$. As $\mathbf{\hat{O}}$ is in pixel-space, we project $\mathbf{\hat{C}}^{\text{3D}}_{i}$ into pixel space using the intrinsics $k$. We compute the 3D offset loss by:
\begin{equation}\label{eq:loss_3d_offset}
    \mathcal{L}^{\text{o3D}} = \frac{1}{\lvert B(\mathbf{I})\rvert} \sum_{i=1}^{|B(\mathbf{I})|}  \bigl\lVert \mathbf{\hat{O}}^{\text{3D}}_{i} - \mathbf{\hat{A}}_{i} - \pi(\mathbf{\hat{C}}^{\text{3D}}_{i}, k)\bigr\rVert_{1},
\end{equation}
where $\pi(\mathbf{\hat{C}}^{\text{3D}}_{i}, k)$ denotes the perspective projection into pixel space. While both $\mathcal{L}^{\text{o2D}}$ and $\mathcal{L}^{\text{o3D}}$ supervise in pixel space, the center location in 2D and 3D (projected to pixel space) of the same object may differ, requiring two separate predictions and losses.

\paragraph{3D Size Loss.} Similar to \cref{eq:loss_2d_size}, we supervise the 3D size $\mathbf{\hat{S}}^{\text{3D}}_i$ using the ground-truth 3D size $\mathbf{S}^{\text{3D}}_i$ by:
\begin{equation}
\mathcal{L}^{\text{s3D}} = \frac{1}{\lvert B(\mathbf{I})\rvert} \sum_{i=1}^{|B(\mathbf{I})|} \bigl\lVert \mathbf{\hat{S}}^{\text{3D}}_i - \mathbf{S}^{\text{3D}}_i\bigr\rVert_{1}.
\end{equation}

\paragraph{Depth Loss.} To supervise depth, we use a Laplacian depth loss \cite{uncloss,gast2018lpn} incorporating predicted uncertainty $\hat{\sigma}_{i}$ to provide a robust depth estimate. Given the ground-truth depth $\overline{z}_{i}$ obtained from $\mathbf{C}^{\text{3D}_{i}}$, we supervise the predicted depth $z_{i}$ by:
\begin{equation}
\mathcal{L}^{\text{d}} = \frac{1}{\lvert B(\mathbf{I})\rvert} \sum_{i=1}^{|B(\mathbf{I})|} \Bigl( \sqrt{2} \frac{\lvert z_{i} - \overline{z}_{i} \rvert}{\hat{\sigma}_{i}} + \log \hat{\sigma}_{i} \Bigr).
\end{equation}

\paragraph{Orientation Loss.} We supervise orientation in allocentric space by:
\begin{equation}
\mathcal{L}^{\text{rot}} = \frac{1}{\lvert B(\mathbf{I})\rvert} \sum_{i=1}^{|B(\mathbf{I})|} \bigl\lVert R^{\text{a}}_{i} - \bar{R}^{\text{a}}_{i} \bigr\rVert_{1},
\end{equation}
where $\bar{R}^{\text{a}}_{i}$ is the ground-truth allocentric orientation obtained from $\bar{R}_{i}$.

Alternatively, for KITTI~\cite{kitti} and Waymo~\cite{waymo}, we follow the multi-bin approach~\cite{multi-bin}. In these datasets, orientation is represented by a single angle $\theta_i \in [-\pi, \pi]$ rather than a full rotation matrix. This angle is discretized into $\xi$ equally spaced bins, where each bin $\delta \in \{1, \dots, \xi\}$ has a fixed center $\Phi_\delta = -\pi + (\delta - 0.5)\frac{2\pi}{\xi}$. The ground-truth bin probability $\Theta_{i,\delta}^{\text{bin}}$ is defined as:
\begin{equation}
\Theta_{i,\delta}^{\text{bin}} = \begin{cases} 1 & \text{if } \lvert \theta_i - \Phi_\delta \rvert \pmod{2\pi} \leq \frac{\pi}{\xi} \\ \text{otherwise} \end{cases}
\end{equation}
The ground-truth residual $\Theta_{i,\delta}^{\text{res}}$ represents the normalized offset of $\theta_i$ from its corresponding bin center $\Phi_\delta$, defined as:
\begin{equation}
\Theta_{i,\delta}^{\text{res}} = \theta_i - \Phi_\delta
\end{equation}
The rotation loss is then formulated as:
\begin{equation}
\mathcal{L}^{\text{rot}} = \frac{1}{\lvert B(\mathbf{I})\rvert} \sum_{i=1}^{|B(\mathbf{I})|} \left[ \text{CE}\bigl(\Theta^{\text{bin}}_{i}, \hat{\Theta}_{i}^{\text{bin}}\bigr) + \sum_{\delta=1}^{\xi} \mathbbm{1}_{\{\Theta_{i,\delta}^{\text{bin}}=1\}} \bigl\lvert \Theta_{i,\delta}^{\text{res}} - \hat{\Theta}_{i,\delta}^{\text{res}}\bigr\rvert \right]
\end{equation}
where $\hat{\Theta}_{i}^{\text{bin}}$ and $\hat{\Theta}_{i,\delta}^{\text{res}}$ are the predicted bin probabilities and residuals, respectively, and the indicator function $\mathbbm{1}$ ensures the residual loss is only computed for the bin containing the ground-truth orientation. We set $\xi=12$.

\paragraph{Classification Loss.} Following YOLOv10~\cite{yolov10}, we supervised classification using a binary cross-entropy loss. While a multi-class task, we supervise classification using a binary cross-entropy loss. YOLOv10~\cite{yolov10} showed that treating each class prediction independently, rather than enforcing a mutually exclusive softmax distribution, is beneficial for downstream accuracy. In particular, we supervised the classification probabilities $p_{i}\in[0, 1]^{|\mathbb{C}|}$, given the classification label $\overline{c}_{i}$ by
\begin{equation}
    \mathcal{L}^{\text{cls}} = - \sum_{i=1}^{|B(\mathbf{I})|}\sum_{j=1}^{|\mathbb{C}|} \left[\mathbbm{1}_{\{j=\overline{c}_{i}\}}\log(p_{i, j}) + (1 - \mathbbm{1}_{\{j=\overline{c}_{i}\}})\log(1 - p_{i, j}),
\right]
\end{equation}
where $\mathbbm{1}$ is again the indicator function that is one if $j$ is equal to $\overline{c}_{i}$, else zero.

\paragraph{Total Loss.} Our total loss is composed of four terms
\begin{equation}
    \mathcal{L} = \lambda^{\text{cls}}\underbrace{\mathcal{L}^{\text{cls}}}_{\text{Classification}} + \underbrace{\mathcal{L}^{\text{2D}}}_{\text{2D detection}} + \underbrace{\mathcal{L}^{\text{3D}}}_{\text{3D detection}} + \lambda^{\text{distill}}\underbrace{\mathcal{L}^{\text{distill}}}_{\text{Distillation}},
\end{equation}
where $\mathcal{L}^{\text{distill}}$ is our distillation loss introduced in \cref{sec:method:distill}. The 2D detection loss $\mathcal{L}^{\text{2D}}$ is computed by
\begin{equation}
\begin{aligned}
\mathcal{L}^{\text{2D}} &= \lambda^{\text{o2D}} \mathcal{L}^{\text{o2D}} + \lambda^{\text{s2D}} \mathcal{L}^{\text{s2D}}
\end{aligned}
\end{equation}
and the 3D detection loss $\mathcal{L}^{\text{3D}}$ is computed by
\begin{equation}
\begin{aligned}
\mathcal{L}^{\text{3D}} &= \lambda^{\text{o3D}}\mathcal{L}^{\text{o3D}} +  \lambda^{\text{s3D}}\mathcal{L}^{\text{s3D}}
+ \lambda^{\text{rot}}\mathcal{L}^{\text{rot}}
+ \lambda^{\text{d}}\mathcal{L}^{\text{d}}.
\end{aligned}
\end{equation}
We weigh the different loss terms with $\lambda^{\text{cls}}=1.0$, $\lambda^{\text{distill}}=0.1$, $\lambda^{\text{o2D}} = 0.02$, $\lambda^{\text{s2D}} = 0.02$, $\lambda^{\text{o3D}}=1.0$, $\lambda^{\text{s3D}} = 1.0$, $\lambda^{\text{rot}} = 1.0$, and $\lambda^{\text{z}} = 1.0$.

\subsection{CGI$_{\text{3D}}$ Details}
\label{sec:supp_cgi_algorithm}
In \cref{sec:speedup}, we introduce the CGI$_{\text{3D}}$ to speed up inference by running the 2D and 3D heads only on high-confidence regions. Here, we provide further details.

\begin{algorithm}[ht!]
    \caption{CGI$_{\text{3D}}$ PyTorch-style pseudo-code. We do not show classification for the sake of simplicity.}
    \label{algo:cgi}
\begin{lstlisting}[style=pythonstyle]
def cgi_3d(feat_8, feat_16, f_cls_8, f_cls_16, f_2d_8,
           f_3d_8, f_2d_16, f_3d_16, k):
    """
    Args:
        feat_8 (Tensor): Features w/ stride 8.
        feat_16 (Tensor): Features w/ stride 16.
        f_cls_8 (Module): Classification head at stride 8
        f_cls_16 (Module): Classification head at stride 16
        f_2d_8 (Module): 2D regression head at stride 8
        f_3d_8 (Module): 3D regression head at stride 8
        f_2d_16 (Module): 2D regression head at stride 16
        f_3d_16 (Module): 3D regression head at stride 16
        k (int): Top-k sampling parameter.
         
    Returns:
        (Tensor, Tensor, Tensor): 2D boxes, 3D boxes & scores
    """
    # 1-2. Dense class scores from stride-8 and stride-16 grids
    p8, p16 = f_cls_8(feat_8), f_cls_16(feat_16)
    # 3-5. TaggedConcat & top-K selection across multi-scale maps
    s8_flat, s16_flat = p8.flatten(1), p16.flatten(1)
    s_all = torch.cat([s8_flat, s16_flat], dim=1)
    scores_k, idx_k = torch.topk(s_all, k, dim=1)
    # Determine origin levels (tags) and local coordinates
    num_p8 = s8_flat.size(1)
    is_l8 = idx_k < num_p8
    adj_idx = torch.where(is_l8, idx_k, idx_k - num_p8)
    # 6. Extract 3x3 patches from respective FPN maps
    q8 = extract_3x3_patches(feat_8, adj_idx)
    q16 = extract_3x3_patches(feat_16, adj_idx)
    q = torch.where(is_l8.view(-1, 1, 1, 1), q8, q16)
    # 7-8. Sparse regression: Apply level-specific heads via masking
    m = is_l8.flatten()
    boxes2d = torch.zeros((indices_k.numel(), 4))
    boxes3d = torch.zeros((indices_k.numel(), 10))
    boxes2d[m],  boxes3d[m]  = f_2d_8(q[m]),  f_3d_8(q[m])
    boxes2d[~m], boxes3d[~m] = f_2d_16(q[~m]), f_3d_16(q[~m])
    return boxes2d, boxes3d, scores_k
\end{lstlisting}
\end{algorithm}

\subsubsection{Algorithmic Details.}
The core of Confidence-Gated 3D Inference (CGI$_{\text{3D}}$) lies in shifting the candidate selection process before the expensive 3D regression stage. As illustrated in \cref{fig:instance_targeting}, the process consists of four key steps: \emph{(1)} computing dense classification scores, \emph{(2)} identifying the top-$k$ most confident object centers ($k$ analyzed in \cref{tab:k_ablation_study}) across feature scales, \emph{(3)} extracting local $3\times3$ patches at these sparse locations, and \emph{(4)} applying the 2D and 3D regression heads exclusively to these high-confidence regions. This sparse inference strategy is detailed using pseudo-code in Algorithm \labelcref{algo:cgi}.

\subsubsection{Comparison with Previous Works.}
Our approach differentiates itself from previous works in multiple aspects. Unlike convolutional approaches such as GUPNet~\cite{gupnet} and MonoLSS~\cite{monolss}, which rely on RoI-Align\cite{mask_rcnn} and consequently bilinear interpolation for feature extraction, CGI3D utilizes a simplified extraction of fixed $3\times3$ patches that can be implemented with simple indexing. While those methods estimate both classification confidence and 2D bounding boxes densely across the entire feature map, our approach requires only dense estimation of classification scores. This allows \ourmethod to focus 2D and 3D box estimation exclusively on high-confidence patches. While MonoDiff~\cite{monodiff} relies on a multi-stage process where an off-the-shelf 2D detector generates proposals for a separate diffusion refinement network, \ourmethod operates as a single, unified network.

Transformer-based models like MonoDETR~\cite{monodetr}, Deformable DETR\cite{deformable_detr}, and Sparse R-CNN\cite{sparse_rcnn} rely on fixed sets of learned queries. Because these queries are typically associated with specific spatial regions, a larger number of them is typically required to ensure sufficient coverage across the entire image. CGI$_{\text{3D}}$ instead leverages the baseline's existing top-$k$ filtering logic to adaptively gate the 3D regression head. By moving this selection process earlier in the pipeline, we bypass expensive 3D reasoning for low-confidence regions using a much smaller $k$ than the typical query count in transformer-based architectures. Distinct from methods like RT-DETR\cite{rt_detr} or RF-DETR\cite{rf_detr}, which introduce dual uncertainty metrics or neural architecture search, CGI$_{\text{3D}}$ requires only the vanilla classification confidence.

Different from existing approaches, CGI$_{\text{3D}}$ is applied exclusively during inference. The 2D and 3D head is only bypassed for proposals that would have been filtered out by the final detector anyway. This allows CGI$_{\text{3D}}$ to improve efficiency while producing virtually the same output as the full/vanilla inference. 

\subsection{Implementation Details}
\label{sec:supp_impl_details}
This section provides additional implementation details, dataset details, and experimental configurations, extending the implementation details provided in the main paper. 

\subsubsection{Hardware, Runtime Evaluation and Training Time.}
All runtime evaluations are also conducted on the same hardware, using a single NVIDIA RTX \num{8000} GPU. Runtime is measured as the total execution time, measured from the start of the forward pass through the completion of all post-processing operations for inference on a single image. This measurement includes associated CPU activity within the post-processing step, but excludes image-to-GPU data transfer. In particular, for exact measurements, we perform a warmup phase followed by \num{1000} forward passes with batch size \num{1} and report the median inference time. While TensorRT speeds up inference, we observe an insignificant variance in downstream accuracy when applied. These stem from numerical quantization and typically lead to a variance in AP of about $\pm$\SI{0.2}{\%}. More specifically, we utilize TensorRT with the following settings: half tensor, simplify, workspace \SI{16}{GB}, image size \num{384}$\times$\num{1280}, and batch size \num{1}.

\subsubsection{Model Architecture.}
We extract multi-scale features at strides 8 and 16. 
We follow our baseline and employ dual prediction heads: a one-to-one head and a one-to-many head.
During training, both heads provide supervision, which improves gradient flow and learning stability.
At inference time, we discard the one-to-many head and use only the one-to-one head.
This design eliminates the need for non-maximum suppression, which accelerates inference while maintaining a favorable accuracy–efficiency trade-off through dense training supervision.

\subsubsection{Training Configuration.}
In addition to the training details described in the main paper, we also use gradient accumulation to simulate a virtual batch size of \num{64}.
For distillation, we use the features from the one-to-one head as targets for the student heads. During initial experiments, we observed that a model size L (between X and M) offers no benefits over the B model size.
We, therefore, exclude the L model from our model family.
CM$_{\text{3D}}$ replaces the prediction-to-ground truth scoring of \baseline. Other matching criteria (\eg, maximum anchor radius) of YOLOv10~\cite{yolov10} are kept.

\begin{figure}[ht!]
    \centering
    \begin{tabularx}{\linewidth}{@{}ccc@{}}
       \sffamily Image 1 & \sffamily Image 2 & \sffamily MixUP image \\
       \includegraphics[width=0.33\linewidth]{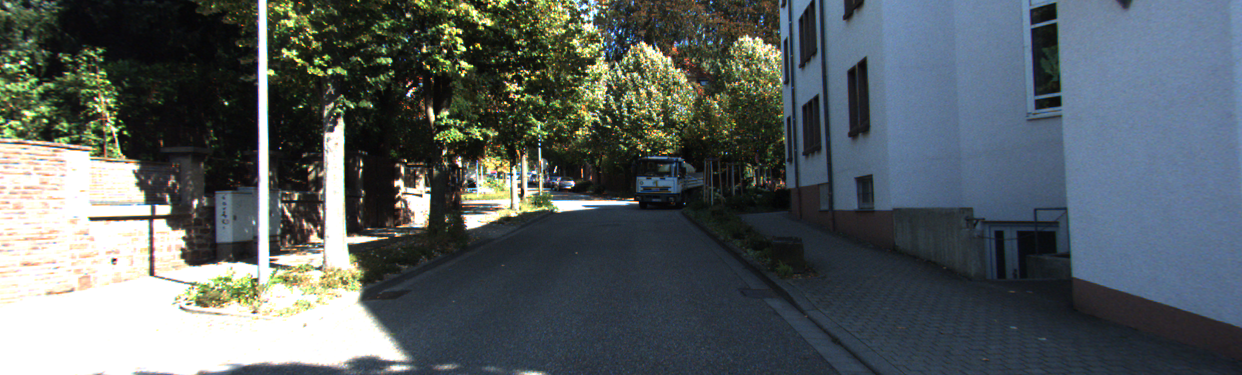} & \includegraphics[width=0.33\linewidth]{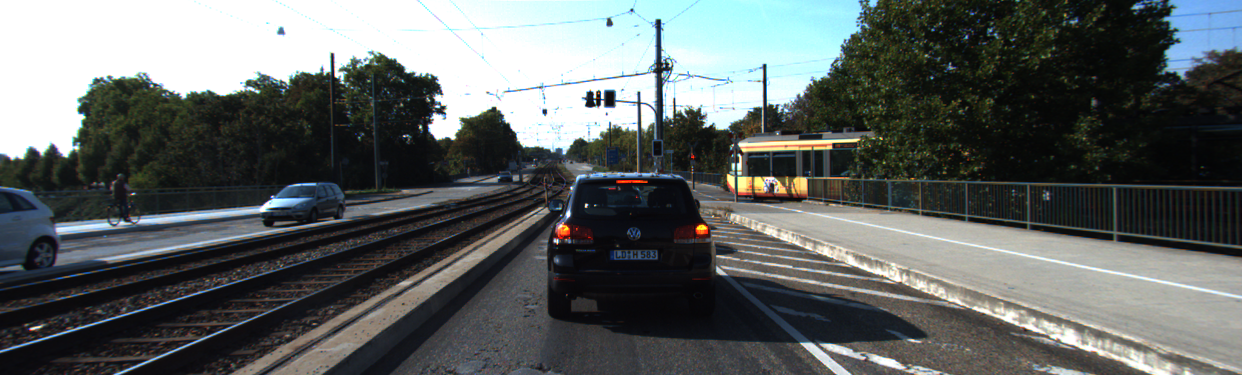} & \includegraphics[width=0.33\linewidth]{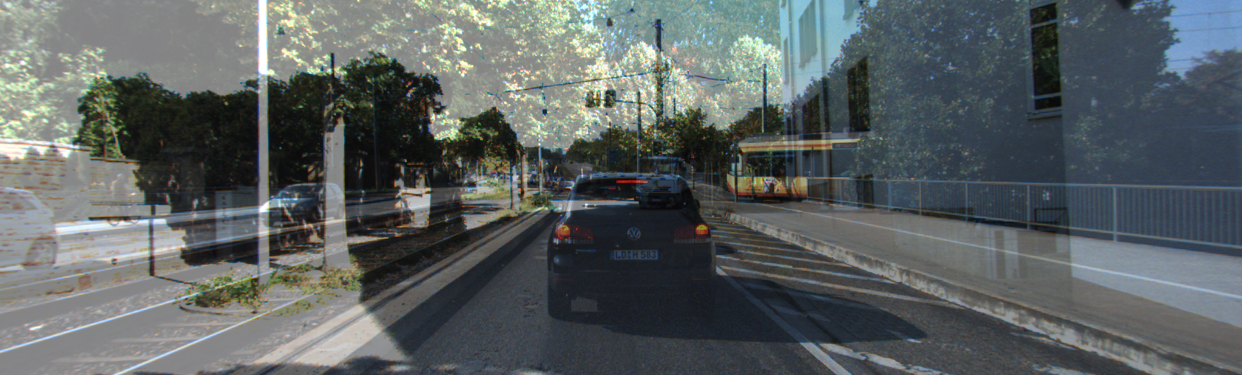} \\

       \includegraphics[width=0.33\linewidth]{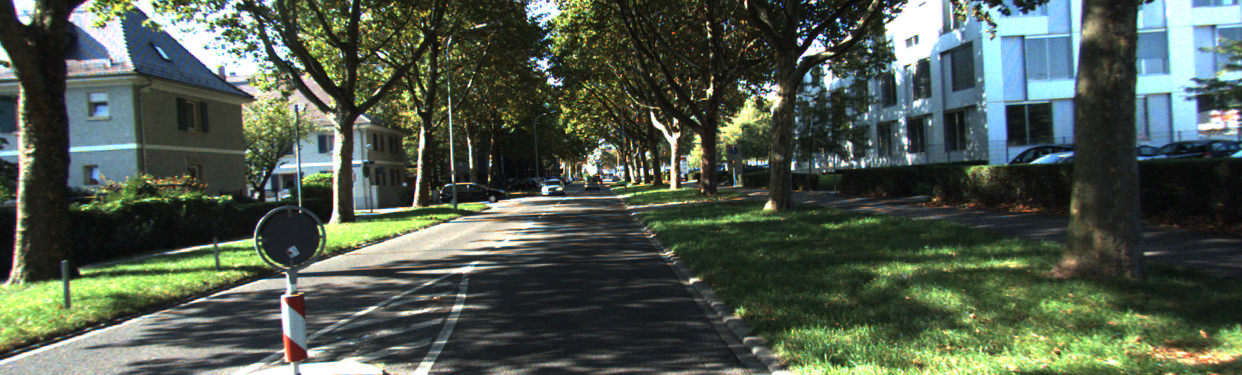} & \includegraphics[width=0.33\linewidth]{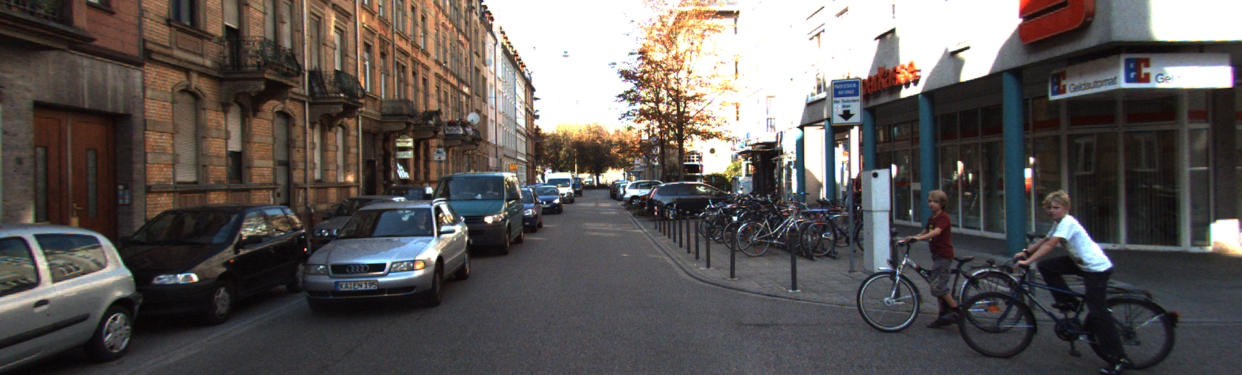} & \includegraphics[width=0.33\linewidth]{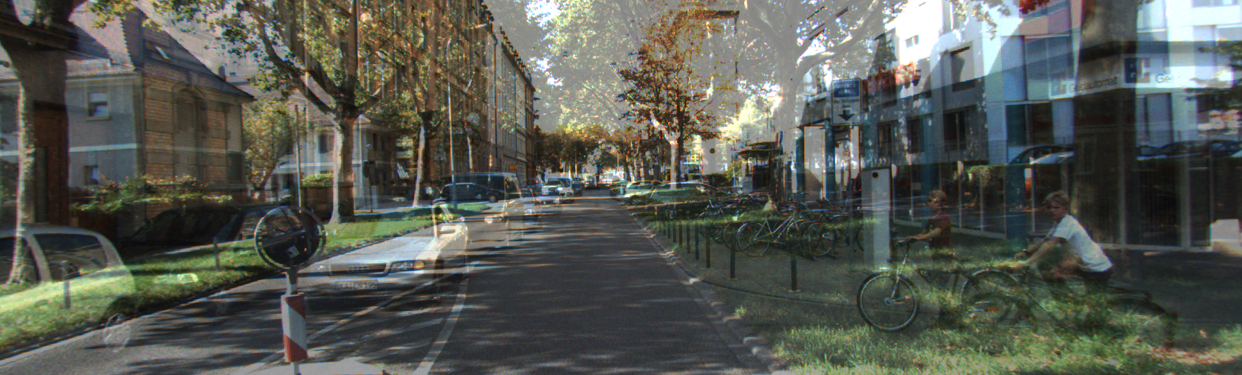} \\

       \includegraphics[width=0.33\linewidth]{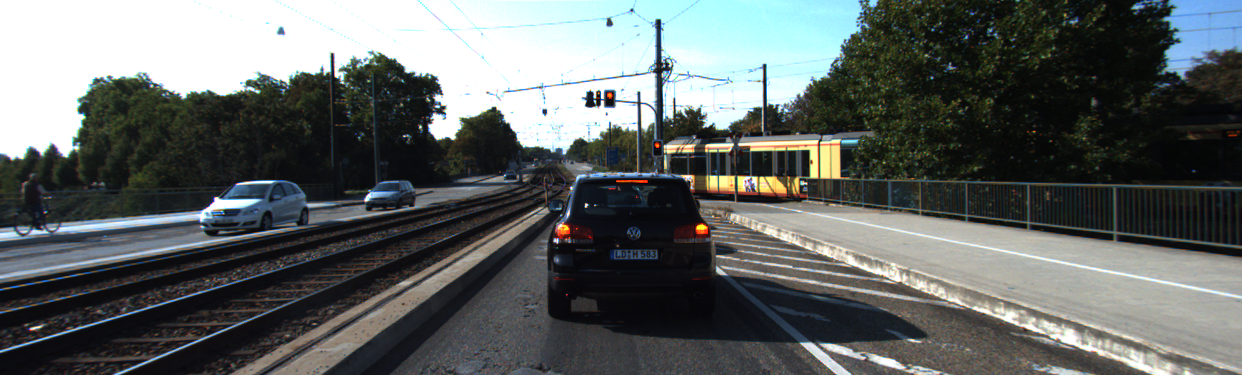} & \includegraphics[width=0.33\linewidth]{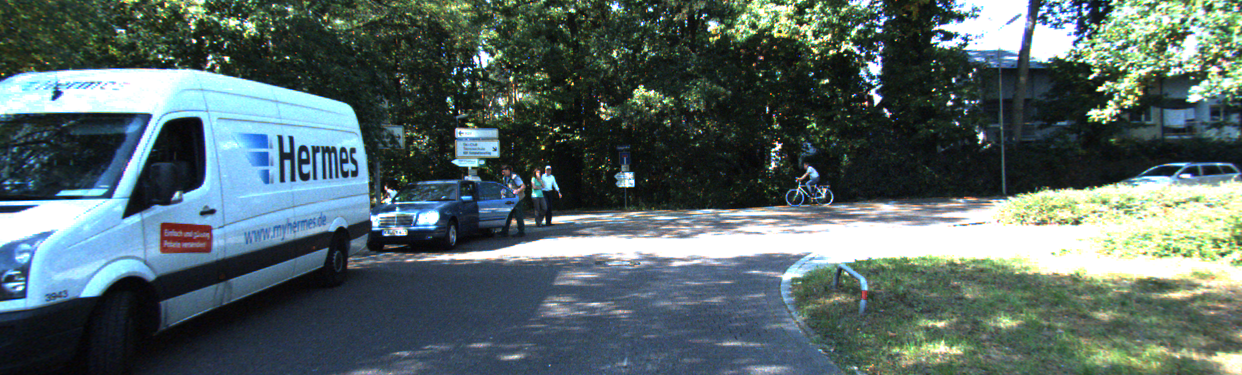} & \includegraphics[width=0.33\linewidth]{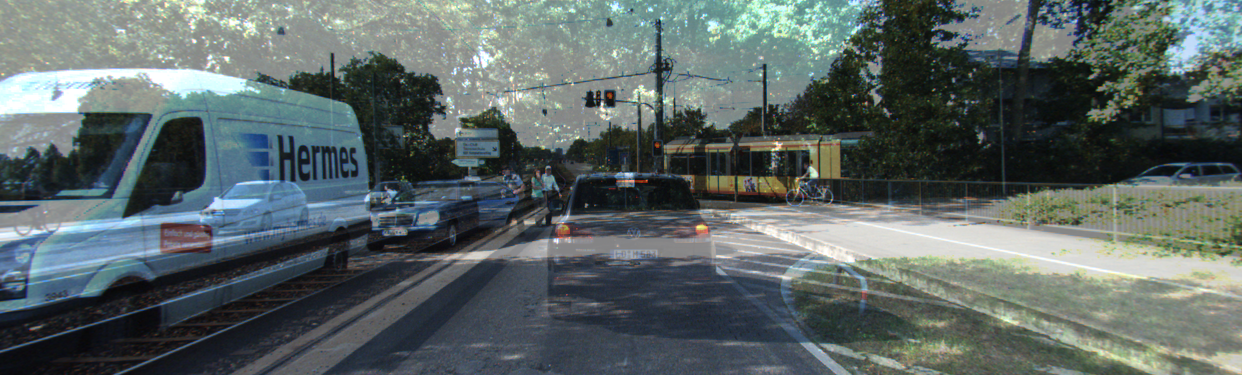} \\

       \includegraphics[width=0.33\linewidth]{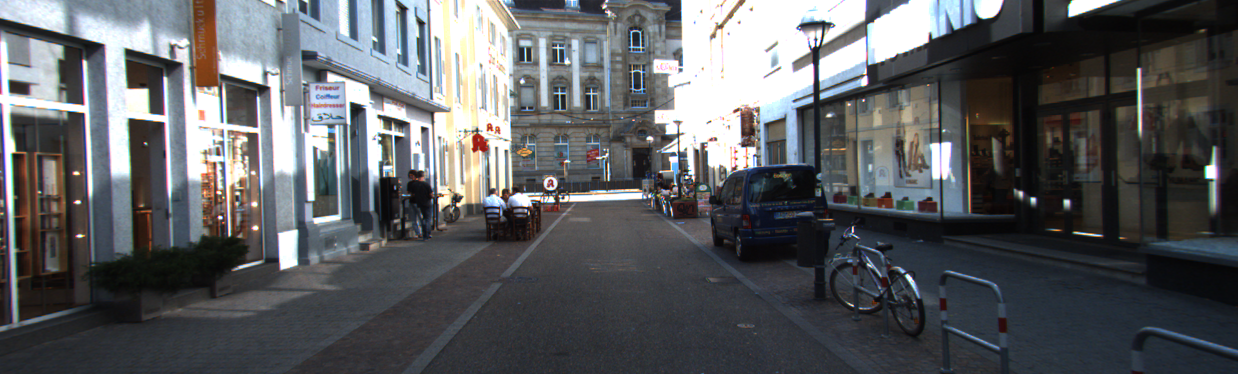} & \includegraphics[width=0.33\linewidth]{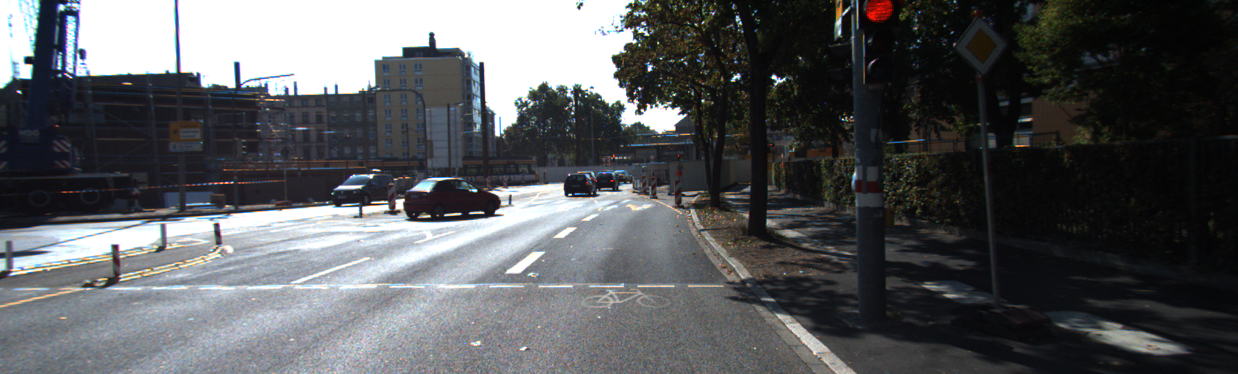} & \includegraphics[width=0.33\linewidth]{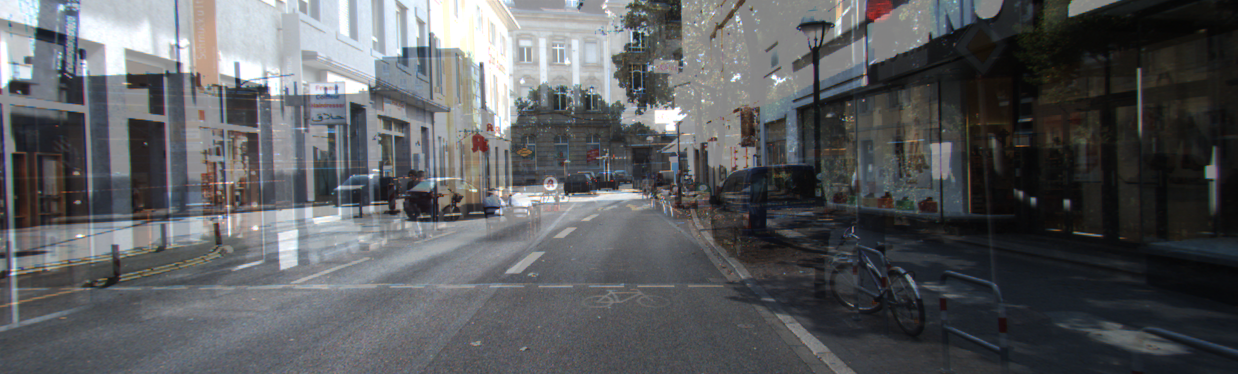}
    \end{tabularx}
    \caption{\textbf{MixUp example.} We provide examples of MixUp-augmented KITTI \cite{kitti} images. First image \emph{(left)}, second image \emph{(middle)}, MixUp image \emph{(right)}. Blending ratio of \num{0.5} is applied. For simplicity, we only apply MixUp and omit other augmentations (\eg, translation and cropping).}
    \label{fig:mixupexample}
\end{figure}

\subsubsection{Data Augmentation.}
We apply the following image augmentations during training: horizontal flipping with probability \num{0.5}, random cropping with probability \num{0.5}, translation with a maximum shift of \num{0.1} times the image size, scaling with resize factors between \num{0.6} and \num{1.4}, and MixUp with probability of \num{0.5} and a blending ratio of \num{0.5} (following previous work \cite{monolss,cdrone}). We empirically observe that the detection accuracy is robust to modest changes of the MixUp probabilities and blending ratios; for simplicity, we keep both at \num{0.5}. In \cref{fig:mixupexample}, we provide examples of MixUp-augmented images.

\subsubsection{Dataset-Specific Settings.}
Both teacher training and student distillation are performed for the same number of epochs: \num{400} on KITTI~\cite{kitti}, \num{100} on Rope3D \cite{rope3d}, \num{45} on Waymo~\cite{waymo}, and \num{24} on NuScenes \cite{nuScenes}.
For Waymo~\cite{waymo} and Rope3D\cite{rope3d}, we use \SI{50}{\%} of the original image resolution to accelerate experiments, while for nuScenes~\cite{nuScenes} we use \SI{75}{\%} of the available resolution.
For top-$k$ selection in CGI$_{\text{3D}}$, we use $k=25$ for KITTI, $k=50$ and Waymo, and $k=200$ for Rope3D. 
We follow GroundMix~\cite{cdrone} and learn virtual depth for Rope3D\cite{rope3d}, as it contains images with varying intrinsic parameters.

\section{Experiments}
\label{sec:supp_experiments}

In this section, we provide additional experimental results and detailed analysis to further validate the components of \ourmethod and support our main paper. We analyze the generalization of A2D2 to other network architectures (\cref{sec:genrealization_a2d2}), investigate sensitivity to initialization/pre-training and backbone selection (\cref{sec:pretraining_backbone}), and provide additional analyses \wrt to design choices and hyperparameters (\cref{sec:a2d2_ablation_supp_detail}). We also provide a high-level ablation and insights into denoising and distillation (\cref{sec:denoise_distill}). Additionally, we ablate the classification score in CM$_{\text{3D}}$ (\cref{sec:cm3d_ablation}), analyze the top-$k$ parameter of CM$_{\text{3D}}$ (\cref{sec:cm3d_topk}), give detailed runtime results including a per-model-size CGI$_{\text{3D}}$ analysis (\cref{sec:runtime_analysis_supp}), and analyze the scaling behavior of student \vs teacher models (\cref{sec:contribution_by_model_size}). Finally, we show additional benchmark results extending our main paper's results (\cref{sec:comparison_sota_supp}), including a cross-dataset generalization study on nuScenes~\cite{nuScenes} (\cref{sec:supp_nuscenes_cross_dataset}), and present additional qualitative results (\cref{sec:qual_results_rope3d}).

\subsection{Generalization of A2D2 to Other Model Architectures}
\label{sec:genrealization_a2d2}
\begin{table}[t]
    \centering
    \setlength{\tabcolsep}{8.7pt}
    \caption{\textbf{Analysis of A2D2 with different detectors.} We extend our analysis of A2D2 in \cref{tab:main_ablation} by demonstrating the effectiveness of A2D2 for other detector architectures. We report \APDDDRFS\ (in \%, $\uparrow$) for the ``Car'' category on the KITTI~\cite{kitti} \emph{validation} set.}
    \small
    \setlength{\tabcolsep}{10.85pt}
    \begin{tabularx}{\linewidth}{
        >{\hspace{-\tabcolsep}\raggedright\columncolor{white}[\tabcolsep][\tabcolsep]}X
        S[table-format=2.2]
        S[table-format=2.2]
        S[table-format=2.2]
        }
        
      \toprule
       & \multicolumn{3}{c}{\textbf{\APDDDRFS Car\ $\uparrow$}} \\
      \cmidrule(lr){2-4}
      \multirow{-2}{*}{\vspace{0.5em}\textbf{Method}} & {\textbf{Easy}} & {\textbf{Mod.}} & {\textbf{Hard}} \\   \midrule

      MonoLSS \cite{monolss} & 25.91 & 18.29 &  15.94 \\
      MonoLSS \cite{monolss} + A2D2 &  27.53 &  20.05 &  16.99 \\
      \textit{Improvements} & \textcolor{tBlueStrong}{\textit{+1.62}} & \textcolor{tBlueStrong}{\textit{+1.76}} & \textcolor{tBlueStrong}{\textit{+1.05}} \\
      \midrule

      MonoCD \cite{monocd} & 24.22 & 18.27 & 15.42 \\   
      MonoCD \cite{monocd} + A2D2 &  28.30 & 21.00 & 18.56 \\      
      \textit{Improvements} & \textcolor{tBlueStrong}{\textit{+4.08}} & \textcolor{tBlueStrong}{\textit{+2.73}} & \textcolor{tBlueStrong}{\textit{+3.14}} \\
      
      \bottomrule
    \end{tabularx}
    \label{tab:supp_ablation_distill_diff_baseline}

\end{table}

While \cref{tab:main_ablation} demonstrates the effectiveness of our A2D2 approach with our \ourmethod model architecture, we provide additional A2D2 results using different model architectures (\cf \cref{tab:supp_ablation_distill_diff_baseline}). In particular, we utilize A2D2 with both MonoLSS~\cite{monolss} and MonoCD~\cite{monocd}.

As both models employ slightly different 2D detection architectures, we retain their original label-assignment strategies and do not use CM$_{\text{3D}}$. As both models also only offer a single model size, we first train supervised and use the resulting model as both the teacher and student (\ie, self-distillation) for A2D2.

Despite architectural differences to \ourmethod, we observe a consistent gain in detection accuracy across all evaluation metrics when using A2D2 with Mono\-LSS and MonoCD (\cf \cref{tab:supp_ablation_distill_diff_baseline}). This demonstrates that distillation using A2D2 can generally be utilized to improve 3D detection accuracy and seems agnostic to model architectures.

\subsection{Backbone Analysis and Pre-Training}
\label{sec:pretraining_backbone}
By default, we employ the YOLOv10~\cite{yolov10} backbone across our experiments as it offers a favorable runtime--accuracy trade-off and provides a versatile range of model sizes. We initialize these parameters using MS-COCO~\cite{coco} pre-trained weights. While MS-COCO provides denser 2D bounding box supervision than ImageNet, the dataset itself is over 100 times smaller. Existing M3D methods predominantly rely on ImageNet~\cite{imagenet} pre-training. Here, we analyze the impact of pre-training and backbone architectures.

\begin{table}[t]
  \centering
    \caption{\textbf{Pre-training analysis.} We analyze the detection accuracy of \ourmethodx and two baselines with and without pre-training. We report \APDDDRFS\ Mod. (in \%, $\uparrow$) for the ``Car'' category on the KITTI~\cite{kitti} \emph{validation} set. MonoLSS~\cite{monolss}, MonoDGP~\cite{monodgp} uses ImageNet~\cite{imagenet} and \ourmethodx uses MS-COCO~\cite{coco} for pre-training.}
    \small
    \begin{tabularx}{\linewidth}{l@{\hspace{5.0em}}c@{\hspace{3.0em}}l}
        \toprule
      \textbf{Method} & \textbf{Pre-training} & \textbf{\APDDDRFS\ Mod.\ Car\ $\uparrow$} \\ \midrule

      MonoDGP \cite{monodgp} & \cmark & \cbarm{tGreenStrong}{6.135em}{0.7em}{22.34} \\
      MonoDGP \cite{monodgp} & \xmark & \cbarm{tBlueLight}{3.825em}{0.7em}{13.93 \textcolor{tRedStrong}{\footnotesize{-8.41}}} \\
      \midrule

      MonoLSS \cite{monolss} & \cmark & \cbarm{tGreenStrong}{5.022em}{0.7em}{18.29} \\
      MonoLSS \cite{monolss} & \xmark & \cbarm{tBlueLight}{4.352em}{0.7em}{15.85\textcolor{tRedStrong}{\footnotesize{ -2.44}}} \\
      \midrule

      \ourmethodx (Ours) & \cmark & \cbarm{tGreenStrong}{6.31em}{0.7em}{22.96} \\
      \ourmethodx (Ours) & \xmark & \cbarm{tBlueLight}{5.28em}{0.7em}{19.21 \textcolor{tRedStrong}{\footnotesize{-3.75}}} \\
      \bottomrule

    \end{tabularx}
    \label{tab:kitti_ablation_no_pretraining}
\end{table}

\begin{table}[t]
  \centering
    \caption{\textbf{Backbone analysis.} We analyze the detection accuracy of \ourmethod and our 3D baseline (\baseline) with different backbone architectures. For reference, we also report results with YOLOv10 backbones, our default choice. We report \APDDDRFS\ Mod. (in \%, $\uparrow$) for the ``Car'' category on the KITTI~\cite{kitti} \emph{validation} set.}
    \footnotesize
    \setlength{\tabcolsep}{3.0pt}
    \begin{tabularx}{\columnwidth}{>{\hspace{-\tabcolsep}\raggedright\columncolor{white}[\tabcolsep][\tabcolsep]}XXS[table-format=2.1]S[table-format=2.2]}
      \toprule
      \textbf{Method} & \textbf{Backbone} & \textbf{Time\ $\downarrow$}  &  \textbf{\APDDDRFS\ Mod.\ Car\ $\uparrow$} \\ \midrule

      \ourmethod & YOLOv10~\cite{yolov10} N & \best 8.2 & 18.38 \\
      \ourmethod & YOLOv10~\cite{yolov10} S & \second 10.9 & 20.67 \\
      \ourmethod & YOLOv10~\cite{yolov10} M & 12.9 & 22.40 \\
      \ourmethod & YOLOv10~\cite{yolov10} B & 14.1 & 22.65 \\
      \ourmethod & YOLOv10~\cite{yolov10} X & 24.3 & 22.96 \\ \midrule

      \baseline & EfficientNetV2~\cite{efficienetnetv2} S &  32.5 & 19.76 \\       \ourmethod  & EfficientNetV2~\cite{efficienetnetv2} S & 24.8 & 
      22.70 \\
      \midrulegray
      
      \baseline & EfficientNetV2~\cite{efficienetnetv2} M &  41.2 & 19.86 \\       \ourmethod & EfficientNetV2~\cite{efficienetnetv2} M &  31.6 &
      \second 23.36 \\
      \midrulegray
      
      \baseline & EfficientNetV2~\cite{efficienetnetv2} L &  58.5 & 19.99 \\
      \ourmethod & EfficientNetV2~\cite{efficienetnetv2} L & 51.4 & \best 
      23.64 \\
      \bottomrule

    \end{tabularx}
    \label{tab:kitti_ablation_backbone}
\end{table}

\subsubsection{Effect of Pre-Trained Weights.}
To assess the impact of pre-training, we compare the accuracy of \ourmethod against other methods~\cite{monodgp,monolss} when trained without any pre-trained weights. As shown in \cref{tab:kitti_ablation_no_pretraining}, our model trained from scratch outperforms MonoLSS~\cite{monolss} even with pre-training. Still, our accuracy drops by \SI{3.75}{\%} AP without pre-training while MonoLSS only drops by \SI{2.44}{\%}. In contrast, MonoDGP~\cite{monodgp} experiences a more significant drop of \SI{8.41}{\%} in AP. We hypothesize that this instability stems from attention, which benefits more from pre-training. Still, to achieve strong detection accuracy, with \ourmethod and other approaches, pre-training is favorable.

\begin{table}[t]
  \centering
    \caption{\textbf{Extended backbone analysis.} We extend the \emph{validation} results in \cref{tab:kitti_ablation_backbone} and evaluate the largest model \ourmethod with EfficientNetV2~\cite{efficienetnetv2} (ENV2) L on the KITTI \cite{kitti} \emph{test} set. We report \APDDDRFS\ and \APBEVRFS\ (both in \%, $\uparrow$) for the category ``Car''.}
    \footnotesize
    \setlength{\tabcolsep}{2.565pt}
    \begin{tabularx}{\columnwidth}{>{\hspace{-\tabcolsep}\raggedright\columncolor{white}[\tabcolsep][\tabcolsep]}XS[table-format=2.2]S[table-format=2.2]S[table-format=2.2]S[table-format=2.2]S[table-format=2.2]S[table-format=2.2]}
      \toprule
      \multirow{2}{*}{\textbf{Method}} & \multicolumn{3}{c}{\textbf{\APDDDRFS}\ $\uparrow$} & \multicolumn{3}{c}{\textbf{\APBEVRFS}\ $\uparrow$} \\ 
      \cmidrule(lr){2-4} \cmidrule(lr){5-7}
      & \textbf{Easy} &  \textbf{Mod.} &  \textbf{Hard} &  \textbf{Easy} &  \textbf{Mod.} &  \textbf{Hard} \\ 
      \midrule

      MonoTAKD \cite{monotakd} & 27.91 & 19.43 & 16.51 & \second 38.75 & \best 27.76 & \second 24.14 \\ \midrule
      \ourmethodx & \second 30.76 & \best 21.20 & \second 18.76 & 38.33 & 26.57 & 23.74\\
      \ourmethod w/ ENV2~\cite{efficienetnetv2} L & \best 31.18 & \second 20.86 & \best 19.10 & \best 39.74 & \second 27.35 & \best 24.43\\
      \bottomrule

    \end{tabularx}\label{tab:kitti_test_set_efficientnetv2}
\end{table}

\subsubsection{Backbone Analysis.} To further analyze different backbone architectures and pre-training protocols, we replace the YOLOv10 backbone with EfficientNetV2 \cite{efficienetnetv2} (ENV2). In particular, we utilize different ENV2 sizes (S, M, and L). ENV2 backbones are pre-trained on ImageNet1k (classification), unlike YOLOv10 backbones, which are pre-trained on MS-COCO with bounding-box supervision. Following our standard distillation procedure, we train the largest variant (w/ ENV2-L backbone) as the teacher and distill knowledge to student models with sizes S, M, and L.

We first report results on the KITTI validation set, in \cref{tab:kitti_ablation_backbone}, and compare against \ourmethod with YOLOv10 backbones and our baseline (\baseline) with ENV2 backbones. We observe that \ourmethod with ENV2 significantly improves accuracy over our baseline with ENV2. While ENV2 backbones are less efficient than YOLOv10 backbones, \ourmethod with ENV2 M and L outperform \ourmethodx in AP. This demonstrates that 2D box pre-training is not strictly needed to obtain strong detection accuracy with \ourmethod, as ENV2 uses ImageNet1k pre-training.

To confirm the strong detection accuracy of \ourmethod with ENV2 L, we report results on the KITTI test set (submission required), in \cref{tab:kitti_test_set_efficientnetv2}. Also on the private test set, \ourmethod with ENV2 L achieves strong detection accuracy. In particular, \ourmethod with ENV2 L outperforms \ourmethodx on \num{5} out of \num{6} accuracy metrics.

\subsection{Distillation and Denoising Ablation}\label{sec:denoise_distill}

\begin{table}[t]
    \centering
    \caption{\textbf{Distillation and denoising ablation.} We extend \cref{tab:main_ablation,tab:ablation_distillation,tab:supp_ablation_distillation} by providing a more high-level ablation bt disentangling distillation (A2D2) and denoising (MixUp). We toggle each component independently and report car \APDDDRFS\ (in \%, $\uparrow$) for different difficulty levels on the KITTI~\cite{kitti} \emph{validation} set using the B model size.}
    \footnotesize\sisetup{table-number-alignment=right}
    \setlength{\tabcolsep}{8.6pt}
    \begin{tabularx}{\columnwidth}{>{\hspace{-\tabcolsep}\centering\columncolor{white}[\tabcolsep][\tabcolsep]}c c S[table-format=2.2] S[table-format=2.2] S[table-format=2.2]}
        \toprule
        {\textbf{Distillation (A2D2)}} & {\textbf{Denoising (MixUp)}} & {\textbf{Easy}\,$\uparrow$} & {\textbf{Mod.}\,$\uparrow$} & {\textbf{Hard}\,$\uparrow$} \\
        \midrule
        \xmark & \xmark & 25.13 & 19.12 & 16.69 \\
        \xmark & \cmark & 26.26 & 20.43 & 17.71 \\
        \cmark & \xmark & \second 26.73 & \second 20.85 & \second 18.14 \\
        \cmark & \cmark & \best 28.44 & \best 22.65 & \best 19.87 \\
        \bottomrule
    \end{tabularx}
    \label{tab:rebuttal_denoising_vs_distillation}
\end{table}

While \cref{tab:main_ablation,tab:ablation_distillation,tab:supp_ablation_distillation} analyze our individual contributions, in \cref{tab:rebuttal_denoising_vs_distillation} we provide a high-level ablation. In particular, we disentangle both distillation (A2D2) and denoising (MixUp) and report detection results on KITTI validation. Without \emph{both} distillation and denoising, we achieve an AP (Mod.) of \SI{19.12}{\%}. Introducing only denoising in the form of MixUp augmentations improves detection accuracy by \SI{1.31}{\%} in AP (Mod.). Only using distillation leads to an even larger increase of \SI{1.73}{\%} in AP (Mod.). Combining both distillation and denoising yields the strongest result accuracy of \SI{22.65}{\%}, improving over no distillation and denoising by \SI{3.53}{\%}, both in AP (Mod.). This demonstrates the effectiveness of both distillation and denoising.

\begin{figure}[t]
    \centering
    \tikzset{every picture/.style={/utils/exec={\sffamily\scriptsize}}}
    \begin{tikzpicture}[spy using outlines={white, magnification=2.0, minimum width=1.19cm, minimum height=1.19cm, fill=white, connect spies}]

        \node[inner sep=0pt, outer sep=0pt] at (-0.3333333\linewidth, 0) {\includegraphics[width=0.3333333333\linewidth]{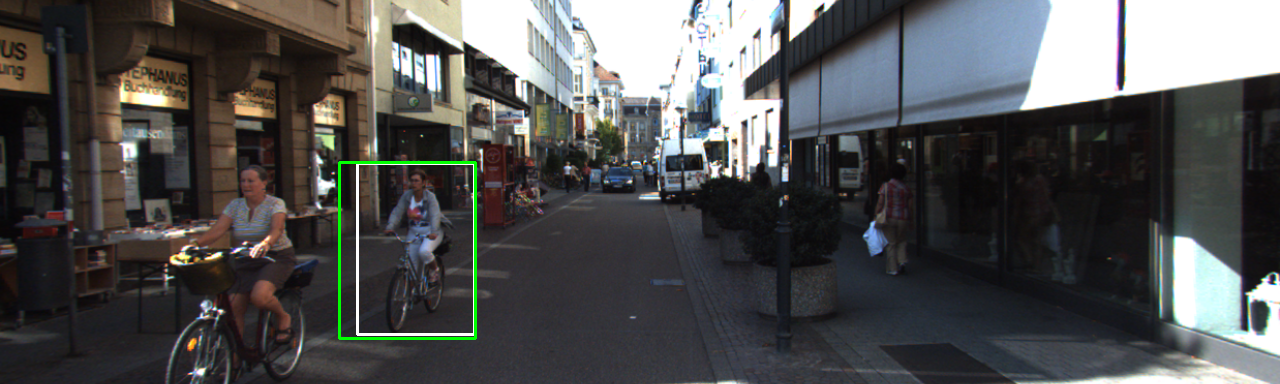}};
        \node[inner sep=0pt, outer sep=0pt] at (0.0\linewidth, 0) {\includegraphics[width=0.3333333333\linewidth]{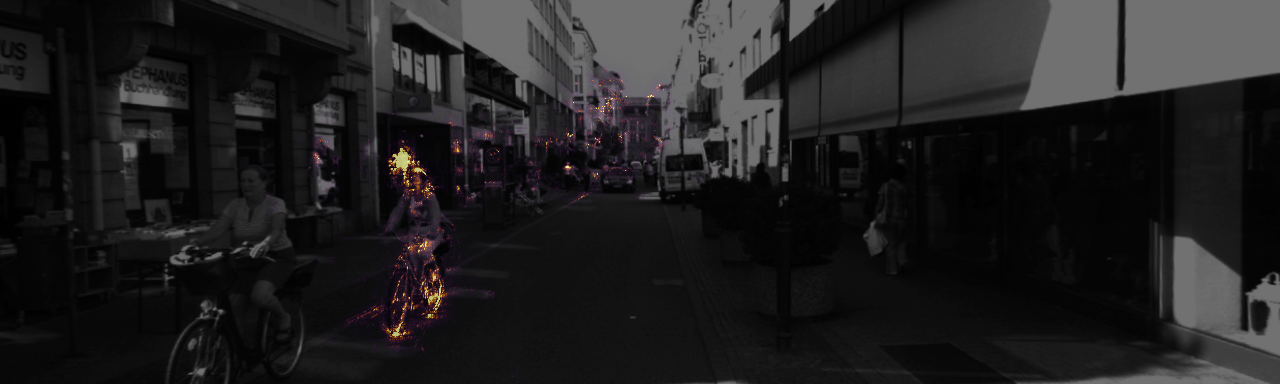}};
        \node[inner sep=0pt, outer sep=0pt] at (0.3333333\linewidth, 0) {\includegraphics[width=0.3333333333\linewidth]{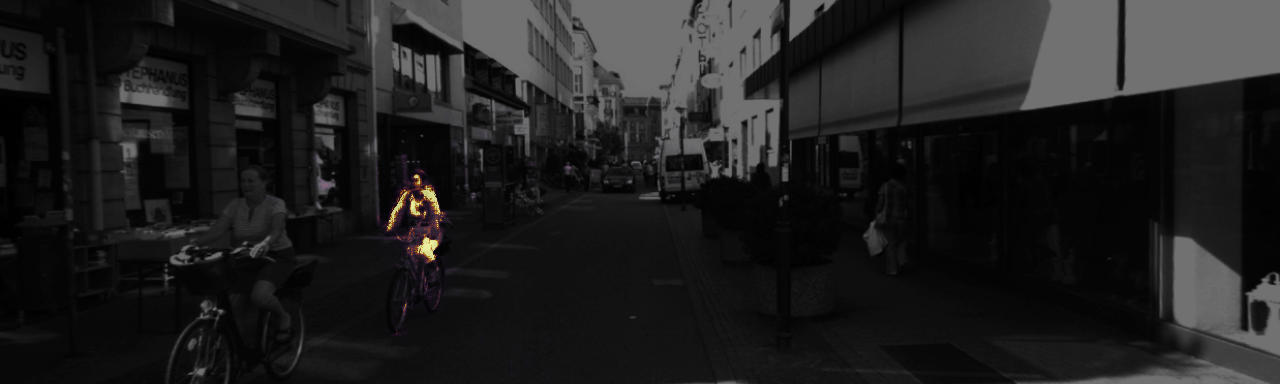}};
        
        \node[inner sep=0pt, outer sep=0pt, white] at (-0.065\linewidth, 0.46) {w/o denoise \& distill.};
        \node[inner sep=0pt, outer sep=0pt, white] at (0.2683333333\linewidth, 0.46) {w/ denoise \& distill.\hphantom{o}};
        \spy on (-0.1, 0.0) in node [draw opacity=1.0, white, fill=white, anchor=center, line width=0.5pt, inner sep=0pt, outer sep=0pt] at (1.425cm, 0.0cm);
        \spy on (3.3, -0.15) in node [draw opacity=1.0, white, fill=white, anchor=center, line width=0.5pt, inner sep=0pt, outer sep=0pt] at (5.485cm, 0.0cm);
    \end{tikzpicture}
    \caption{\textbf{Saliency analysis.} Integrated gradients~\cite{intgrad} \wrt object depth (right cyclist) on KITTI \emph{val}. Without denoising \& distillation \emph{(middle)}, saliency leaks into noisy background regions. With both \emph{(right)}, saliency concentrates on the object. Ground truth detection is in white, and for reference, we visualize the prediction of \ourmethod\ with denoising \& distillation in green, indicated in the input image \emph{(left)}.}
    \label{fig:ig_supp}
\end{figure}

In \cref{fig:ig_supp}, we extend \cref{fig:ig_main} and provide an additional saliency example. Similarly, without MixUp-based denoising and distillation (A2D2), saliency is diffuse and is located in noisy background regions and not perpetually concentrated on the detected object. In contrast, \ourmethod\ with both denoising and distillation predominantly focuses on the detected object. This demonstrates that denoising and distillation enable the student to learn more expressive features that focus on the object to be detected.



\subsection{A2D2 Ablation}\begin{table}[t]
    \caption{\textbf{Extended ablation of A2D2.} We extend the detailed ablation of A2D2 in \cref{tab:ablation_distillation} and report \APDDDRFS\ (in \%, $\uparrow$) for the ``Car'' category on the KITTI~\cite{kitti} \emph{validation} set by adding different components.}
    \centering
    \small
    \setlength{\tabcolsep}{6.5pt}
    \begin{tabularx}{\linewidth}{
        >{\hspace{-\tabcolsep}\raggedright\columncolor{white}[\tabcolsep][\tabcolsep]}X
        S[table-format=2.2]
        S[table-format=2.2]
        S[table-format=2.2]
        }
      \toprule

      \textbf{Method} & {\textbf{Easy\ $\uparrow$}} & {\textbf{Mod.\ $\uparrow$}} & {\textbf{Hard\ $\uparrow$}} \\ \midrule
      \ourmethodb (Ours) & \best 28.44 & \best 22.65 & \best 19.87 \\
      \hspace{0.5em} Ours w/ absolute quality indicators & \second 28.03 & \second  22.30 & \second 19.68\\
      \hspace{0.5em} Ours w/ simple distillation & 26.10 & 20.50 & 17.73 \\ 
      \hspace{0.5em} Ours w/ MixSKD~\cite{MixSKD} & 25.14 & 19.68 & 18.22\\
       \midrule
      Ours w/o A2D2 & 26.26 & 20.43 & 17.71 \\ 
      \bottomrule 
      \label{tab:supp_ablation_distillation}
      \end{tabularx}
\end{table}

\label{sec:a2d2_ablation_supp_detail}
Building upon the A2D2 study in \cref{tab:ablation_distillation} of the main paper, we provide further ablation studies in \cref{tab:supp_ablation_distillation} to evaluate additional methodological variations and alternative strategies. We use \APDDDRFS\ Moderate (Mod.) as the primary metric. Replacing the relative depth error with absolute depth error for the teacher-quality weight reduces AP by \SI{0.35}{\%}. We hypothesize that this overemphasizes near-range objects (with naturally smaller absolute residuals), which skews the data distribution and, thereby, harms generalization.

Replacing our sophisticated distillation pipeline with vanilla distillation (\ie, without importance and quality weighting, using backbone instead of depth feature for distillation, and without augmentation) leads to a substantial drop of \SI{2.15}{\%} in AP and only a minimal gain of \SI{0.07}{\%} \vs not applying knowledge distillation at all. This highlights the crucial role of our proposed distillation components. We hypothesize this minimal improvement occurs because, without the quality weighting, the student only benefits from the difference in model capacity; however, the B and X models tend to perform similarly (\cf \cref{tab:exp_kitti_val}), leading to negligible gains.

Finally, we evaluate replacing our distillation strategy (A2D2) with MixSKD \cite{MixSKD}. MixSKD also uses an augmentation-based self-distillation framework but for image classification. We show the methodical differences to A2D2 below. This substitution results in the largest performance drop of \SI{2.97}{\%} in AP, an expected result. MixSKD is primarily designed for classification tasks, whereas our distillation approach (A2D2) was specifically designed for the geometric and structured nature of monocular 3D detection. 

To clarify this significant performance gap, we provide a detailed comparison highlighting the core methodological differences that explain the observed \SI{2.97}{\%} drop in AP. 
\textit{First}, MixSKD is tailored for image classification tasks, whereas our method is designed for M3D. As a result, MixSKD distills classification or backbone features, whereas our approach focuses on distilling depth features. 
\textit{Second}, MixSKD performs distillation across all spatial features. This means, in the case of MixSKD, most of the distillation happens for background rather than foreground features. In contrast, A2D2 distills instance-specific features only. 
\textit{Third}, MixSKD relies on interpolated features for distillation, whereas our framework distills teacher features.
\textit{Fourth}, MixSKD performs online self-distillation, while we perform offline distillation (\cf A2D2 ablation in the main paper).
\textit{Fifth}, we employ quality and importance weighting to focus distillation on valuable teacher knowledge, a component absent in MixSKD.
\textit{Sixth}, beyond the core feature distillation loss, MixSKD requires additional components, including a discriminator head, distinct auxiliary branches, and multiple logit distillation losses. In contrast, our method outperforms it with a simpler design relying solely on a single feature distillation loss.

\begin{table}[t]
    \centering
    \caption{\textbf{CM$_{\text{3D}}$ classification score ablation.} We study the effect of removing the classification score from the CM$_{\text{3D}}$ matching criterion. We report car \APDDDRFS\ (in \%, $\uparrow$) for all difficulty levels on the KITTI~\cite{kitti} \emph{validation} set using the B model size.}
    \footnotesize\sisetup{table-number-alignment=right}
    \setlength{\tabcolsep}{4.0pt}
    \renewcommand{\arraystretch}{0.8}
    \begin{tabularx}{\columnwidth}{>{\hspace{-\tabcolsep}\raggedright\columncolor{white}[\tabcolsep][\tabcolsep]}X S[table-format=2.2] S[table-format=2.2] S[table-format=2.2]}
        \toprule
        \textbf{Method} & {\textbf{Easy}\,$\uparrow$} & {\textbf{Mod.}\,$\uparrow$} & {\textbf{Hard}\,$\uparrow$} \\
        \midrule
        \ourmethodb\ w/o class.\ score in CM$_{\text{3D}}$ & 26.21 & 20.73 & 18.19 \\
        \ourmethodb\ (Ours)                                & \best 28.44 & \best 22.65 & \best 19.87 \\
        \bottomrule
    \end{tabularx}
    \label{tab:supp_cm3d_ablation}
\end{table}

\subsection{CM$_{\text{3D}}$ Ablation}
\label{sec:cm3d_ablation}
\cref{tab:supp_cm3d_ablation} ablates the role of the classification score within CM$_{\text{3D}}$. The score serves two purposes: separating semantic categories and discriminating foreground from background. While category separation may appear straightforward on KITTI with only three classes, the \SI{1.92}{\%} drop in car mod.\ \APDDDRFS\ confirms that the score remains highly relevant, likely driven by the foreground/background discrimination role.

\begin{table}[t]
  \centering

  \caption{\textbf{Top-$k$ CGI\textsubscript{3D} analysis.} We evaluate the trade-off between detection accuracy and computational efficiency on the KITTI~\cite{kitti} \emph{validation} set for different $k$ values. We report \APDDDRFS\ Mod. (in \%, $\uparrow$) for different categories, runtime in ms, and GFLOPS. Runtime is reported for a single-image forward pass on a single NVIDIA RTX 8000 \emph{without} TensorRT. Setting $k$ to \num{25} leads to the best trade-off with virtually no loss in detection accuracy (\cf \cref{tab:ablationd_runtime}).}
  \footnotesize
  \setlength{\tabcolsep}{10.0pt}
  \begin{tabularx}{\linewidth}{>{\hspace{-\tabcolsep}\raggedright\columncolor{white}[\tabcolsep][\tabcolsep]}XS[table-format=2.2]S[table-format=2.2]S[table-format=2.2]S[table-format=2.2]S[table-format=2.2]S[table-format=2.2]}
    \toprule
    & & & \multicolumn{4}{c}{\textbf{\APDDDRFS\ Mod. $\uparrow$}} \\
    \cmidrule(lr){4-7}
    \textbf{\textit{k}} & \textbf{Time $\downarrow$} & \textbf{GFLOPs $\downarrow$} & \textbf{Car} & \textbf{Ped.} & \textbf{Cyc.} & \textbf{Avg.} \\ 
    \midrule
    5   & 9.69  & 14.13 & 18.51 & 9.76  & 3.92 & 10.73 \\
    10  & 9.69  & 14.13 & 18.41 & 10.44 & 4.22 & \second 11.02 \\
    \rowcolor{gray!20} 
    25  & 8.20  & 14.20 & 18.38 & 10.62 & 4.12 & \best 11.04 \\
    50  & 9.72  & 14.43 & 18.38 & 10.57 & 4.09 & 11.01 \\
    100 & 9.83  & 14.78 & 18.38 & 10.56 & 4.08 & 11.00 \\
    200 & 9.95  & 15.47 & 18.38 & 10.56 & 4.08 & 11.00 \\
    500 & 10.16 & 17.53 & 18.38 & 10.56 & 4.08 & 11.00 \\
    \bottomrule
  \end{tabularx}
  \label{tab:k_ablation_study}
\end{table}\label{sec:cm3d_topk}
\subsection{Analysis on the top-$k$ Parameter of CGI$_{\text{3D}}$}
\label{sec:ablation_top_k}
Since the hyperparameter $k$ is the most critical factor in the CGI$_{\text{3D}}$ architecture, as it controls exactly how many object candidates are passed to the 3D head, we conduct a dedicated analysis study in \cref{tab:k_ablation_study} to analyze its effect on the accuracy-runtime trade-off. We find that the highest 3D detection accuracy is achieved at $k=\text{\num{25}}$, which represents approximately only \SI{0.25}{\%} of all possible spatial locations on the KITTI feature maps. This suggests that the classification head effectively filters background noise and pre-selects the most relevant object candidates. While increasing $k$ beyond this point introduces additional GFLOPs and increases runtime, it yields no further gains in AP. This result confirms that \ourmethod properly captures the sparse nature of 3D scenes and operates with high efficiency.

\subsection{Runtime Analysis}
\label{sec:runtime_analysis_supp}
In the main paper, we demonstrate efficient real-time inference of \ourmethod (\cf \cref{tab:exp_kittI_test_fast}) and the effectiveness of CGI$_{\text{3D}}$ in reducing runtime (\cf \cref{tab:ablationd_runtime}). In this section, we provide further runtime and efficiency analysis. Specifically, we first show a runtime comparison across different hardware devices, then analyze the runtime of individual model components, and finally extend the CGI$_{\text{3D}}$ runtime analysis across all model sizes.

\begin{table}[t]
    \caption{\textbf{Runtime analysis: \ourmethod \vs baseline.} We compare \ourmethodn against our baseline (\baselinen) in terms of runtime on the KITTI~\cite{kitti} \emph{validation} set. We report runtime for a single-image forward pass in ms. CPU inference uses the Intel Xeon Gold 6254, and GPU inference uses a single NVIDIA RTX 8000.}
    \centering
    \footnotesize
    \setlength{\tabcolsep}{6.0pt} 
    \begin{tabularx}{\linewidth}{
        >{\hspace{-\tabcolsep}\raggedright\columncolor{white}[\tabcolsep][\tabcolsep]}Xrrrr
        }
      \toprule

      \textbf{Method} & {\textbf{Baseline $\downarrow$}} & {\textbf{Ours $\downarrow$}} & {\textbf{Change Abs.  $\downarrow$}} & {\textbf{Change Rel.  $\downarrow$}} \\
      \midrule

      CPU             & \SI{541.7}{ms} & \SI{243.4}{ms} & \textcolor{tBlueStrong}{\SI{-298.3}{ms}} & \textcolor{tBlueStrong}{\SI{-55.1}{\%}} \\
      GPU             & \SI{16.2}{ms}  & \SI{8.2}{ms}   & \textcolor{tBlueStrong}{\SI{-8.0}{ms}}   & \textcolor{tBlueStrong}{\SI{-49.4}{\%}} \\
      GPU w/ TensorRT & \SI{1.9}{ms}   & \SI{1.2}{ms}   & \textcolor{tBlueStrong}{\SI{-0.7}{ms}}   & \textcolor{tBlueStrong}{\SI{-36.8}{\%}} \\

      \bottomrule 
    \end{tabularx}
    \label{tab:supp_runtime_comparison}
\end{table}
\subsubsection{Runtime Comparison Across Devices.}
To evaluate the hardware-agnostic benefits of our approach, we compare the inference efficiency of \ourmethodn against the baseline \baselinen across CPU, standard GPU, and TensorRT environments in \cref{tab:supp_runtime_comparison}. Our modifications yield substantial speed-ups across all platforms: \SI{55}{\%} on the CPU and \SI{49}{\%} on the standard GPU, confirming benefits in compute-bound regimes. Critically, the speed-up decreases to \SI{37}{\%} in the sub-\SI{2}{ms} TensorRT environment. This clear dependency demonstrates the transition to an overhead-bound state in highly optimized hardware, where fixed costs (\eg, kernel launch latency) dominate the total runtime and thus limit the realized benefit of computational savings.
Additionally, in terms of memory footprint, \ourmethodn\ demonstrates superior efficiency, requiring a peak memory of \SI{169.4}{MB} and reserved memory of \SI{224.4}{MB}, compared to \SI{211.2}{MB} (peak) and \SI{249.6}{MB} (reserved) for \baselinen.

\begin{table}[t]
    \centering
    \caption{\textbf{Runtime breakdown.} We decompose runtime into different model components and compare LeAD-M3D N against our baseline (YOLOv10-M3D N) on the KITTI \cite{kitti} \emph{validation} set. We report runtime for a single-image forward pass in ms on a single NVIDIA RTX 8000 without using TensorRT.}
    \footnotesize
    \setlength{\tabcolsep}{4pt}
    \begin{tabularx}{\linewidth}{
        >{\hspace{-\tabcolsep}\raggedright\columncolor{white}[\tabcolsep][\tabcolsep]}X
        rrrr
        }
      \toprule
      \textbf{Component} & \textbf{Baseline} $\downarrow$ & \textbf{Ours} $\downarrow$ & \textbf{Change Abs.  $\downarrow$} & \textbf{Change Rel. $\downarrow$} \\ \midrule

      Backbone                & \SI{8.18}{ms}  & \SI{8.18}{ms} & ---             & --- \\
      \midrulegray
      Classification head     & \SI{0.41}{ms}  & \SI{0.41}{ms} & ---             & --- \\
      Patch extraction        & ---            & \SI{0.26}{ms} & \textcolor{tRedStrong}{\SI{+0.26}{ms}} & N/A \\
      2D/3D regression heads  & \SI{7.37}{ms}  & \SI{0.57}{ms} & \textcolor{tBlueStrong}{\SI{-6.80}{ms}} & \textcolor{tBlueStrong}{\SI{-92.3}{\%}} \\
      Decoding                & \SI{0.28}{ms}  & \SI{0.28}{ms} & ---             & --- \\ \midrule
      Total head              & \SI{8.06}{ms}  & \SI{1.52}{ms} & \textcolor{tBlueStrong}{\SI{-6.54}{ms}} & \textcolor{tBlueStrong}{\SI{-81.1}{\%}} \\
      Total                   & \SI{16.24}{ms} & \SI{9.70}{ms} & \textcolor{tBlueStrong}{\SI{-6.54}{ms}} & \textcolor{tBlueStrong}{\SI{-40.3}{\%}} \\
      \bottomrule
    \end{tabularx}
    \label{tab:supp_ablation_runtime}
\end{table}
\subsubsection{Runtime Decomposition.}
To identify which specific operations contribute to the overall runtime, we present a detailed runtime decomposition in \cref{tab:supp_ablation_runtime}. While the CGI$_{\text{3D}}$ patch extraction introduces a minor \SI{0.26}{ms} overhead, it enables a substantial \SI{92}{\%} reduction in the 2D/3D head's computational cost. For this patch extraction, we evaluated a custom CUDA kernel but found that simple indexing proved equally efficient. We therefore adopt the simpler implementation.

\begin{table}[t]
    \centering
    \caption{\textbf{CGI$_{\text{3D}}$ analysis by model size.} Runtime (ms, $\downarrow$) and car mod.\ \APDDDRFS\ (\%, $\uparrow$) on the KITTI~\cite{kitti} \emph{validation} set with and without CGI$_{\text{3D}}$, evaluated across all five model sizes of \ourmethod, extending \cref{tab:ablationd_runtime}.}
    \footnotesize\sisetup{table-number-alignment=center}
    \setlength{\tabcolsep}{2.0pt}
    \begin{tabularx}{\columnwidth}{>{\hspace{-\tabcolsep}\raggedright\columncolor{white}[\tabcolsep][\tabcolsep]}X S[table-format=2.1] S[table-format=2.1] S[table-format=2.1] S[table-format=2.1] S[table-format=2.1] S[table-format=2.1] S[table-format=2.1] S[table-format=2.1] S[table-format=2.1] S[table-format=2.1]}
        \toprule
         & \multicolumn{2}{c}{\textbf{N}} & \multicolumn{2}{c}{\textbf{S}} & \multicolumn{2}{c}{\textbf{M}} & \multicolumn{2}{c}{\textbf{B}} & \multicolumn{2}{c}{\textbf{X}} \\
        \cmidrule(lr){2-3} \cmidrule(lr){4-5} \cmidrule(lr){6-7} \cmidrule(lr){8-9} \cmidrule(lr){10-11}
        \raisebox{3.0pt}[0pt][0pt]{\multirow{-2}{*}{\textbf{CGI$_{\text{3D}}$}}} & {\textbf{Time}} & {\textbf{AP}} & {\textbf{Time}} & {\textbf{AP}} & {\textbf{Time}} & {\textbf{AP}} & {\textbf{Time}} & {\textbf{AP}} & {\textbf{Time}} & {\textbf{AP}} \\
        \specialrule{0.5pt}{1.0pt}{2pt}
        \xmark & 15.5 & 18.4 & 16.0 & 20.7 & 20.1 & 22.3 & 20.9 & 22.7 & 30.9 & 22.9 \\
        \cmark & \best 9.7 & 18.4 & \best 10.2 & 20.7 & \best 13.3 & 22.3 & \best 13.9 & 22.7 & \best 23.6 & 22.9 \\
        \specialrule{0.5pt}{1.5pt}{2pt}
        \textbf{Time} $\Delta$ & {-5.8} & {---} & {-5.8} & {---} & {-6.8} & {---} & {-7.0} & {---} & {-7.3} & {---} \\
        \bottomrule
    \end{tabularx}
    \label{tab:supp_cgi3d_model_sizes}
\end{table}

\subsubsection{CGI$_{\text{3D}}$ Runtime Across Model Sizes.}
\cref{tab:supp_cgi3d_model_sizes} extends \cref{tab:ablationd_runtime} by reporting the runtime and detection accuracy with and without CGI$_{\text{3D}}$ for all five model sizes. CGI$_{\text{3D}}$ consistently reduces runtime by \SIrange{5.8}{7.3}{ms} across all variants while accuracy remains unchanged. The slight increase in savings at larger model sizes is likely due to secondary effects such as reduced GPU memory bandwidth pressure and GPU synchronization overhead when processing sparse rather than dense feature maps.

\subsection{Student \vs Teacher Analysis}
\label{sec:contribution_by_model_size}
\begin{table}[t]
  \centering
    \caption{\textbf{Student \vs Teacher.} We analyze the accuracy of our teacher model against our student models with different sizes (N, S, M, B, and X). For reference, we also report our baseline (YOLOv10-M3D) and improvements over the baseline. We report \APDDDRFS\ Mod. (in \%, $\uparrow$) for the ``Car'' category on the KITTI~\cite{kitti} \emph{val.} set.}
    \small
    \begin{tabularx}{\linewidth}{l@{\hspace{1.85em}}l}
        \toprule
      \textbf{Method} & \textbf{\APDDDRFS\ Mod.\ Car\ $\uparrow$} \\ \midrule

      \textbf{Teacher} \\
      \ourmethodx w/o A2D2 & \cbarm{tRedLight}{16.83em}{0.7em}{20.87} \\
      \midrule
      \textbf{Baselines / Students} \\
      \baselinen & \cbarm{tBlueLight}{11.71em}{0.7em}{14.52} \\
      \ourmethodn & \cbarm{tGreenStrong}{14.83em}{0.7em}{18.38 \textcolor{tBlueStrong}{+3.86}} \\
      \midrulegray
      \baselines & \cbarm{tBlueLight}{14.40em}{0.7em}{17.85} \\
      \ourmethods & \cbarm{tGreenStrong}{16.67em}{0.7em}{20.67 \textcolor{tBlueStrong}{+2.82}} \\
      \midrulegray
      \baselinem & \cbarm{tBlueLight}{15.53em}{0.7em}{19.26} \\
      \ourmethodm & \cbarm{tGreenStrong}{18.07em}{0.7em}{22.40 \textcolor{tBlueStrong}{+3.14}} \\
      \midrulegray
      \baselineb & \cbarm{tBlueLight}{15.81em}{0.7em}{19.60} \\
      \ourmethodb & \cbarm{tGreenStrong}{18.27em}{0.7em}{22.65 \textcolor{tBlueStrong}{+3.05}} \\
      \midrulegray
      \baselinex & \cbarm{tBlueLight}{16.11em}{0.7em}{19.98} \\
      \ourmethodx & \cbarm{tGreenStrong}{18.52em}{0.7em}{22.96 \textcolor{tBlueStrong}{+2.98}} \\

      \bottomrule

    \end{tabularx}
    \label{tab:supp_ablation_val_a2d2}
\end{table}

As introduced in \cref{sec:method:distill}, our distillation strategy always utilizes the \ourmethodx w/o A2D2 model size as the teacher. This ensures that even the smallest students benefit from the high capacity of our largest model. In \cref{tab:supp_ablation_val_a2d2}, we evaluate the effectiveness of this approach by comparing distilled student models against both the baseline and the teacher.

First, we observe a significant improvement over the baseline across all scales. Interestingly, our results show a consistent accuracy gain of approximately \SI{3}{\%} in \APDDDRFS\space across nearly all model sizes. The smallest variant, \ourmethodn, achieves the largest improvement with a gain of \SI{3.86}{\%} in \APDDDRFS. We hypothesize that this smallest model suffers from a significant capacity gap and consequently benefits more substantially from the guidance provided by the large teacher.

Second, we compare the student models directly to the teacher to assess the efficiency of knowledge transfer. Notably, our second-smallest model already achieves a detection accuracy that is almost on par with the teacher's accuracy, while the benefit of \ourmethodx (or largest model size) over its teacher variant is \SI{2.09}{AP}.

Finally, we investigate the effect of teacher size on student accuracy. Using \ourmethodx as teacher, \ourmethodn achieves \SI{18.38}{\%} Mod.\ \APDDDRFS. Reducing the teacher to \ourmethodb\ decreases student accuracy to \SI{18.08}{\%}, confirming that a larger teacher consistently benefits student performance.

\begin{table}[ht!]

    \caption{
        \textbf{Extended KITTI \cite{kitti} test results.} We extend \cref{tab:exp_kittI_test_fast,tab:exp_kitti_test_sota} and report results on the KITTI \cite{kitti} \emph{test} set for the category ``Car'' using \APDDDRFS\ and \APBEVRFS\ (both in \%, $\uparrow$). \emph{Extra} indicates the use of auxiliary training data. \emph{Params} reports no.\ of model parameters in millions. \emph{GFLOPs} measured for single-image inference. \emph{Time} is reported in ms for single-image inference without TensorRT on an NVIDIA RTX 8000 GPU. \noCode indicates models with no public code available. For these models, we report the authors' stated runtime. If code is publicly available, we report runtime on our hardware.
    }
    \setlength{\tabcolsep}{1.4pt}
    \fontsize{6pt}{7pt}\selectfont
    \begin{tabularx}{\textwidth}{>{\hspace{-\tabcolsep}\raggedright\columncolor{white}[\tabcolsep][\tabcolsep]}XlcrrcS[table-format=1.2]S[table-format=1.2]S[table-format=1.2]S[table-format=1.2]S[table-format=1.2]S[table-format=1.2]}
    \toprule
    \multirow{3}{*}{\textbf{Method}} & \multirow{3}{*}{\textbf{Venue}} & \multirow{3}{*}{\textbf{Extra}} & \multirow{3}{*}{\textbf{Par.\,$\downarrow$}} & \multirow{3}{*}{\textbf{FLOP.\,$\downarrow$}} & \multirow{3}{*}{\textbf{Time\,$\downarrow$}} & \multicolumn{3}{c}{\textbf{\APDDDRFS\,$\uparrow$}} & \multicolumn{3}{c}{\textbf{\APBEVRFS\,$\uparrow$}}\\
    \cmidrule(lr){7-9} \cmidrule(lr){10-12}
    & & & & & & \textbf{Easy} & \textbf{Mod.} & \textbf{Hard} & \textbf{Easy} & \textbf{Mod.} & \textbf{Hard}\\ \midrule

    MonoDistill~\cite{monodistill} & {\fontsize{4pt}{5pt}\selectfont ICLR'22} & LiDAR & 50.1 & 159 & 46.7 & 22.97 & 16.03 & 13.60 & 31.87 & 22.59 & 19.72 \\ 
    ADD~\cite{add} & {\fontsize{4pt}{5pt}\selectfont AAAI'23} & LiDAR & {---\;\,} & {---\,} & ---\noCode & 25.61 & 16.81 & 13.79 & 35.20 & 23.58 & 20.08\\ 
    MonoNeRD~\cite{mononerd} & {\fontsize{4pt}{5pt}\selectfont CVPR'23} & LiDAR & \second 6.6 & 4220 & 1380.3 & 22.75 & 17.13 & 15.63 & 31.13 & 23.46 & 20.97 \\
    HSRDN~\cite{hsrdn} & {\fontsize{4pt}{5pt}\selectfont TIV'24} & LiDAR & {---\;\,} & {---\,} & ---\noCode & 26.37 & 16.39 & 14.22 & 36.49 & 23.96 & 19.67\\ 
    MonoFG~\cite{monofg} & {\fontsize{4pt}{5pt}\selectfont ACM'24} & LiDAR &  {---\;\,} & {---\,} & 56\noCode\; & 24.35 & 16.46 & 13.84 & 32.42 & 21.64 & 18.60\\ 
    MonoSTL~\cite{monostl} & {\fontsize{4pt}{5pt}\selectfont TCSVT'24} & LiDAR & 50.0 & 159 & 25.7 & 25.33 & 16.13 & 13.35 & 33.87 & 22.42 & 19.04\\ 
    MonoDSSMs-A & {\fontsize{4pt}{5pt}\selectfont ACCV'24} & LiDAR & 23.6 & {---\,} & 28.8\noCode & 21.47 & 14.55 & 11.78 & 28.84 & 19.54 & 16.30 \\
    MonoSGC~\cite{monosgc} & {\fontsize{4pt}{5pt}\selectfont TIV'24} & LiDAR & 23.1 & 173 & 35.0 & 27.01 & 16.77 & 14.61 & 35.78 & 23.27 & 19.92\\
    OccupancyM3D~\cite{occupancym3d} & {\fontsize{4pt}{5pt}\selectfont CVPR'24} & LiDAR & 28.3 & 389 & 213.9 & 25.55 & 17.02 & 14.79 & 35.38 & 24.18 & 21.37\\  
    DPL$_{FLEX}$~\cite{dpl_flex} & {\fontsize{4pt}{5pt}\selectfont CVPR'24} & Unlabeled & 21.5 & 152 & 28.9 & 24.19 & 16.67 & 13.83 & 33.16 & 22.12 & 18.74 \\ 
    MonoSG~\cite{monosg} & {\fontsize{4pt}{5pt}\selectfont RAL'25} & Stereo & {---\;\,} & {---\,} & 36\noCode\, & 25.77 & 16.70 & 14.22 & 33.46 & 22.12 & 19.16 \\
    DK3D~\cite{dk3d} & {\fontsize{4pt}{5pt}\selectfont TPAMI'25} & LiDAR & 19.6 & {---\,} & 30.1\noCode & 25.63 & 16.82 & 13.81 & 35.23 & 23.59 & 20.10 \\
    MonoTAKD~\cite{monotakd} & {\fontsize{4pt}{5pt}\selectfont CVPR'25} & LiDAR & 46.9 & 810 & 270.1 & 27.91 & 19.43 & 16.51 & 38.75 & 27.76 & 24.14 \\  

    \midrule
    HomoLoss~\cite{homoflex} & {\fontsize{4pt}{5pt}\selectfont CVPR'22} & Geom. & 21.5 & 152 & 28.9 & 21.75 & 14.94 & 13.07 & 29.60 & 20.68 & 17.81\\
    MonoDDE~\cite{monodde} & {\fontsize{4pt}{5pt}\selectfont CVPR'22} & Geom. & {---\;\,} & {---\,} & 36\noCode\, & 25.53 & 16.59 & 14.53 & 33.41 & 22.81 & 19.57 \\
    PDR~\cite{pdr} & {\fontsize{4pt}{5pt}\selectfont TCSVT'23} & Geom. & {---\;\,} & {---\,} & 29\noCode\, & 23.69 & 16.14 & 13.78 & {---\;\,} & {---\;\,} & {---\;\,} \\
    MonoATT~\cite{monoatt} & {\fontsize{4pt}{5pt}\selectfont CVPR'23} & Geom. & {---\;\,} & {---\,} & 56\noCode\, & 24.72 & 17.37 & 15.00 &  36.87 & 24.42 & 21.88\\
    MonoUNI~\cite{monouni} & {\fontsize{4pt}{5pt}\selectfont NeurIPS'23} & Geom. & 22.7 & 122 & 23.2 & 24.75 & 16.73 & 13.49 & 33.28 & 23.05 & 16.39\\ 
    MonoCD~\cite{monocd} & {\fontsize{4pt}{5pt}\selectfont CVPR'24} & Geom. & 21.8 & 171 & 28.1 & 25.53 & 16.59 & 14.53 & 33.41 & 22.81 & 19.57 \\
    MoGDE~\cite{MoGDE} & {\fontsize{4pt}{5pt}\selectfont TPAMI'25} & Geom. & {---\;\,} & {---\,} & ---\noCode & 27.25 & 17.93 & 15.80 & 38.84 & 26.02 & 23.27 \\
    MonoDGP~\cite{monodgp} & {\fontsize{4pt}{5pt}\selectfont CVPR'25} & Geom. & 43.3 & 276 & 68.1 & 26.35 & 18.72 & 15.97 & 35.24 & 25.23 & 22.02 \\

    \midrule

    MonoCon~\cite{monocon} & {\fontsize{4pt}{5pt}\selectfont AAAI'22} & {---} & 19.6 & 115 & 15.5 & 22.50 & 16.46 & 13.95 & 31.12 & 22.10 & 19.00\\
    Cube R-CNN~\cite{brazil2023omni3d} & {\fontsize{4pt}{5pt}\selectfont CVPR'23} & {---} & 47.2 & 142 & 27.3 & 23.59 & 15.01 & 12.56 & 31.70 & 21.20 & 18.43 \\
    MonoDETR~\cite{monodetr} & {\fontsize{4pt}{5pt}\selectfont ICCV 23} & {---} & 37.7 & 119 & 47.8 & 25.00 & 16.47 & 16.38 & 33.60 & 22.11 & 18.60\\
    DDML~\cite{depth-discriminative} & {\fontsize{4pt}{5pt}\selectfont NeurIPS'23} & {---} & 19.6 & 115 & 15.5 & 23.31 & 16.36 & 13.73 & {---\;\,} & {---\;\,} & {---\;\,} \\
    MonoPSTR~\cite{monopstr} & {\fontsize{4pt}{5pt}\selectfont TIM'24} & {---} & 38.0 & 96 & 35.0 & 26.15 & 17.01 & 13.70 & 34.79 & 22.88 & 19.40 \\ 
    FD3D~\cite{fd3d} & {\fontsize{4pt}{5pt}\selectfont AAAI'24} & {---} & {---\;\,} & {---\,} & 40\noCode\, &  25.38 & 17.12 & 14.50 & 34.20 & 23.72 & 20.76 \\ 
    MonoLSS~\cite{monolss} & {\fontsize{4pt}{5pt}\selectfont 3DV'24} & {---} & 21.5 & 127 & 20.2 & 26.11 & 19.15 & 16.94 & 34.89 & 25.95 & 22.59\\
    MonoDiff~\cite{monodiff} & {\fontsize{4pt}{5pt}\selectfont CVPR'24} & {---} & {---\;\,} & {---\,}& 86\noCode\, & {30.18} & {21.02} & 18.16 & {---\;\,} & {---\;\,} & {---\;\,}\\
    MonoMAE~\cite{monomae} & {\fontsize{4pt}{5pt}\selectfont NeurIPS'24} & {---} & {---\;\,} & {---\,} & 38\noCode\, & 25.60 & 18.84 & 16.78 & 34.14 & 24.93 & 21.76\\
    GATE3D~\cite{gate3d} & {\fontsize{4pt}{5pt}\selectfont CVPRW'25} & {---} & {---\;\,} & {---\,} & 82\noCode\, & 26.07 & 18.85 & 16.76 & 33.94 & 25.06 & 22.04 \\
    IDEAL-M3D~\cite{ideal-m3d} & {\fontsize{4pt}{5pt}\selectfont WACV'26} & {---} & 21.5 & 127 & 20.2 & 27.06 & 18.87 & 16.73 & 35.33 & 25.44 & 22.25\\
    MonoA$^2$\cite{monoa2} & {\fontsize{4pt}{5pt}\selectfont PR'26} & {---} & {---\;\,} & {---\,} & 30\noCode\, & 23.24 & 17.55 & 15.26 & 31.71 & 23.14 & 20.45 \\
    \midrule

    \ourmethodn & {\fontsize{4pt}{5pt}\selectfont (Ours)} & {---} & \best 3.8 & \best 14 & \best 9.7 & 24.31 & 16.49 & 14.14 & 32.22 & 21.72 & 19.41 \\
    \ourmethods & {\fontsize{4pt}{5pt}\selectfont (Ours)} & {---} & 10.1 & \second 38 & \second 10.2 & 27.28 & 18.87 & 16.37 & 34.86 & 24.17 & 21.32 \\
    \ourmethodm &  {\fontsize{4pt}{5pt}\selectfont (Ours)} & {---}  & 19.7 & 88 & 13.3 & 28.08 & 19.47 & 17.66 & 36.21 & 25.46 & 22.89 \\
    \ourmethodb &  {\fontsize{4pt}{5pt}\selectfont (Ours)} & {---} & 24.9 & 133 & 13.9 & \second 29.10 & \second 20.17 & \second 18.34 & \second 37.65 & \best 26.63 & \best 23.75\\
    \ourmethodx &  {\fontsize{4pt}{5pt}\selectfont (Ours)} & {---} & 36.3 & 218 & 23.6 & \best 30.76 & \best 21.20 & \best 18.76 & \best 38.33 &  \second 26.57 & \second 23.74 \\
    \bottomrule

    \end{tabularx}
    \label{tab:exp_kitti_test_full}
\end{table}

\subsection{Comparison with State-of-the-Art Methods}
\label{sec:comparison_sota_supp}
In this section, we provide the full evaluation tables for our approach across all benchmarks, due to constraints in the main paper. We detail our accuracy on the KITTI\cite{kitti} test (including ``Pedestrian'' and ``Cyclist'' categories) and validation sets, provide full metric reports for Rope3D\cite{rope3d} and Waymo\cite{waymo}, discuss the impact of different training protocols on Waymo, and provide cross-dataset generalization results on nuScenes~\cite{nuScenes} (\cref{sec:supp_nuscenes_cross_dataset}).

\subsubsection{KITTI~\cite{kitti} Test Set for Cars.}
While \cref{tab:exp_kittI_test_fast,tab:exp_kitti_test_sota} of the main paper compares \ourmethod against the most critical baselines, \cref{tab:exp_kitti_test_full} provides an extended overview of results for the ``Car'' category, including runtime, GFLOPs, parameters, and BEV metrics.
\ourmethod\ obtains the highest score on the primary metric, \APDDDRFS, demonstrating strong detection accuracy. In contrast, MonoTAKD~\cite{monotakd} reports higher numbers on \APBEVRFS. We attribute this to the inherent design difference: \APBEVRFS\space is tightly coupled to raw depth quality, a task where MonoTAKD gains a significant advantage by being trained with additional dense LiDAR depth supervision. This extra data modality simplifies depth prediction and inflates depth-error sensitive metrics. More importantly, MonoTAKD's dependency on compute-intensive 3D volume processing results in an inference speed that is more than an order of magnitude slower than ours. As shown in \cref{tab:kitti_test_set_efficientnetv2}, when utilizing a different backbone, such as EfficientNetv2, \ourmethod\ outperforms MonoTAKD~\cite{monotakd} in five out of six metrics while maintaining a runtime over \num{5}$\times$ faster and training purely on image data.

\begin{table}[t!]
    \caption{
        \textbf{Detailed KITTI \cite{kitti} test results.} We extend \cref{tab:exp_kitti_test_full} and report \APDDDRFS\ for the ``Car'', ``Pedestrian'', and the ``Cyclist'' category (all in \%, $\uparrow$). We also report \APDDDRFS\ (in \%, $\uparrow$) averaged over categories. All \APDDDRFS\ scores are decomposed into easy, moderate, and hard objects, following the standard protocol. \emph{Extra} indicates the use of auxiliary training data.
    }

    \setlength{\tabcolsep}{1.2pt}
    \fontsize{4pt}{5pt}\selectfont
    \centering
     \begin{tabularx}{\textwidth}{>{\hspace{-\tabcolsep}\raggedright\columncolor{white}[\tabcolsep][\tabcolsep]}XcS[table-format=2.2]S[table-format=2.2]S[table-format=2.2]S[table-format=2.2]S[table-format=2.2]S[table-format=2.2]S[table-format=2.2]S[table-format=2.2]S[table-format=2.2]S[table-format=2.2]S[table-format=2.2]S[table-format=2.2]}
    \toprule
    \multirow{3}{*}{\textbf{Method}} & \multirow{3}{*}{\textbf{Extra}} & 
    \multicolumn{3}{c}{\textbf{\text{Car.} \APDDDRFS\,$\uparrow$}} & 
    \multicolumn{3}{c}{\textbf{\text{Ped.}  \APDDDRFF\,$\uparrow$}} &
    \multicolumn{3}{c}{\textbf{\text{Cycl.} \APDDDRFF\,$\uparrow$}} & 
    \multicolumn{3}{c}{\textbf{\text{Avg.} \APDDDRF\,$\uparrow$}} \\  
    \cmidrule(lr){3-5} \cmidrule(lr){6-8} \cmidrule(lr){9-11} \cmidrule(lr){12-14}  
    
    & & {\textbf{Easy}} & {\textbf{Mod.}} & {\textbf{Hard}} & {\textbf{Easy}} & {\textbf{Mod.}} & {\textbf{Hard}} & {\textbf{Easy}} & {\textbf{Mod.}} & {\textbf{Hard}} & {\textbf{Easy}} & {\textbf{Mod.}} & {\textbf{Hard}} \\ \midrule 

    CMKD \cite{cmkd} & LiDAR & 25.09 & 16.99 & 15.30 & 17.79 & 11.69 & 10.09 & 9.60 & 5.24 & 4.50 & 17.49 & 11.37 & 9.96\\
    MonoRUn \cite{monorun} & LiDAR & 19.65 & 12.30 & 10.58 & 10.88 & 6.78 & 5.83 & 1.01 & 0.61 & 0.48 & 10.51 & 6.56 & 5.63\\
    CaDDN \cite{CaDDN} & LiDAR & 19.17 & 13.41 & 11.46 & 12.87 & 8.14 & 6.76 & 7.00 & 3.41 & 3.30 & 13.01 & 8.32 & 7.17 \\
    DD3D \cite{dd3d} & LiDAR & 23.22 & 16.34 & 14.20 & 13.91 & 9.30 & 8.05 & 2.39 & 1.52 & 1.31 & 13.17 & 8.64 & 7.85\\
    AutoShape \cite{autoshape} & Shapes & 22.47 & 14.17 & 11.36 & 5.46 & 3.74 & 3.03 & 5.99 & 3.06 & 2.70 & 11.31 & 6.99 & 5.70\\
    MonoEF \cite{monoef} & Odometry &  21.29 & 13.87 & 11.71 & 4.27 & 2.79 & 2.21 & 1.80 & 0.92 & 0.71 & 9.12 & 5.86 & 4.88\\
    MonoDistill \cite{monodistill} & LiDAR & 24.31 & 18.47 & 15.76 & 12.79 & 8.17 & 3.39 & 5.53 & 2.81 & 2.40 & 14.21 & 9.82 & 7.18\\
    MonoJSG \cite{monojsg} & LiDAR & 24.69 & 16.14 & 13.64 & 11.02 & 7.49 & 6.41 & 5.45 & 3.21 & 2.57 & 13.72 & 8.95 & 7.54\\
    DCD \cite{dcd} & Shapes & 23.81 & 15.90 & 13.21 & 10.37 & 6.73 & 6.28 & 4.72 & 2.74 & 2.41 & 12.97 & 8.31 & 7.30\\
    Pseudo-Stereo \cite{pseudo_stereo} & LiDAR & 23.74 & 17.74 & 15.14 & 16.95 & 10.82 & 9.26 & 11.22 & 6.18 & 5.21 & 17.30 & 11.58 & 9.87 \\
    MonoPGC \cite{monopgc} & LiDAR & 24.68 & 17.17 & 14.14 & 14.16 & 9.67 & 8.26 & 5.88 & 3.30 & 2.85 & 14.91 & 10.05 & 8.42 \\
    MonoNeRD \cite{mononerd} & LiDAR & 22.75 & 17.13 & 15.63 &  13.20 & 8.26 & 7.02 & 4.79 & 2.48 & 2.16 & 13.58 & 8.88 & 8.27\\
    OPA-3D \cite{opa-3d} & LiDAR & 24.60 & 17.05 & 14.25 & 16.64 & 11.04 & 9.38 & 7.52 & 4.79 & 4.22 & 16.25 & 11.23 & 9.28\\
    HSRDN \cite{hsrdn} & LiDAR & 26.37 & 16.39 & 14.22 & 14.35 & 9.90 & 8.71 & 13.29 & 6.85 & 6.24 & 18.00 & 11.04 & 9.72 \\
    OccupancyM3D \cite{occupancym3d} & LiDAR & 25.55 & 17.02 & 14.79 & 14.68 & 9.15 & 7.80 & 7.37 & 3.56 & 2.84 & 15.87 & 9.91 & 8.48\\
    DPL$_{FLEX}$ \cite{dpl_flex} & Unlabeled & 24.19 & 16.67 & 13.83 & 11.66 & 7.52 & 6.16 & 8.41 & 4.51 & 3.59 & 14.75 & 9.57 & 7.86 \\
    MonOri \cite{monori} & LiDAR & 25.20 & 16.77 & 14.45 & 18.97 & 12.76 & 11.00 & 9.47 & 5.87 & 5.35 & 17.88 & 11.80 & 10.27 \\
    MonoSG \cite{monosg} & Stereo & 25.77 & 16.70 & 14.22 & 14.72 & 9.53 & 7.87 & 4.18 & 2.65 & 2.09 & 16.12 & 9.62 & 8.06 \\
    MonoTAKD \cite{monotakd} & LiDAR & 27.91 & 19.43 & 16.51 & 16.15 & 10.41 & 9.68 & 13.54 & 7.23 & 6.86 & 19.20 & 12.36 & 11.01 \\
    \midrule

    MonoPair \cite{monopair} & Geometry & 13.04 &  9.99 & 8.65 & 10.02 & 6.68 & 5.53 & 3.79 & 2.12 & 1.83 & 8.95 & 6.26 & 4.79\\
    GUPNet \cite{gupnet} & Geometry & 20.11 & 14.20 & 11.77 & 14.95 & 9.76 & 8.41 & 5.58 & 3.21 & 2.66 & 13.55 & 9.06 & 7.61\\
    MonoFlex \cite{monoflex} & Geometry & 19.94 & 13.89 & 12.07 & 9.43 & 6.31 & 5.26 & 4.17 & 2.35 & 2.04 & 11.18 & 7.52 & 6.46\\
    MonoDDE \cite{monodde} & Geometry & 24.93 & 17.14 & 15.10 & 11.13 & 7.32 & 6.67 & 5.94 & 3.78 & 3.33 & 13.45 & 9.41 & 8.37\\
    DEVIANT \cite{deviant} & Geometry & 21.88 & 14.46 & 11.89 & 13.43 & 8.65 & 7.69 & 5.05 & 3.13 & 2.59 & 14.45 & 8.75 & 7.39 \\
    HomoLoss \cite{homoflex} & Geometry & 21.75 & 14.94 & 13.07 & 11.87 & 7.66 & 6.82 & 5.48 & 3.50 & 2.99 & 13.03 & 8.70 & 7.62\\
    YOLOBU \cite{yolobu} & Geometry & 22.43 & 16.21 & 13.73 & 11.68 & 7.58 & 6.22 & 5.25 & 2.83 & 2.31 & 13.12 & 8.87 & 7.42\\
    MonoRCNN++ \cite{monorcnn++} & Geometry & 20.08 & 13.72 & 11.34 & 12.26 & 7.90 & 6.62 & 3.17 & 1.81 & 1.75 & 11.84 & 7.81 & 6.88\\
    SSD-MonoDETR \cite{ssd-monodetr} & Geometry & 24.52 & 17.88 & 15.69 & 12.64 & 9.88 & 8.58 & 7.79 &  5.76 & 4.33 & 14.98 & 11.17 & 9.53 \\
    PDR \cite{pdr} & Geometry & 23.69 & 16.14 & 13.78 & 11.61 & 7.72 & 6.40 & 2.72 & 1.57 & 1.50 &   12.67 & 9.77 & 7.23 \\
    MonoATT \cite{monoatt} & Geometry & 24.72 & 17.37 & 15.00 & 13.20 & 8.26 & 7.02 & 4.79 & 2.48 & 2.16 & 14.23 & 9.37 & 8.06 \\
    MonoUNI \cite{monouni} & Geometry & 24.75 & 16.73 & 13.49 & 15.78 & 10.34 & 8.74 & 7.34 & 4.28 & 3.78 & 15.96 & 10.45 & 8.67 \\ 
    MoGDE \cite{MoGDE} & Geometry & 27.25 & 17.93 & 15.80 & 11.27 & 8.33 & 7.67 & 7.02 & 3.96 & 3.41 & 15.18 & 10.07 & 8.96 \\ 
    MonoDGP \cite{monodgp} & Geometry & 26.35 & 18.72 & 15.97 & 15.04 & 9.89 & 8.38 & 5.28 & 2.82 & 2.65 & 15.56 & 10.48 & 9.00 \\ 
    \midrule

    MonoCon \cite{monocon} & {---} & 22.50 & 16.46 & 13.95 & 13.10 & 8.41 & 6.94 & 2.80 & 1.92 & 1.55 & 12.80 & 8.93 & 7.33\\
    MonoDTR \cite{monodtr} & {---} & 24.52 & 18.57 & 15.51 & 15.33 & 10.18 & 8.61 & 5.05 & 3.27 & 3.19 & 14.97 & 10.67 & 9.10\\
    Cube R-CNN \cite{brazil2023omni3d} & {---} & 23.59 & 15.01 & 12.56 &  11.17 & 6.95 & 5.87 & 3.65 & 2.67 & 2.28 & 12.80 & 8.21 & 6.90\\
    DDML \cite{depth-discriminative} & {---} & 23.31 & 16.36 & 13.73 & 14.90 & 10.28 & 8.70 & 5.38 & 2.89 & 2.83 & 14.53 & 9.84 & 8.42\\
    MonoPSTR \cite{monopstr} & {---} & 26.15 & 17.01 & 13.70 & 13.26 & 8.84 & 6.99 & 7.11 & 4.75 & 4.21 & 15.51 & 10.20 & 8.30 \\
    MonoLSS \cite{monolss} & {---} & 26.11 & 19.15 & 16.94 & \best 17.09 & \best 11.27 & \best 10.00 & 7.23 & 4.34 & 3.92 & 16.81 & 11.59 & 10.29\\ 
    MonoDiff \cite{monodiff} & {---} & \second 30.18 & \second 21.02 & 18.16 & 13.51 & 8.94 & 7.28 & 8.52 & \second 5.55 & 4.35 & 17.40 & 11.84 & 9.93  \\ 
    MonoMM \cite{monomm} & {---} & 21.13 & 15.67 & 12.97 & 14.86 & 9.95 & 8.34 & 6.82 & 3.82 & 3.75 & 14.27 & 9.81 & 8.35 \\
    GATE3D \cite{gate3d} & {---} & 26.07 & 18.85 & 16.76 & 16.25 & 10.53 & 8.91 & {---} & {---} & {---} & {---} & {---} & {---} \\
    MonoA$^2$ \cite{monoa2} & {---} & 23.24 & 17.55 & 15.26 & 12.95 & 8.51 & 7.56 & 4.39 & 2.28 & 2.31 & 13.53 & 9.45 & 8.38 \\
    IDEAL-M3D \cite{ideal-m3d} & {---} & 27.06 & 18.87 & 16.73 & 13.73 & 8.50 & 7.52 & 6.93 & 4.12 & 3.71 & 15.91 & 10.50 & 9.32 \\
    \midrule

    \ourmethodn (Ours) & {---} & 24.31 & 16.49 & 14.14 &  12.78 & 8.18 & 6.80 & 6.11 & 3.34 & 3.10 & 14.40 & 9.34 & 8.01\\
    \ourmethods (Ours) & {---} & 27.28 & 18.87 & 16.37 & 15.91 & 9.95 & 8.44 & 4.61 & 2.38 & 2.40 & 15.93 & 10.40 & 9.07\\
    \ourmethodm (Ours) & {---} & 28.08 & 19.47 & 17.66 & 15.91 & 10.15 & 8.62 & 8.95 & 5.02 & 4.53 & 17.65 & 11.55 & 10.27 \\
    \ourmethodb (Ours) & {---} & 29.10 & 20.17 & \second 18.34 & 16.03 & 10.47 & 8.91 & \best 10.11 & \best 6.05 & \best 5.11 & \second 18.44 & \second 12.23 & \second 10.79 \\
    \ourmethodx (Ours) & {---} & \best 30.76 & \best 21.20 & \best 18.76 & \second 16.68 & \second 10.98 & \second 9.20 & \second 9.28 & 5.48 & \second 4.70 & \best 18.91 & \best 12.55 & \best 10.89 \\
    \bottomrule
    \end{tabularx}
    \label{tab:exp_kitti_test_car_ped_cyc_avg}
\end{table}
\begin{table}[t!]

    \caption{\textbf{Extended KITTI \cite{kitti} validation results.} We extend the test results in \cref{tab:exp_kittI_test_fast,tab:exp_kitti_test_sota} with \emph{validation} results on KITTI. We report \APDDDRFS\ (in \%, $\uparrow$). \emph{Time} is reported in ms for single-image inference without TensorRT on an NVIDIA RTX 8000 GPU. \emph{Extra} indicates the use of auxiliary training data. \noCode indicates models with no public code available. For these models, we report the authors' stated runtime. If code is publicly available, we report runtime on our hardware.}
    
    \fontsize{6pt}{7pt}\selectfont
    \setlength{\tabcolsep}{5.65pt}
    \begin{tabularx}{\columnwidth}{>{\hspace{-\tabcolsep}\raggedright\columncolor{white}[\tabcolsep][\tabcolsep]}XcrS[table-format=2.2]}
    \toprule
    \textbf{Method} & \textbf{Extra} & \textbf{Time\ $\downarrow$} & \textbf{\APDDDRFS\ $\uparrow$}\\ \midrule
    MonoDistill \cite{monodistill} & LiDAR & 46.7 & 16.03 \\ 
    ADD \cite{add} & LiDAR & ---\noCode  & 16.81\\ 
    MonoNeRD \cite{mononerd} & LiDAR & 1380.3 & 19.96 \\
    HSRDN \cite{hsrdn} & LiDAR & ---\noCode & 13.61\\ 
    MonoFG \cite{monofg} & LiDAR & 56\noCode\; & 16.46\\ 
    MonoSTL \cite{monostl} & LiDAR & 25.7 & 17.14\\ 
    MonoSGC \cite{monosgc} & LiDAR & 35.0 & 19.55\\
    OccupancyM3D \cite{occupancym3d} & LiDAR & 213.9 & 19.96\\  
    DPL$_\mathit{FLEX}$ \cite{dpl_flex} & Unlabeled & 28.9 & 19.84\\ 
    MonoSG \cite{monosg} & Stereo & 36\noCode\; & 20.77\\
    MonoTAKD \cite{monotakd} & LiDAR & 270.1 & 22.61\\  

    \midrule
    HomoLoss \cite{homoflex} & Geometry & 28.9\noCode & 16.89\\
    MonoDDE \cite{monodde} & Geometry & 36\noCode\; & 19.75\\
    MonoATT \cite{monoatt} & Geometry & 56\noCode\; & 22.47 \\
    MonoUNI \cite{monouni} & Geometry & 23.2 & 16.73\\ 
    MonoCD \cite{monocd} & Geometry & 28.1 & 19.37\\
    MoGDE \cite{MoGDE} & Geometry & ---\noCode & 20.35\\
    MonoDGP \cite{monodgp} & Geometry & 68.1 & 22.34\\ 

    \midrule

    MonoCon \cite{monocon} & --- & 15.5 & 19.01\\
    MonoDETR \cite{monodetr} & --- & 47.8 & 16.47\\
    DDML \cite{depth-discriminative} & --- & 15.5 & 19.43 \\
    MonoPSTR \cite{monopstr} & --- & 35.0 & 17.01\\ 
    FD3D \cite{fd3d} & --- & 40\noCode\; & 20.23\\ 
    MonoLSS \cite{monolss} & --- & 20.2 & 18.29\\
    MonoDiff \cite{monodiff} & --- & 86\noCode\; & 22.02 \\
    MonoMAE \cite{monomae} & --- & 38\noCode\; & 20.90\\
    MonoA$^2$ \cite{monoa2} & --- & 30\noCode\; & 21.04 \\
    \midrule

    \baselinen (Baseline) & --- & 16.2 & 14.52 \\
    \ourmethodn (Ours) & --- & \best 8.2 & 18.38 \\
    \midrulegray
    \baselines (Baseline) & --- & 16.7 & 17.85 \\
    \ourmethods  (Ours) & --- & \second \underline{10.9} & 20.67\\
    \midrulegray
    \baselinem (Baseline) & --- & 21.0 & 19.26 \\
    \ourmethodm   (Ours) & ---  & 12.9 & 22.40\\
    \midrulegray
    \baselineb (Baseline) & --- & 21.5 & 19.60 \\
    \ourmethodb   (Ours) & --- & 14.1 & \second \underline {22.65}\\
    \midrulegray
    \baselinex (Baseline) & --- & 31.7 & 19.98 \\
    \ourmethodx   (Ours) & --- & 24.3 & \best 22.96\\

    \bottomrule

    \end{tabularx}
    \label{tab:exp_kitti_val}
\end{table}

\subsubsection{KITTI~\cite{kitti} Test Set for All Classes.}
Whereas the main paper focuses on the ``Car'' category, we here extend our evaluation to the ``Pedestrian'' and ``Cyclist'' categories.
\Cref{tab:exp_kitti_test_car_ped_cyc_avg} reports the precision for all three categories as well as the average precision.
Our two largest models achieve the highest average precision among methods that do not use extra training data.

While MonoLSS~\cite{monolss} achieves higher precision on the ``Pedestrian'' category, we hypothesize that this is due to their Learnable Sample Selection (LSS) module. This module lets the model shift the focus of the heads to meaningful areas of the feature map, which might be particularly relevant for the elongated bounding boxes of the ``Pedestrian'' category. However, its overall average precision is still significantly lower than ours.

Among methods with extra data, only MonoTAKD~\cite{monotakd} surpasses our average accuracy.
While our car and pedestrian accuracy is higher, MonoTAKD achieves a higher cyclist accuracy.
We hypothesize that this is due to additional LiDAR supervision, which provides denser training signals for the ``Cyclist'' category.
This category is extremely underrepresented in KITTI, making extra supervision more valuable.

\subsubsection{KITTI~\cite{kitti} Validation Set.}
While we utilize the KITTI \emph{test} set as our primary benchmark in the main paper (\cf \cref{tab:exp_kittI_test_fast,tab:exp_kitti_test_sota}), as it requires a formal server submission without accessible labels, we provide an additional comparison on the \emph{validation} set here to further validate our findings. 
\cref{tab:exp_kitti_val} compares \ourmethod with other methods on the KITTI~\cite{kitti} validation set.
\ourmethodx achieves the highest accuracy among all variants and \ourmethod offers the best accuracy-to-runtime trade-off.
We also observe consistent gains across all model sizes compared to our baseline, confirming the effectiveness of our proposed components.

\begin{table}[t!]
    \centering

     \caption{\textbf{Extended Rope3D~\cite{rope3d} results.} We extend \cref{tab:exp_rope3d_light} and compare with with state-of-the-art M3D methods on the Rope3D heterlog.\ benchmark. We report AP and Rope score (both in \%) for the ``Big Vehicle'' and ``Car'' class using different IoU thresholds. \emph{Extra} indicates additional data used (\eg, ground-truth ground-plane input).}
    \fontsize{6pt}{7pt}\selectfont
    \centering
    \setlength{\tabcolsep}{1.1pt}
    \begin{tabularx}{\linewidth}{
        lc
        *{8}{S[table-format=2.2]}
        }
    
      \toprule
      \multirow{4}{*}{\textbf{Method}} & \multirow{4}{*}{\textbf{Extra}} & 
      \multicolumn{4}{c}{\textbf{IoU = 0.5\,$\uparrow$}} & \multicolumn{4}{c}{\textbf{IoU = 0.7\,$\uparrow$}}\\
      \cmidrule(lr){3-6} \cmidrule(lr){7-10}
      & & \multicolumn{2}{c}{\textbf{Car}} & \multicolumn{2}{c}{\textbf{Big Vehicle}} & \multicolumn{2}{c}{\textbf{Car}} & \multicolumn{2}{c}{\textbf{Big Vehicle}}\\
      \cmidrule(lr){3-4} \cmidrule(lr){5-6} \cmidrule(lr){7-8} \cmidrule(lr){9-10} 
      & & \textbf{AP} & \textbf{Rope} & \textbf{AP} & \textbf{Rope} & \textbf{AP} & \textbf{Rope} & \textbf{AP} & \textbf{Rope} \\ \midrule
      M3D-RPN~\cite{m3drpn} & Ground plane & 36.33 & 48.16 & 24.39 & 37.81 & 11.09 & 28.17 & 3.39 & 21.01 \\
      MonoDLE~\cite{monodle} & Ground plane & 31.33 & 43.68 & 23.81 & 36.21 & 12.16 & 28.39 & 3.02 & 19.96 \\
      MonoFlex~\cite{monoflex} & Ground plane & 37.27 & 48.58 & \best 47.52 & \best 55.86 & 11.24 & 27.79 & \best 13.10 & \best 28.22 \\
      BEVHeight~\cite{bevheight} & Ground plane & 29.65 & 42.48 & 13.13 & 28.08 & 5.41 & 23.09 & 1.16 & 18.53 \\
      CoBEV~\cite{cobev} & Ground plane, Geometry & 31.25 & 43.74 & 16.11 & 30.73 & 6.59 & 24.01 & 2.26 & 19.71 \\ 
      MOSE ~\cite{mose} & Ground plane, Geometry & 25.62 & {---} & 11.04 & {---}  & {---}  & {---}  & {---}  & {---}  \\ \midrule

      M3D-RPN~\cite{m3drpn} & {---} &  21.74 & 36.40 & 21.49 & 35.49 & 6.05 & 23.84 & 2.78 & 20.82 \\
      Kinematic3D \cite{kinematic3d} & {---} & 23.56 & 37.05 & 13.85 & 28.58 & 5.82 & 23.06 & 1.27 & 18.92 \\
      MonoDLE~\cite{monodle} & {---} & 19.08 & 33.72 & 19.76 & 33.07 & 3.77 & 21.42 & 2.31 & 19.55 \\
      MonoFlex \cite{monoflex} & {---} & 32.01 & 44.37 & 13.86 & 28.47 & 10.86 & 27.39 & 0.97 & 18.18 \\
      BEVFormer~\cite{bevformer} & {---} & 25.98 & 39.51 & 8.81 & 24.67 & 3.87 & 21.84 & 0.84 & 18.42 \\
      BEVDepth~\cite{bevdepth} & {---} & 9.00 & 25.80 & 3.59 & 20.39 & 0.85 & 19.38 & 0.30 & 17.84 \\
      MonoCon \cite{monocon} & {---} & 38.07 & 49.44 & 18.66 & 32.89 & 10.71 & 27.55 & 1.61 & 19.25 \\ 
      GroundMix \cite{cdrone} & {---} & \second 47.72 & \second 57.26 & 32.12 & 43.64 & 12.86 & 29.37 & 3.90 & 21.06 \\ \midrule

      \baselinen (Baseline) & {---} & 27.67 & 41.05 & 14.12 & 30.21 & 8.57 & 24.72 & 0.92 & 18.60 \\
      \ourmethodn (Ours) & {---} & 38.00 & 49.39 & 17.88 & 33.29 & 9.67 & 25.46 & 1.80 & 19.17\\
      \midrulegray

      \baselines (Baseline) & {---} & 34.78 & 46.87 & 21.62 & 36.34 & 9.01 & 25.19 & 2.12 & 19.67 \\
      \ourmethods (Ours) & {---} & 43.52 & 53.89 & 26.78 & 40.50 & 13.33 & 28.59 & 4.16 & 21.25 \\
      \midrulegray

      \baselinem (Baseline) & {---} & 43.23 & 53.68 & 30.65 & 43.62 & 11.88 & 27.53 & 3.88 & 21.12\\
      \ourmethodm (Ours) & {---} & 45.92 & 55.84 & 34.69 & 46.85 & 14.31 & 29.62 & 4.95 & 22.14 \\
      \midrulegray

      \baselineb (Baseline) & {---} & 42.67 & 53.25 & 32.45 & 45.07 & 12.55 & 28.15 & 4.65 & 21.83 \\
      \ourmethodb (Ours) & {---} & 46.30 & 56.14 & 34.75 & 46.90 & \second 15.05 & \second 30.13 & 5.40 & 22.41 \\
      \midrulegray

      \baselinex (Baseline) & {---} & 44.06 & 54.38 & 31.31 & 44.18 & 13.87 & 29.33 & 4.12 & 21.54 \\
      \ourmethodx (Ours) & {---} & \best 49.50 & \best 58.72 & \second 37.74 & \second 49.31 & \best 16.45 & \best 31.34 & \second 8.71 & \second 25.15 \\

      \bottomrule
    \end{tabularx}

    \label{tab:exp_rope3d_full}
\end{table}

\subsubsection{Rope3D~\cite{rope3d}.}
In \cref{tab:exp_rope3d_light} of the main paper, we provide a summary of Rope3D results. Here, we present the complete benchmark table including the \APDDDRFF.
In particular, \cref{tab:exp_rope3d_full} provides the complete Rope3D~\cite{rope3d} benchmark table.
The results are consistent with those shown in the main paper, further confirming our method's effectiveness across diverse datasets and viewpoints. MonoFlex~\cite{monoflex} achieves higher accuracy on the big vehicle class when supplied with ground-plane inputs. However, its Big Vehicle accuracy degrades significantly by \SI{92.6}{\%} in \APDDDS\ when ground-plane inputs are unavailable.

\subsubsection{Waymo~\cite{waymo}.}
We complement the overall accuracy results from \cref{tab:exp_waymo_light} by reporting \APH\ and accuracy across different distance thresholds.
\Cref{tab:exp_waymo_full} presents the full Waymo~\cite{waymo} results, comparing the baseline with our approach. These results, obtained using the training and validation split defined by DEVIANT~\cite{deviant}, are consistent with the findings in the main paper and show that our method provides clear improvements over the baseline.

\begin{table}[t!]
    \centering
    \medskip
    \caption{\textbf{Detailed Waymo \cite{waymo} validation results.} We report detailed results of \ourmethod and our baseline (\baseline) on the Waymo \cite{waymo} validation set, extending \cref{tab:exp_waymo_light}. We report \AP\ and \APH (both in \%, $\uparrow$) for for Level 1 and Level 2 difficulty as well as different IoU thresholds. We follow the DEVIANT\cite{deviant} training and validation setting.}
    \setlength{\tabcolsep}{0.9pt}
    \centering
    \fontsize{4pt}{5pt}\selectfont
    \begin{tabularx}{\linewidth}{
        >{\centering\arraybackslash}p{0.4cm}
        l
        S[table-format=2.2]
        S[table-format=2.2]
        S[table-format=1.2]
        S[table-format=1.2]
        S[table-format=2.2]
        S[table-format=2.2]
        S[table-format=1.2] 
        S[table-format=1.2]
        S[table-format=2.2]
        S[table-format=2.2]
        S[table-format=1.2]
        S[table-format=1.2]
        S[table-format=2.2]
        S[table-format=2.2]
        S[table-format=1.2]
        S[table-format=1.2]@{}
        }
        
      \toprule
      \multirow{5}{*}{\textbf{IoU}} & \multirow{5}{*}{\textbf{Method}} & 
      \multicolumn{8}{c}{\textbf{Level 1}} & 
      \multicolumn{8}{c}{\textbf{Level 2}} \\
      \cmidrule(lr){3-10} 
      \cmidrule(lr){11-18}       
      & & \multicolumn{4}{c}{\textbf{\AP\,$\uparrow$}} & \multicolumn{4}{c}{\textbf{\APH\,$\uparrow$}} &  \multicolumn{4}{c}{\textbf{\AP\,$\uparrow$}} & \multicolumn{4}{c}{\textbf{\APH\,$\uparrow$}}\\
      
      \cmidrule(lr){3-6} 
      \cmidrule(lr){7-10} 
      \cmidrule(lr){11-14} 
      \cmidrule(lr){15-18}
      & & {\textbf{All}} & {\textbf{0-30}} & {\textbf{30-50}} & {\textbf{50-}} & {\textbf{All}} & {\textbf{0-30}} & {\textbf{30-50}} & {\textbf{50-}} & {\textbf{All}} & {\textbf{0-30}} & {\textbf{30-50}} & {\textbf{50-}} & {\textbf{All}} & {\textbf{0-30}} & {\textbf{30-50}} & {\textbf{50-}} \\ \midrule

      \multirow{12}{*}{0.7} & 
      \baselinen  & 2.34 & 6.00 & 0.73 & 0.07 & 2.31 & 5.93 & 0.72 & 0.07 & 2.06 & 5.90 & 0.66 & 0.05 & 2.04 & 5.83 & 0.66 & 0.05 \\
      & \ourmethodn  & 2.96 & 7.68 & 0.74 & 0.06 & 2.93 & 7.60 & 0.73 & 0.06 & 2.61 & 7.55 & 0.67 & 0.05 & 2.58 & 7.48 & 0.66 & 0.05 \\
      \cmidrulewaymogray
      
      & \baselines  & 2.71 & 7.00 & 0.97 & 0.09 & 2.68 & 6.92 & 0.96 & 0.09 & 2.39 & 6.88 & 0.88 & 0.07 & 2.37 & 6.81 & 0.87 & 0.07 \\
      & \ourmethods  & 3.53 & 8.82 & 1.23 & 0.11 & 3.49 & 8.73 & 1.22 & 0.10 & 3.11 & 8.67 & 1.11 & 0.08 & 3.08 & 8.59 & 1.11 & 0.08 \\
      \cmidrulewaymogray
      
      & \baselinem  & 2.95 & 8.73 & 1.05 & 0.09 & 2.93 & 8.65 & 1.04 & 0.09 & 2.61 & 8.60 & 0.95 & 0.07 & 2.58 & 8.52 & 0.95 & 0.07 \\
      & \ourmethodm  & 3.98 & 9.95 & 1.50 & 0.09 & 3.94 & 9.86 & 1.49 & 0.09 & 3.51 & 9.80 & 1.36 & 0.07 & 3.48 & 9.71 & 1.35 & 0.07 \\
      \cmidrulewaymogray
      
      & \baselineb  & 3.13 & 8.19 & 1.29 & 0.10 & 3.10 & 8.11 & 1.28 & 0.09 & 2.76 & 8.06 & 1.17 & 0.07 & 2.73 & 7.99 & 1.16 & 0.07 \\
      & \ourmethodb  & 4.29 & 10.82 & 1.58 & 0.08 & 4.25 & 10.72 & 1.57 & 0.08 & 3.78 & 10.65 & 1.43 & 0.06 & 3.75 & 10.55 & 1.43 & 0.06 \\
      \cmidrulewaymogray
      
      & \baselinex  & 3.34 & 8.93 & 1.39 & 0.12 & 3.30 & 8.84 & 1.38 & 0.12 & 2.95 & 8.80 & 1.26 & 0.09 & 2.92 & 8.71 & 1.26 & 0.09 \\
      & \ourmethodx  & 4.81 & 12.08 & 1.72 & 0.09 & 4.76 & 11.97 & 1.71 & 0.09 & 4.24 & 11.90 & 1.56 & 0.07 & 4.20 & 11.79 & 1.55 & 0.07 \\
      \midrule

      \multirow{12}{*}{0.5} & 
      \baselinen  & 10.49 & 24.88 & 4.40 & 0.52 & 10.35 & 24.55 & 4.36 & 0.51 & 9.26 & 24.53 & 4.14 & 0.41 & 9.14 & 24.20 & 4.11 & 0.40 \\
      & \ourmethodn  & 12.14 & 28.72 & 4.58 & 0.49 & 12.00 & 28.38 & 4.54 & 0.48 & 10.73 & 28.32 & 4.15 & 0.38 & 10.60 & 27.99 & 4.11 & 0.37 \\
      \cmidrulewaymogray
      
      & \baselines  & 11.18 & 26.12 & 5.58 & 0.69 & 11.04 & 25.80 & 5.54 & 0.68 & 9.88 & 25.77 & 5.08 & 0.54 & 9.76 & 25.45 & 5.04 & 0.53 \\
      & \ourmethods  & 13.24 & 30.59 & 6.41 & 0.81 & 13.09 & 30.23 & 6.36 & 0.80 & 12.08 & 30.17 & 5.81 & 0.64 & 11.94 & 29.82 & 5.76 & 0.63 \\
      \cmidrulewaymogray
      
      & \baselinem  & 11.69 & 29.66 & 6.33 & 0.60 & 11.56 & 29.27 & 6.28 & 0.59 & 10.34 & 29.29 & 5.76 & 0.47 & 10.22 & 28.90 & 5.71 & 0.46 \\
      & \ourmethodm  & 14.55 & 33.54 & 7.11 & 0.63 & 14.40 & 33.20 & 7.07 & 0.62 & 12.86 & 33.12 & 6.68 & 0.50 & 12.74 & 32.78 & 6.64 & 0.49\\
      \cmidrulewaymogray
      
      & \baselineb  &12.89 & 30.51 & 7.07 & 0.66 & 12.76 & 30.18 & 7.01 & 0.66 & 11.40 & 30.13 & 6.43 & 0.52 & 11.28 & 29.80 & 6.37 & 0.51 \\
      & \ourmethodb  & 15.04 & 34.87 & 7.47 & 0.47 & 14.88 & 34.51 & 7.42 & 0.47 & 13.29 & 34.44 & 6.76 & 0.37 & 13.15 & 34.07 & 6.71 & 0.37\\
      \cmidrulewaymogray
      
      & \baselinex  & 13.30 & 31.58 & 7.57 & 0.83 & 13.15 & 31.18 & 7.52 & 0.83 & 11.77 & 31.18 & 6.90 & 0.65 & 11.63 & 30.79 & 6.85 & 0.65 \\
      & \ourmethodx  & 16.46 & 38.42 & 8.30 & 0.58 & 16.28 & 38.01 & 8.25 & 0.58 & 14.54 & 37.93 & 7.55 & 0.46 & 14.39 & 37.52 & 7.50 & 0.45\\

      \bottomrule
    \end{tabularx}
    \label{tab:exp_waymo_full}
\end{table}
\begin{table}[t!]
    \centering
    \caption{\textbf{Additional Waymo~\cite{waymo} validation results.} While we follow the DEVIANT\cite{deviant} training and validation protocol in the main paper (\cf \cref{tab:exp_waymo_light}) and \cref{tab:exp_waymo_full}, here we report results consistent with the training and validation setting by CaDDN~\cite{CaDDN}. We compare against respective models following this setting and report \APDDDF\ and \APHDDDF\ (both in \%, $\uparrow$). We also report runtime in ms for single-image inference on an NVIDIA RTX 8000. \emph{Extra} indicates the use of auxiliary training data.}
    \fontsize{6pt}{7pt}\selectfont
    \setlength{\tabcolsep}{5.65pt}
    \begin{tabularx}{\linewidth}{
        >{\hspace{-\tabcolsep}\raggedright\columncolor{white}[\tabcolsep][\tabcolsep]}X c
        r
        S[table-format=2.2]
        S[table-format=2.2]
        }
        
      \toprule
      \textbf{Method} & \textbf{Extra} & {\textbf{Time\ $\downarrow$}} & {\textbf{\APDDDF\ $\uparrow$}} & {\textbf{\APHDDDF\ $\uparrow$}} \\ 
      \midrule
      CaDDN ~\cite{CaDDN} & LiDAR & 547.0 & 17.54 & 17.31 \\
      CMKD~\cite{cmkd} & LiDAR & 544.2 & 12.99 & 12.90 \\
      DID-M3D \cite{DID-M3D} & LiDAR & 96.7 & 20.66 & 20.47 \\ 
      MonoNeRD ~\cite{mononerd} & LiDAR & 1275.5 & \second 31.18 & \second 30.70 \\
      OccupancyM3D \cite{occupancym3d} & LiDAR & 555.8 & 28.99 & 28.66 \\
      \midrule
      M3D-RPN ~\cite{m3drpn} & --- & \second 65.4 & 3.79 & 3.63 \\
      \ourmethodx (Ours) & --- & \best 28.0 & \best 33.03 & \best 32.71 \\ 
      \bottomrule
    \end{tabularx}
    \label{tab:waymo_alt_eval}

\end{table}

\subsubsection{The DEVIANT~\cite{deviant} \vs CaDDN~\cite{CaDDN} Waymo Protocols.} 
The original Waymo dataset is a multi-view benchmark. CaDDN \cite{CaDDN} adapted it for M3D by using every third frame for training and filtering objects not visible in the front-view camera. Since then, two primary versions have emerged:
\begin{itemize}
    \item The CaDDN protocol applies the same initial object filtering to both the training and validation sets.
    \item The alternative DEVIANT \cite{deviant} protocol is more restrictive. It additionally removes objects whose projected 3D centers fall outside the image from the training set, but retains them in the validation set. 
\end{itemize}
The DEVIANT protocol has become more common in the literature \cite{monoxiver,monocon,monocd,monodgp,ssd-monodetr}). We therefore adopt it for our main experiments (\cf the main paper and \cref{tab:exp_waymo_full}).

For completeness, however, we also provide results on the CaDDN version in \cref{tab:waymo_alt_eval}.
\ourmethodx achieves the highest accuracy while being \num{2} to \num{43}-times faster than existing models. Our analysis of prior work reveals that the primary computational bottlenecks in methods like OccupancyM3D~\cite{occupancym3d}, MonoNeRD~\cite{mononerd}, CaDDN~\cite{CaDDN}, and CMKD~\cite{cmkd} stem from two key stages: lifting the 2D image to a 3D volume and the subsequent 3D volume processing. Specifically, CaDDN~\cite{CaDDN} faces an additional slowdown due to the use of computationally intensive fully-connected layers for depth estimation.

\begin{table}[t]
    \centering
    \caption{\textbf{Cross-dataset generalization to nuScenes.} We evaluate \ourmethodb on the frontal nuScenes~\cite{nuScenes} \emph{val.}\ split following the cross-dataset protocol of DEVIANT~\cite{deviant}, training on KITTI~\cite{kitti} only. We additionally report \ourmethodb\ trained directly on nuScenes as an in-domain reference. We report mean absolute depth error (m, $\downarrow$) across different depth ranges.}
    \footnotesize\sisetup{table-number-alignment=center}
    \setlength{\tabcolsep}{8pt}
    \begin{tabularx}{\columnwidth}{>{\hspace{-\tabcolsep}\raggedright\columncolor{white}[\tabcolsep][\tabcolsep]}X c S[table-format=1.2] S[table-format=1.2] S[table-format=1.2] S[table-format=1.2]}
        \toprule
        \textbf{Method} & \textbf{Train} & \textbf{0--20\,$\downarrow$} & \textbf{20--40\,$\downarrow$} & \textbf{40--$\infty$\,$\downarrow$} & \textbf{All\,$\downarrow$} \\
        \midrule
        GUPNet~\cite{gupnet}   & KITTI & 0.82 & 1.70 & 6.20 & 1.45 \\
        MonoUNI~\cite{monouni} & KITTI & 0.72 & 1.79 & 4.98 & 1.43 \\
        MonoLSS~\cite{monolss} & KITTI & 0.59 & 2.01 & 5.40 & 1.42 \\
        MonoMAE~\cite{monomae} & KITTI & 0.71 & 1.57 & 4.95 & 1.40 \\
        MonoCon~\cite{monocon} & KITTI & 0.78 & 1.65 & 6.02 & 1.40 \\
        MonoCD~\cite{monocd}   & KITTI & 0.73 & 1.59 & 5.78 & 1.33 \\
        DEVIANT~\cite{deviant} & KITTI & 0.76 & 1.60 & 4.50 & 1.26 \\
        \midrule
        \ourmethodb & KITTI      & \second 0.58 & \second 1.34 & \second 3.69 & \second 1.07 \\
        \ourmethodb & nuScenes & \best 0.43 & \best 0.94 & \best 1.67 & \best 0.84 \\
        \bottomrule
    \end{tabularx}
    \label{tab:rebuttal_nuscenes}
\end{table}

\newlength{\imgW}
\newlength{\imgWW}
\setlength{\imgW}{0.46\textwidth} 
\setlength{\imgWW}{0.5\textwidth} 
\setlength{\extrarowheight}{0.3em}

\begin{figure*}[ht!] 
\centering
\renewcommand{\arraystretch}{1.1}
\begin{tabular}{c@{\hspace{0.8em}}c@{\hspace{0.8em}}c@{}} 

\sffamily RGB w/ Ours & \sffamily BEV \\

\includegraphics[width=\imgW]{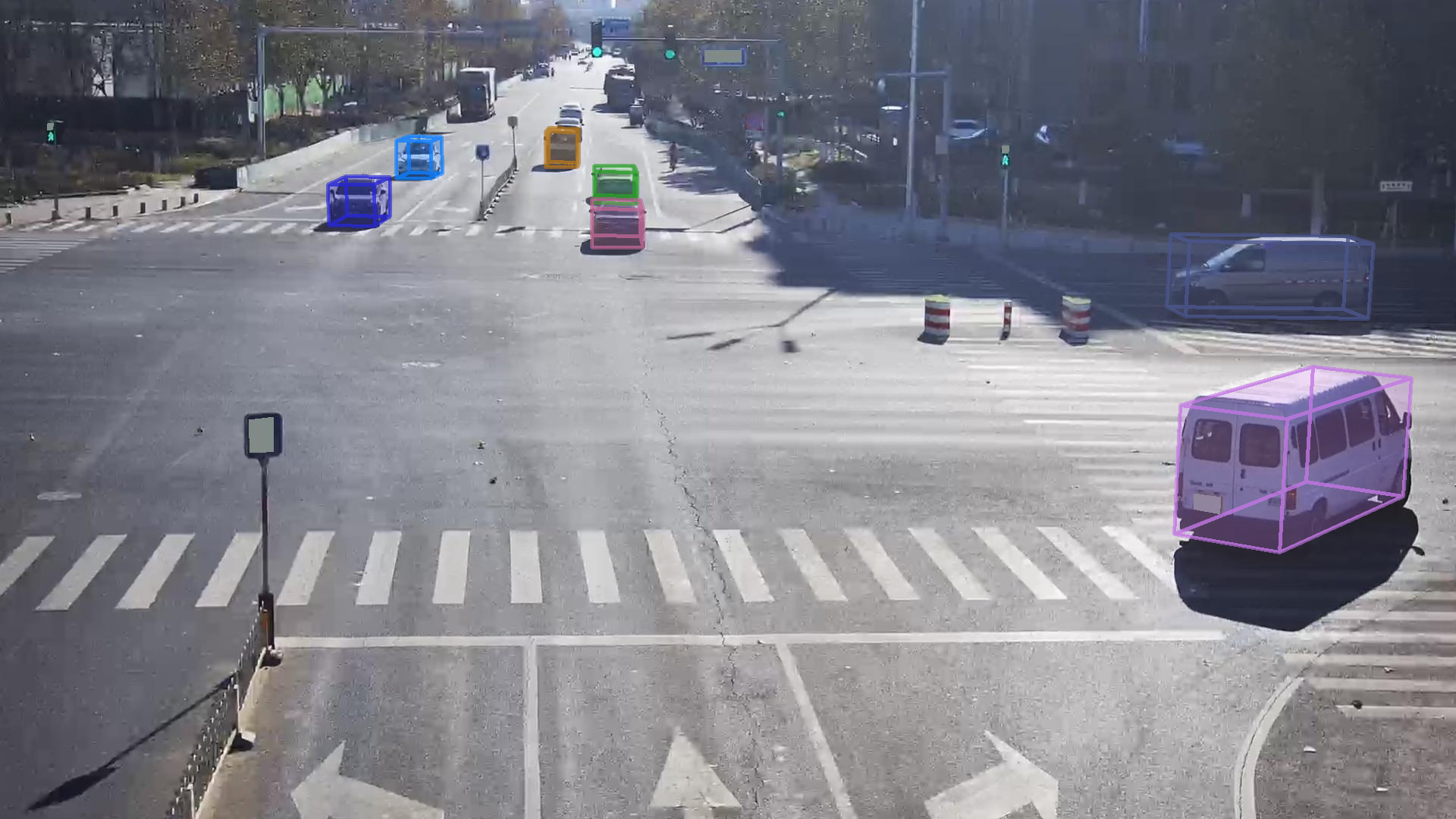} &
\begin{tikzpicture}[spy using outlines={black, magnification=3.5, minimum width=0.3\imgWW, minimum height=0.3\imgWW, fill=white, connect spies}]
        \node[inner sep=0pt, outer sep=0pt] at (0, 0) {\includegraphics[width=\imgWW]{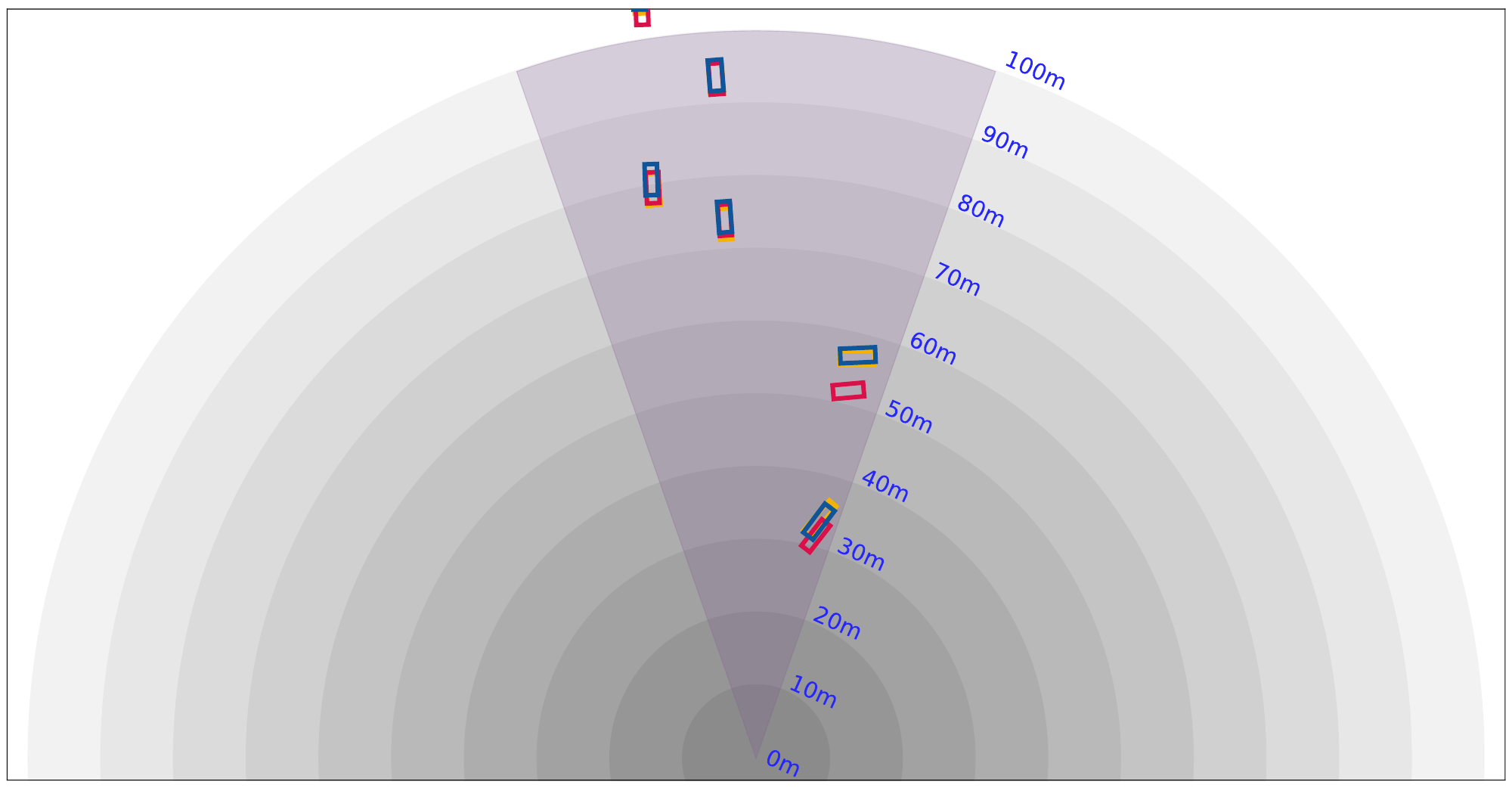}};\spy on (-0.23, 0.745) in node [draw opacity=1.0, black, fill=white, anchor=center, line width=0.2pt, inner sep=0pt, outer sep=0pt] at (-0.345\imgWW, -0.104\imgWW);
        \spy on (0.4, 0.1) in node [draw opacity=1.0, black, fill=white, anchor=center, line width=0.2pt, inner sep=0pt, outer sep=0pt] at (0.345\imgWW, -0.104\imgWW);
\end{tikzpicture} \\

\includegraphics[width=\imgW]{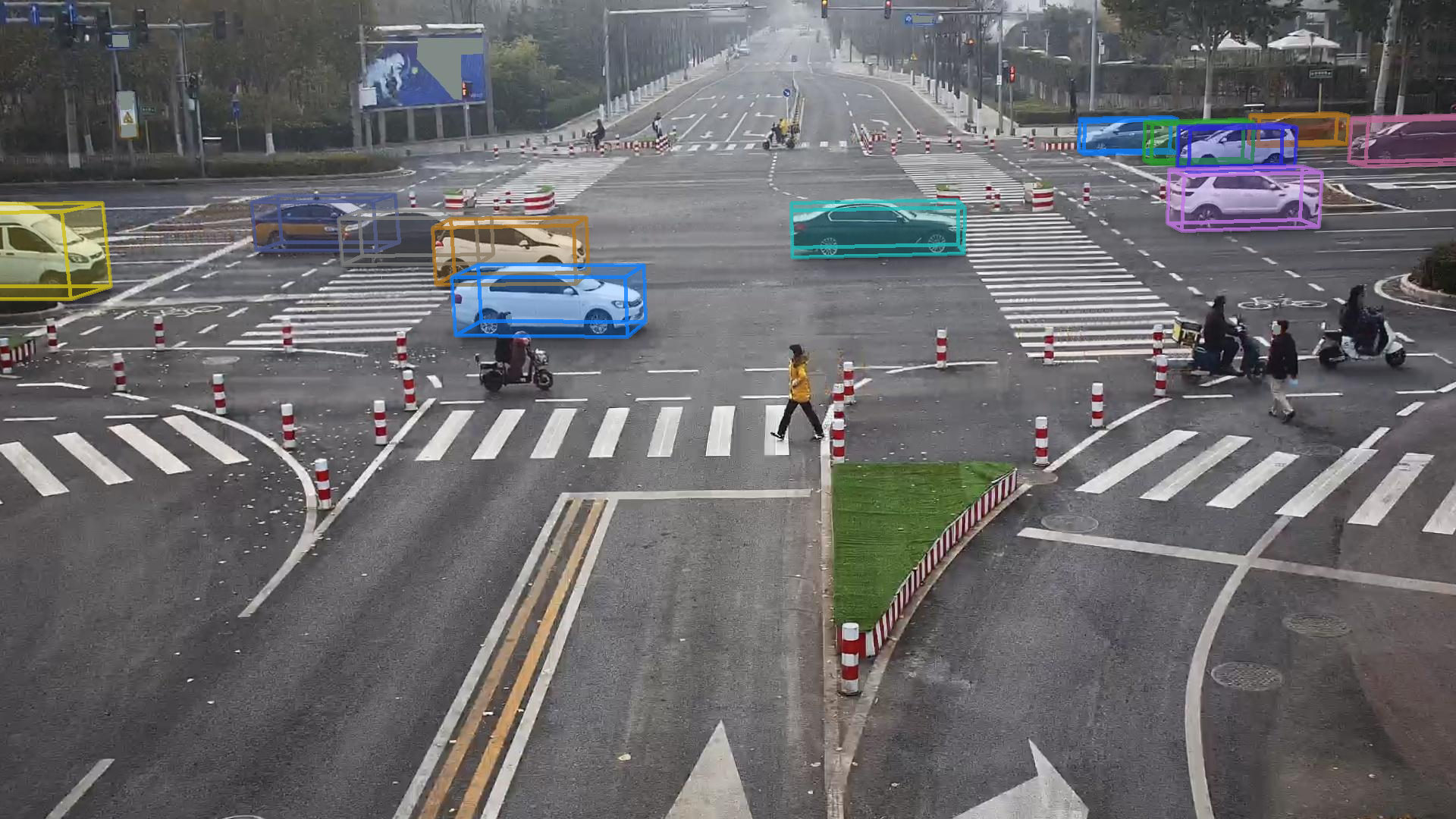} &
\begin{tikzpicture}[spy using outlines={black, magnification=3.5, minimum width=0.3\imgWW, minimum height=0.3\imgWW, fill=white, connect spies}]
        \node[inner sep=0pt, outer sep=0pt] at (0, 0) {\includegraphics[width=\imgWW]{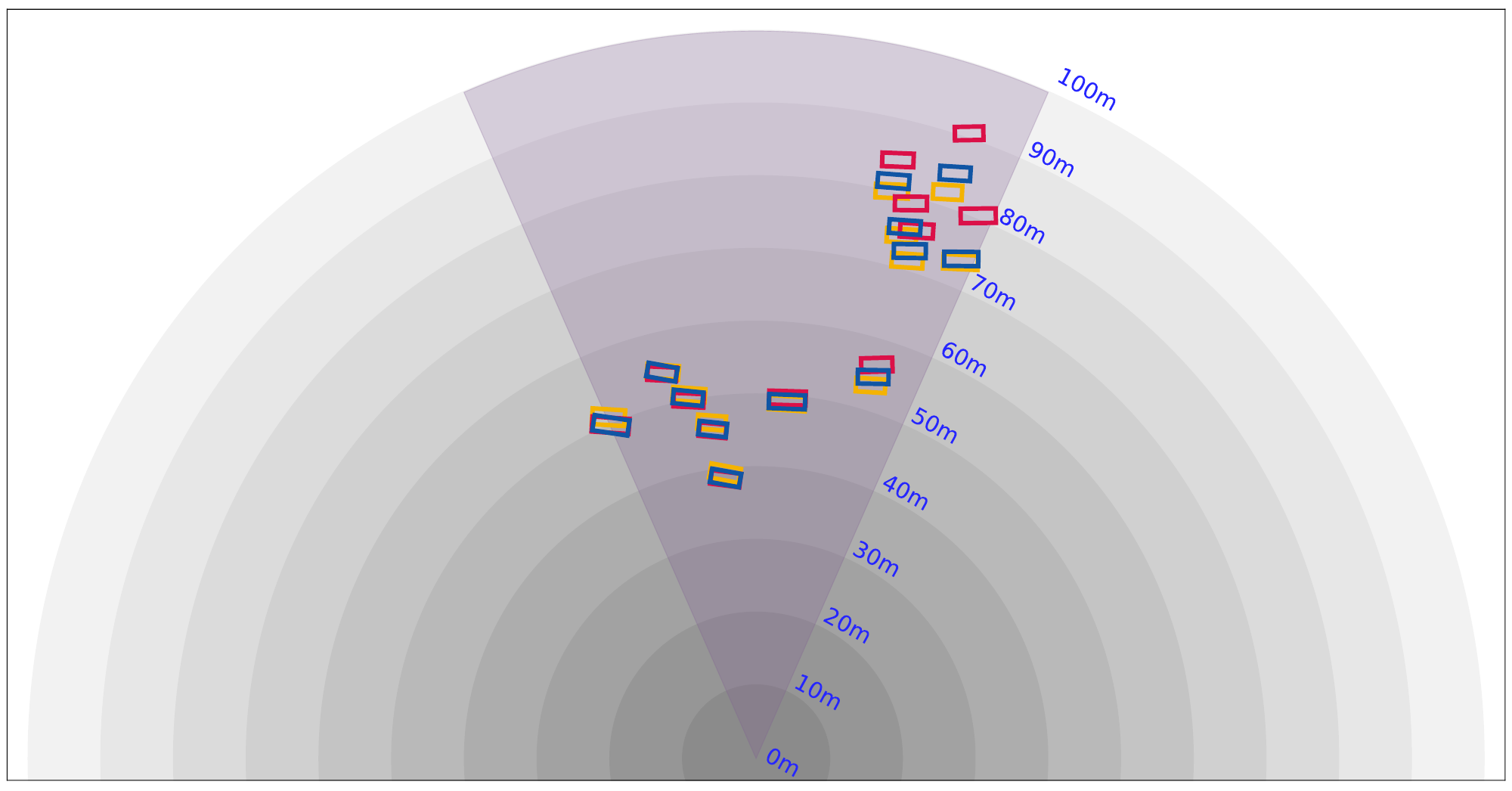}};\spy on (-0.2, -0.1) in node [draw opacity=1.0, black, fill=white, anchor=center, line width=0.2pt, inner sep=0pt, outer sep=0pt] at (-0.345\imgWW, -0.104\imgWW);
        \spy on (0.6, 0.745) in node [draw opacity=1.0, black, fill=white, anchor=center, line width=0.2pt, inner sep=0pt, outer sep=0pt] at (0.345\imgWW, -0.104\imgWW);
\end{tikzpicture} \\

\includegraphics[width=\imgW]{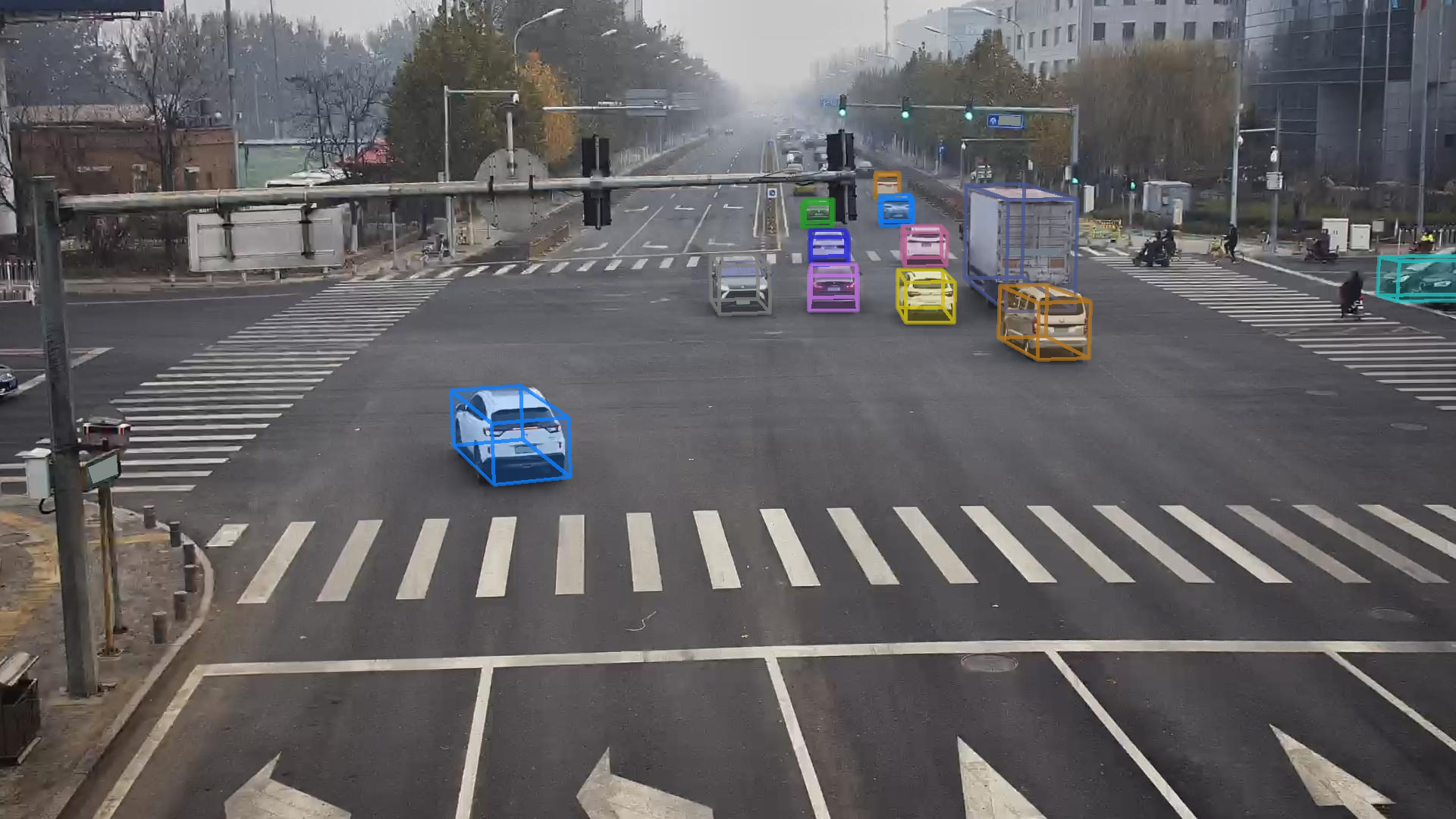} &
\begin{tikzpicture}[spy using outlines={black, magnification=3.5, minimum width=0.3\imgWW, minimum height=0.3\imgWW, fill=white, connect spies}]
        \node[inner sep=0pt, outer sep=0pt] at (0, 0) {\includegraphics[width=\imgWW]{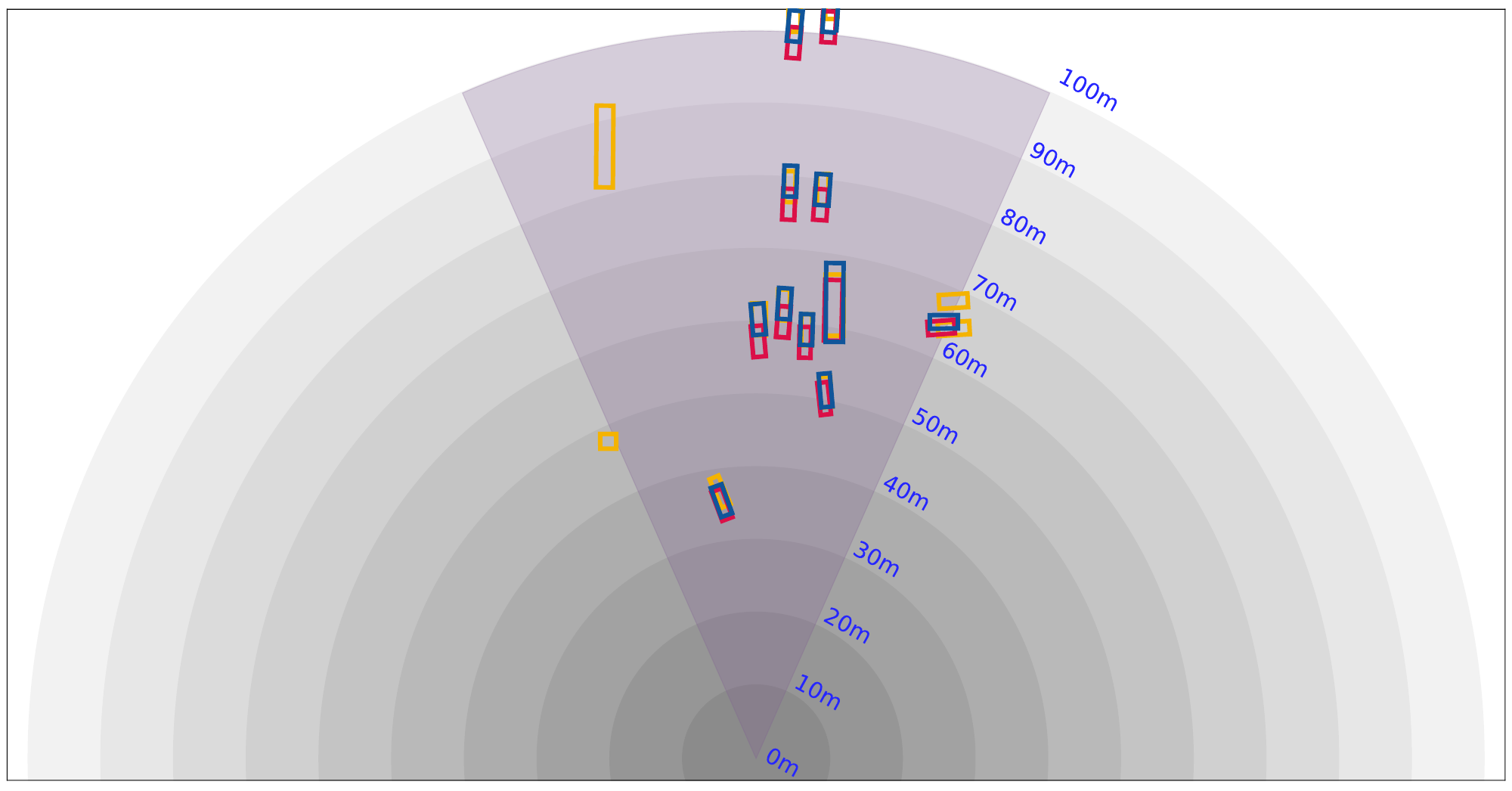}};\spy on (-0.2, -0.4) in node [draw opacity=1.0, black, fill=white, anchor=center, line width=0.2pt, inner sep=0pt, outer sep=0pt] at (-0.345\imgWW, -0.104\imgWW);
        \spy on (0.2, 0.75) in node [draw opacity=1.0, black, fill=white, anchor=center, line width=0.2pt, inner sep=0pt, outer sep=0pt] at (0.345\imgWW, -0.104\imgWW);
\end{tikzpicture} \\

\includegraphics[width=\imgW]{figures/qual_rope3d/rope3d_150666_sj8fas10n151d20211202air_420_1637548012_1637553059_209_obstacle_ours.png} &
\begin{tikzpicture}[spy using outlines={black, magnification=3.5, minimum width=0.3\imgWW, minimum height=0.3\imgWW, fill=white, connect spies}]
        \node[inner sep=0pt, outer sep=0pt] at (0, 0) {\includegraphics[width=\imgWW]{figures/qual_rope3d/rope3d_150666_sj8fas10n151d20211202air_420_1637548012_1637553059_209_obstacle_cr}};\spy on (-0.3, 0.75) in node [draw opacity=1.0, black, fill=white, anchor=center, line width=0.2pt, inner sep=0pt, outer sep=0pt] at (-0.345\imgWW, -0.104\imgWW);
        \spy on (0.4, 0.1) in node [draw opacity=1.0, black, fill=white, anchor=center, line width=0.2pt, inner sep=0pt, outer sep=0pt] at (0.345\imgWW, -0.104\imgWW);
\end{tikzpicture} \\

\end{tabular}
\caption{
\textbf{Qualitative results on the Rope3D~\cite{rope3d} heterologous validation set.} \ourmethodx achieves more accurate depth estimates than \baselinex. We highlight improved detections in BEV. Best viewed in color and with zoom. {BEV color coding}: Ground truth \colorindicator{qualGTcr}, \baselinex\ \colorindicator{qualBLcr}, \ourmethodx\ \colorindicator{qualOURcr}, and field of view \colorindicator{qualFOV}.
}
\label{fig:qualit_rope3d}
\end{figure*}

\subsubsection{Cross-Dataset Generalization to nuScenes~\cite{nuScenes}.}
\label{sec:supp_nuscenes_cross_dataset}
We evaluate on the cross-dataset generalization benchmark proposed by Kumar~\etal~\cite{deviant}, training on KITTI~\cite{kitti} and evaluating on nuScenes~\cite{nuScenes}. As shown in \cref{tab:rebuttal_nuscenes}, \ourmethodb outperforms all competing KITTI-trained methods, achieving an overall mean absolute depth error of \SI{1.07}{m} \vs \SI{1.26}{m} for DEVIANT~\cite{deviant}, demonstrating the generalizability of our approach. Training directly on nuScenes further improves accuracy to \SI{0.84}{m} overall. Complementary to this, the main paper evaluates generalization in the opposite direction, training on nuScenes and evaluating on KITTI using \APDDDRFF\ (\cf \cref{tab:rebuttal_domgen}).

\subsection{Qualitative Results on Rope3D}
\label{sec:qual_results_rope3d}
In \cref{fig:qualit_rope3d}, we provide qualitative results on Rope3D \cite{rope3d}. Compared with KITTI \cite{kitti}, Rope3D spans a substantially wider depth range, which makes accurate monocular depth estimation even more challenging. Additionally, Rope3D includes more diverse viewpoints. Both methods exhibit some false negatives under heavy occlusion, but our \ourmethod consistently produces more accurate depth estimates than the \baseline. In many examples, the improvement is significant, with depth errors reduced by roughly \num{5}–\SI{8}{m} \wrt to the baseline.

\section{Future Work}
\label{sec:future_work}
Our \ourmethod framework establishes a solid foundation for efficient real-time monocular 3D detection. While we focus on supervised M3D and demonstrate strong results, \ourmethod also provides a basis for future research beyond this specific setting. \emph{First}, extending A2D2 to the semi-supervised setting could enable training on large-scale datasets of labeled and unlabeled images. As suggested by our strong domain-generalization results, scaling data could potentially further improve robustness and generalization to different domains.

\emph{Second}, while we focus on standard M3D benchmarks, extending \ourmethod to the open-vocabulary setting could enable a powerful model. While current open-vocabulary M3D approaches utilize pseudo-annotations and self-training, A2D2 could be utilized as an effective learning scheme in this setting. Data scaling and learning open-vocabulary representations through A2D2 could together provide a promising avenue toward a foundational zero-shot approach for M3D.

\emph{Third}, while we focus on the \emph{monocular} setting, the most widespread and general setting, many real-world applications provide video. Distilling and denoising both within the spatial and temporal domains using A2D2 could enable learning of temporally consistent and accurate representations that aid detection in videos. While A2D2 could potentially function out of the box for multi-frame settings, enforcing extra consistency by exploiting the temporal domain through specific augmentations could be an effective strategy. Similar principles could apply to multi-view configurations, where cross-camera geometric constraints could further regularize distillation.

\end{document}